\documentclass{article}
\PassOptionsToPackage{numbers,compress}{natbib}
\usepackage[final,main]{neurips_2026}
\usepackage[utf8]{inputenc} 
\usepackage[T1]{fontenc}    
\usepackage{hyperref}       
\usepackage{url}            
\usepackage{booktabs}       
\usepackage{amsfonts}       
\usepackage{nicefrac}       
\usepackage{microtype}      
\usepackage{xcolor}         
\usepackage{amsmath}
\usepackage{graphicx}
\usepackage{subcaption}
\usepackage{multirow}
\usepackage{afterpage}
\title{MESS: Multi-Exposure Sequence Synthesis for Generalizable Image Enhancement}

\author{%
  Ruodai Cui$^{1,3}$ \quad Shuaizheng Liu$^{1,2}$ \quad
  Rongyuan Wu$^{1,2}$ \quad Lei Zhang$^{1,2}$\thanks{Corresponding author. This work is partly supported by the PolyU-OPPO Joint Innovative Research Center.} \\
  $^{1}$Department of Computing, The Hong Kong Polytechnic University \\
  $^{2}$OPPO Research Institute \quad
  $^{3}$Qualcomm Technologies, Inc. \\
  \texttt{\{ruodai.cui, shuaizheng.liu, rong-yuan.wu\}@connect.polyu.hk} \\
  \texttt{cslzhang@comp.polyu.edu.hk}
}

\begin{document}

\maketitle

\begin{abstract}
Despite the significant progress in image enhancement under adverse illuminations, such as low-light image enhancement, exposure correction, and backlit image enhancement, existing models show limited generalizability to different scenarios. A major bottleneck lies in the lack of large-scale and diverse training data, as these tasks typically rely on manually captured, precisely aligned image pairs that are expensive to scale up. Although synthetic data have been explored as an alternative, existing synthesis methods are limited in modeling complex adverse illumination conditions and camera imaging pipelines. To address this issue, we propose a novel framework to synthesize realistic training pairs for generalizable image enhancement in the wild. Specifically, we first derive multi-exposure sequences (MES) from high-bit RAW images by rendering them under different exposure settings through an emulated ISP pipeline. Using these RAW-derived MES as supervision, we train a diffusion-based RGB-to-MES generator to synthesize plausible exposure sequences from a single 8-bit RGB image, extending the synthesis pipeline from limited RAW collections to large-scale in-the-wild RGB data. For each source image, we randomly sample one synthesized exposure variant as the degraded input and use the original high-quality RGB image as the target. With the proposed \textbf{MES S}ynthesis (\textbf{MESS}) approach, we construct a dataset of 500K training pairs and train a lightweight network, which demonstrates significantly better generalization performance than existing models across various image enhancement tasks. Code and data are available at \url{https://github.com/PolyU-VCLab/MESS}.
\end{abstract}

\section{Introduction}
\begin{figure}[t]
    \centering
    \begin{tabular*}{\linewidth}{@{}p{0.25\linewidth}@{}p{0.25\linewidth}@{}p{0.25\linewidth}@{}p{0.25\linewidth}@{}}
        \centering Input &
        \centering Rformer-UHDLL &
        \centering Rformer-Ours &
        \centering GT
    \end{tabular*}
    \includegraphics[width=\linewidth]{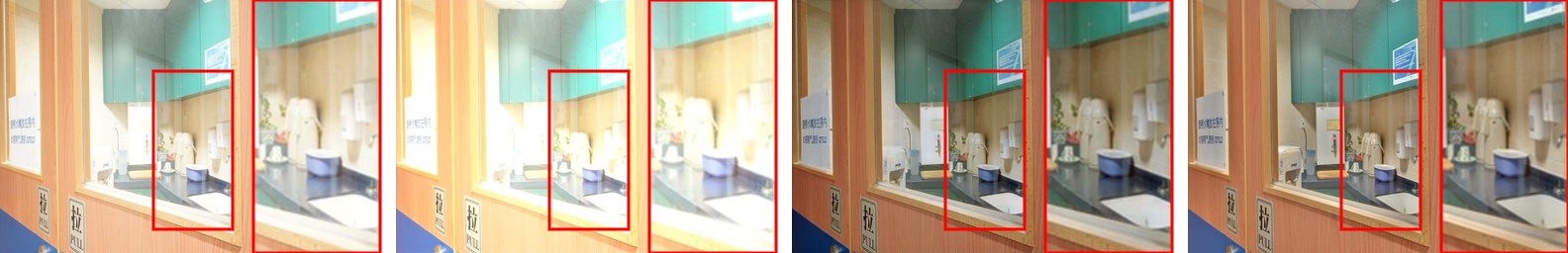} \\
    \includegraphics[width=\linewidth]{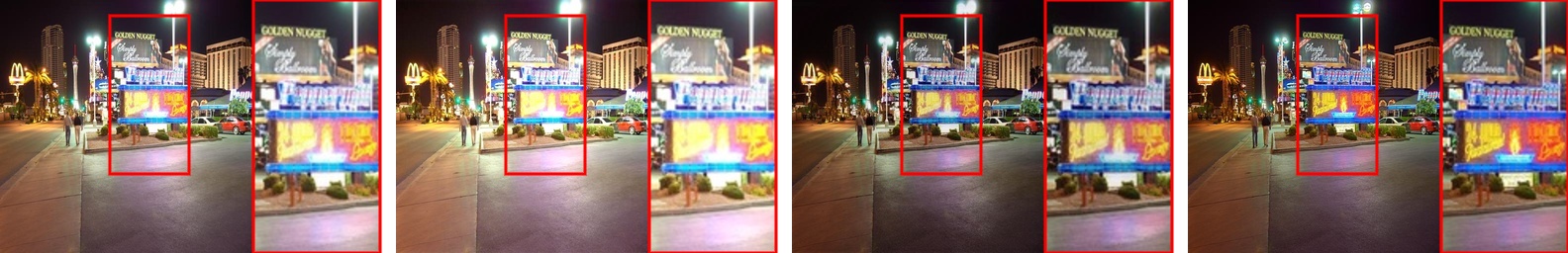}
    \caption{Visual comparisons of a recent state-of-the-art low-light image enhancement method, Rformer~\cite{Cai2023Retinexformer}, trained on the UHDLL dataset~\cite{li2023uhdfour} (2nd column) and on our large-scale synthetic dataset (3rd column). Zoomed-in regions are highlighted with red boxes for better comparison.}
    \label{fig:1st_demo}
    \vspace{-5mm}
\end{figure}

Images captured under adverse illumination conditions often suffer from underexposure, overexposure, and low contrast, resulting in poor visibility and degraded visual quality~\cite{Wang2013NPE,Guo2016LIME,Liu2021Benchmarking}. Many image enhancement methods have been developed for specific scenarios, including low-light image enhancement (LLIE)~\cite{Chen2018SID,Wei2018RetinexNet,Guo2020ZeroDCE,Jiang2021EnlightenGAN,Liu2021RUAS,Cai2023Retinexformer}, exposure correction (EC)~\cite{Wang2019DeepUPE,Afifi2021Learning,Huang2022ENC,Wang2023DAEC,Li2024CSEC}, and backlit image enhancement (BIE)~\cite{Liang2023CLIPLIT,gaintseva2024RAVE}. Despite their progress, most existing methods are designed for a single enhancement task and trained on narrow data distributions, showing limited generalization performance. As shown in Fig.~\ref{fig:1st_demo}, a model trained for LLIE may fail when applied to exposure correction, producing inaccurate brightness adjustment and losing local details. Moreover, real-world images often contain mixed illumination degradations, e.g., underexposed and overexposed regions appearing simultaneously, which further increases the difficulty of image enhancement in the wild. Although recent methods attempt to improve generalization by using larger models or more complex designs~\cite{Wang2023IAGC,wang2023exposurediffusion,Cai2023Retinexformer,Wang2024QuadPrior,Lamba2021ExtremeDark}, their performance under diverse real-world conditions remains limited.

We argue that a key bottleneck for generalizable illumination enhancement is the lack of large-scale, diverse, and reliable paired training data. Existing datasets are typically built from manually captured and carefully aligned image pairs, where a degraded image is paired with a better-exposed reference~\cite{Chen2018SID,Wei2018RetinexNet,Cai2018SICE,Liu2021Benchmarking}. However, this process is costly and difficult to scale, resulting in limited coverage of real-world scenes and degradation types. As a consequence, models trained on such datasets often struggle to generalize beyond their target task or data domain.

\begin{figure}[t]
    \centering
    \includegraphics[width=1\linewidth]{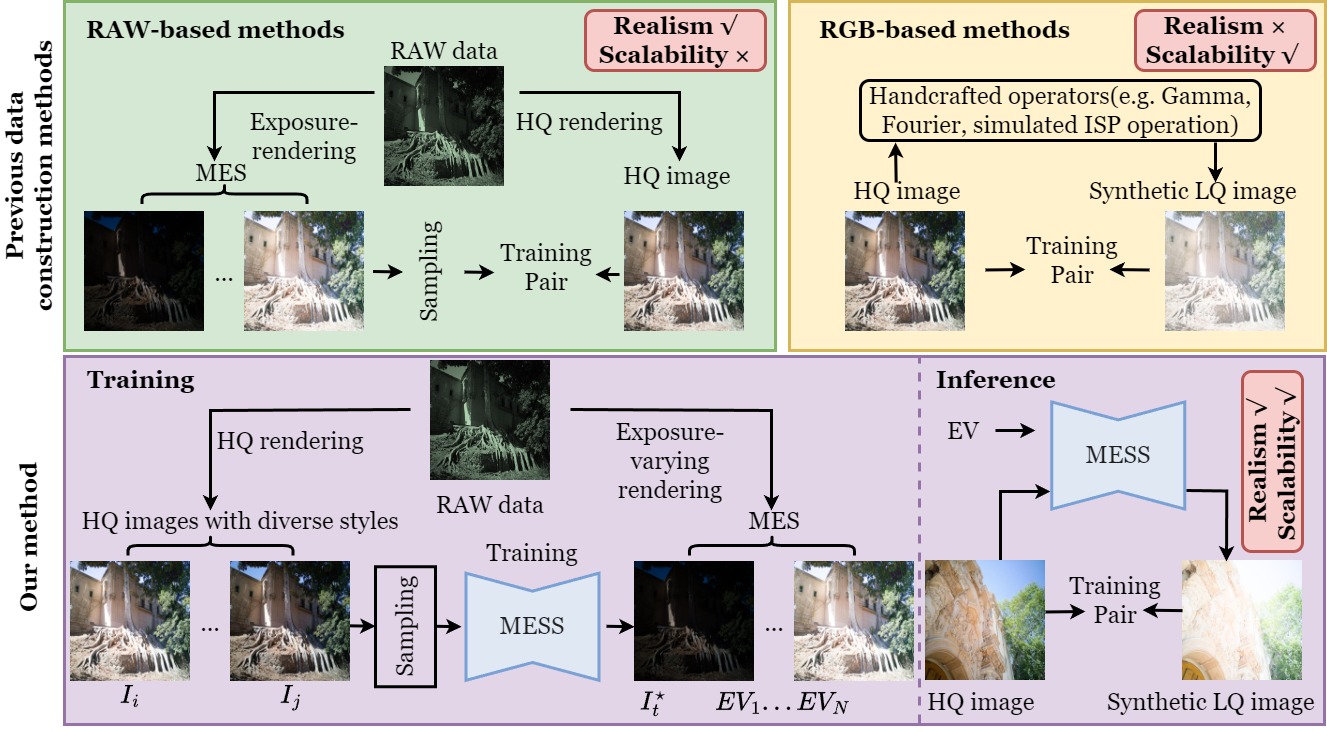}
    \caption{\textbf{Comparison of different data construction strategies for image illumination enhancement.}
    RAW-based methods (top left) produce realistic exposure variations but lack scalability, while RGB-based methods (top right) are scalable but often less realistic. In contrast, our proposed \textbf{MESS} (bottom) transfers RAW-derived exposure priors to in-the-wild RGB images, enabling both realistic and scalable synthesis of training pairs. Notions are defined in \S\ref{sec:method}.}
    \label{fig:method_cmp}
    \vspace{-3mm}
\end{figure}

A natural alternative is to synthesize paired training data by applying controlled degradations to high-quality images. However, realistic and scalable synthesis is a challenging task. As shown in the top-left of Fig.~\ref{fig:method_cmp}, one line of research performs exposure manipulation in the high-bit RAW domain using image signal processing (ISP) operators~\cite{Afifi2021Learning,Cui2024UEC,Cui2025UNICE}. These methods produce realistic exposure variation, but their scalability is constrained by the limited availability of high-quality RAW data. Another line of work (top-right of Fig.~\ref{fig:method_cmp}) starts from 8-bit RGB images and reverse the ISP to mimic exposure changes in a pseudo-RAW space~\cite{Xing2021InvISP,Kim2024ParamISP}. While more scalable, such approximations often fail to capture the complexity of real camera pipelines and illumination variations. Once a high-bit RAW image has been rendered into 8-bit RGB space, its dynamic range is already clipped or compressed, making faithful inverse exposure manipulation fundamentally difficult.

In this work, we propose \textbf{M}ulti-\textbf{E}xposure \textbf{S}equences \textbf{S}ynthesis (\textbf{MESS}), a framework that transfers RAW-derived exposure variations to large-scale in-the-wild RGB images. Given a high-quality RGB image, MESS synthesizes a realistic multi-exposure sequence (MES), from which degraded-target pairs can be automatically constructed for training generalizable enhancement models. The key idea is to use RAW images to provide realistic supervision for exposure variation, while learning an RGB-to-MES generator that transfers such variation to ordinary RGB images at scale. In this way, MESS combines the physical realism of RAW-based exposure rendering with the scalability of large RGB image collections, as shown in the bottom of Fig.~\ref{fig:method_cmp}.

In particular, MESS is built upon three components. First, we exploit RAW images to construct two types of supervision: multi-style high-quality renderings of the same scene, and camera-consistent exposure renderings generated by varying only the exposure setting while keeping the underlying ISP fixed. Based on this design, we introduce a frequency-based enhancement module to reduce rendering-style discrepancies while preserving scene structure and local details. Second, we incorporate a diffusion prior to compensate for information loss in 8-bit RGB images and synthesize structurally plausible missing content. Third, we design an exposure-conditioned rendering module that injects the target exposure condition and performs exposure-aware feature modulation, producing outputs that better match realistic camera characteristics.

Based on MESS, we curate 500K high-quality in-the-wild RGB images and synthesize exposure-perturbed counterparts to automatically construct a large-scale paired training set. For each source image, one synthesized exposure variant is sampled from the generated MES as the degraded input, while the original high-quality image is used as the target. To validate the effectiveness of the synthesized data, we train a lightweight enhancement network on the resulting dataset. Despite its compact model size, the trained model generalizes better than existing methods across multiple real-world illumination enhancement benchmarks.

Our contributions are summarized as follows. (1) We propose \textbf{MESS}, a new framework for scalable and realistic training data synthesis for generalizable image enhancement by transferring RAW-derived exposure variations to in-the-wild RGB images. (2) We develop an MES synthesis pipeline, including frequency-based enhancement for style-robust scene representation, diffusion-based intermediate synthesis for detail completion, and exposure-conditioned rendering for realistic exposure-aware output generation. (3) We build a large-scale synthetic dataset of 500K paired samples with MESS, and show that even a lightweight enhancement model trained on this dataset can achieve strong generalization across multiple real-world image enhancement tasks.

\section{Related Work}

\textbf{Image Illumination Enhancement Methods and Datasets}.
Existing work on illumination enhancement can be broadly categorized into three related but distinct tasks: low-light image enhancement (LLIE)~\cite{Hai2023R2RNet, jiang2024lightendiffusion, li2023uhdfour, zerodcepp}, exposure correction (EC)~\cite{Li2024CSEC, Cui2024UEC, huang2022FECNet, Afifi2021Learning, huang2023sample_relationship}, and backlit image enhancement (BIE)~\cite{Liang2023CLIPLIT, Lv2022BacklitNet, ueda2020backlit}. Many methods can achieve strong performance on their target benchmarks by using specialized architectures such as CNNs, Transformers, and diffusion models~\cite{jiang2024lightendiffusion, wang2023exposurediffusion, feng2024difflight, Li2024CSEC}. However, most of them are developed for a single task and trained on relatively narrow data distributions, limiting the generalization to real-world images with mixed degradations.
This limitation is related to the problems of existing datasets. Popular benchmarks such as LOLv2Real~\cite{Yang2021SGM}, UHD-LL~\cite{li2023uhdfour}, LSRW~\cite{Hai2023R2RNet}, BAID~\cite{Lv2022BacklitNet}, BAID~\cite{Lv2022BacklitNet} and SICE~\cite{Cai2018SICE} are limited in scale, scene diversity and degradation coverage. As shown in Table~\ref{tab:dataset}, LOLv1 and LOLv2Real contain only a few hundred images, while larger datasets such as LSRW~\cite{Hai2023R2RNet}, UHD-LL~\cite{li2023uhdfour}, BAID~\cite{Lv2022BacklitNet}, and SICE~\cite{Cai2018SICE} still contain only a few thousand images. This limited scale makes it difficult to achieve strong generalization across diverse real-world scenarios. Some studies have explored ISP-like operations in the RAW domain, including MSEC~\cite{Afifi2021Learning} and LCDP~\cite{Wang2022LCDPNet}, which partially alleviates this issue by enabling more physically grounded exposure simulation through ISP rendering~\cite{Afifi2021Learning, Wang2022LCDPNet}. However, these methods are still constrained by the availability of high-quality RAW datasets.

\textbf{Synthetic Data for Illumination Enhancement}.
Synthetic data provide a practical alternative to the costly collection of real paired data for image enhancement. For example, LLNet~\cite{Lore2017LLNet} generates degraded samples through gamma-based perturbations of high-quality reference images, whereas CSNorm~\cite{Yao2023CSNorm} adopts a Fourier-transform-based degradation strategy. However, such approaches often struggle to produce physically realistic degradations.
Since illumination variation is approximately linear in the RAW domain, mapping sRGB images back to RAW space enables more physically grounded simulation of illumination degradations. Motivated by this observation, several ISP-aware synthesis pipelines have been proposed. For instance, Unprocessing~\cite{Brooks2019Unprocessing} and CycleISP~\cite{Zamir2020CycleISP} show that explicit modeling of the camera ISP can generate more realistic paired data and improve generalization to real-world images. InvISP~\cite{Xing2021InvISP} formulates the ISP as an invertible pipeline to reconstruct RAW images from sRGB inputs, while ParamISP~\cite{Kim2024ParamISP} further incorporates camera metadata, such as ISO and exposure time, into the ISP model. Following this line of research, recent LLIE-oriented synthesis pipelines~\cite{Long2025SynLLIE,Wang2025ISPDrivenLLIE} first map sRGB images back to RAW space, then simulate low-light degradations and ISP variations, and finally render the results back to sRGB for training.
Nevertheless, the mapping from RAW to RGB is inherently irreversible, as the dynamic range is clipped or compressed during the imaging pipeline. Hence, simplified ISP operators alone are insufficient to faithfully recover RAW from sRGB images, which limits the realism and fidelity of the synthesized data.

\begin{table}[t]
\caption{Summary of existing datasets for image enhancement.
``\#images'' denotes the number of training images, while
``\#scenes'' denotes the number of unique scenes. For datasets
LOLv2Real~\cite{Yang2021SGM}, UHD-LL~\cite{li2023uhdfour},
LSRW~\cite{Hai2023R2RNet}, and BAID~\cite{Lv2022BacklitNet},
the exact number of scenes is not provided. We treat images with a
CLIP similarity score greater than 0.9 as duplicated images from the same scene.}
\label{tab:dataset}
\centering
\begin{tabular}{c l r r|c l r r}
\hline
Task & Dataset & \#images & \#scenes & Task & Dataset & \#images & \#scenes \\
\hline
LLIE & LOLv2Real~\cite{Yang2021SGM}        &   789  &   263 & EC   & LCDP~\cite{Wang2022LCDPNet}         & 1,700 & 1,700 \\
LLIE & LSRW~\cite{Hai2023R2RNet}           & 5,650  & 4,157 & EC   & SICE~\cite{Cai2018SICE}             & 4,800 &   589 \\
LLIE & UHD-LL~\cite{li2023uhdfour}      & 2,150  &   724 & BIE  & BAID~\cite{Lv2022BacklitNet}        & 3,000 & 1,782 \\
EC   & MSEC~\cite{Afifi2021Learning}   & 24,330 & 5,000 & BIE  & Backlit300~\cite{Liang2023CLIPLIT}  &   305 &   305 \\
\hline
\end{tabular}
\vspace{-2mm}
\end{table}

\section{Multi-Exposure Sequence Synthesis: Methodology}
\label{sec:method}

This section describes the proposed MESS framework and its training strategy.
We first present the data preparation procedure and overall training pipeline in \S\ref{subsec:framework}, including RAW-guided supervision for learning camera-consistent exposure variation and RGB-based synthesis for large-scale paired data construction.
We then introduce the three main components: frequency-based enhancement in \S\ref{subsec:frequency}, diffusion-based detail enhancement in \S\ref{subsec:diffusion}, and exposure-conditioned rendering in \S\ref{subsec:rendering}.
Finally, we summarize the training objectives in \S\ref{subsec:objectives}.

\subsection{Training Framework Overview}
\label{subsec:framework}

\textbf{Training Data Preparation.}
As shown in the bottom left of Fig.~\ref{fig:method_cmp}, we construct RAW-guided triplets \(\{I_i, I_j, I_t^{\star}\}\) from each high-bit RAW image \(X\).
Specifically, \(I_i = \mathcal{T}_{\mathrm{sty}}(X; s_i)\) and \(I_j = \mathcal{T}_{\mathrm{sty}}(X; s_j)\), where \(\mathcal{T}_{\mathrm{sty}}\) denotes style-varied rendering, and \(s_i\) and \(s_j\) are rendering-style parameters. Meanwhile, \(I_t^{\star} = \mathcal{T}_{\mathrm{cam}}(X; e_t)\) is the image rendered under the target exposure using a camera-consistent ISP pipeline, where \(\mathcal{T}_{\mathrm{cam}}\) denotes camera-consistent exposure rendering and \(e_t \in [0,1]\) is the normalized exposure value.
The image pair \(\{I_i, I_j\}\) is used to learn a style-robust scene representation, while the image \(I_t^{\star}\) supervises exposure-controlled synthesis. More details of the training data construction process can be found in \textbf{Appendix A}.

\textbf{Training Pipeline.}
MESS aims to synthesize realistic multi-exposure sequences (MES) from high-quality RGB images in-the-wild, enabling scalable paired training data construction.
Given a source RGB image \(I_i\) and a normalized target exposure value \(e_t\), MESS predicts an exposure-controlled output \(\hat{I}_t = \mathrm{MESS}(I_i, e_t)\).
Both \(\hat{I}_t\) and the RAW-rendered reference \(I_t^{\star}\) correspond to the same target exposure.
By varying \(e_t\), the model generates an MES for the same input image.

\begin{figure}
    \centering
    \includegraphics[width=1\linewidth]{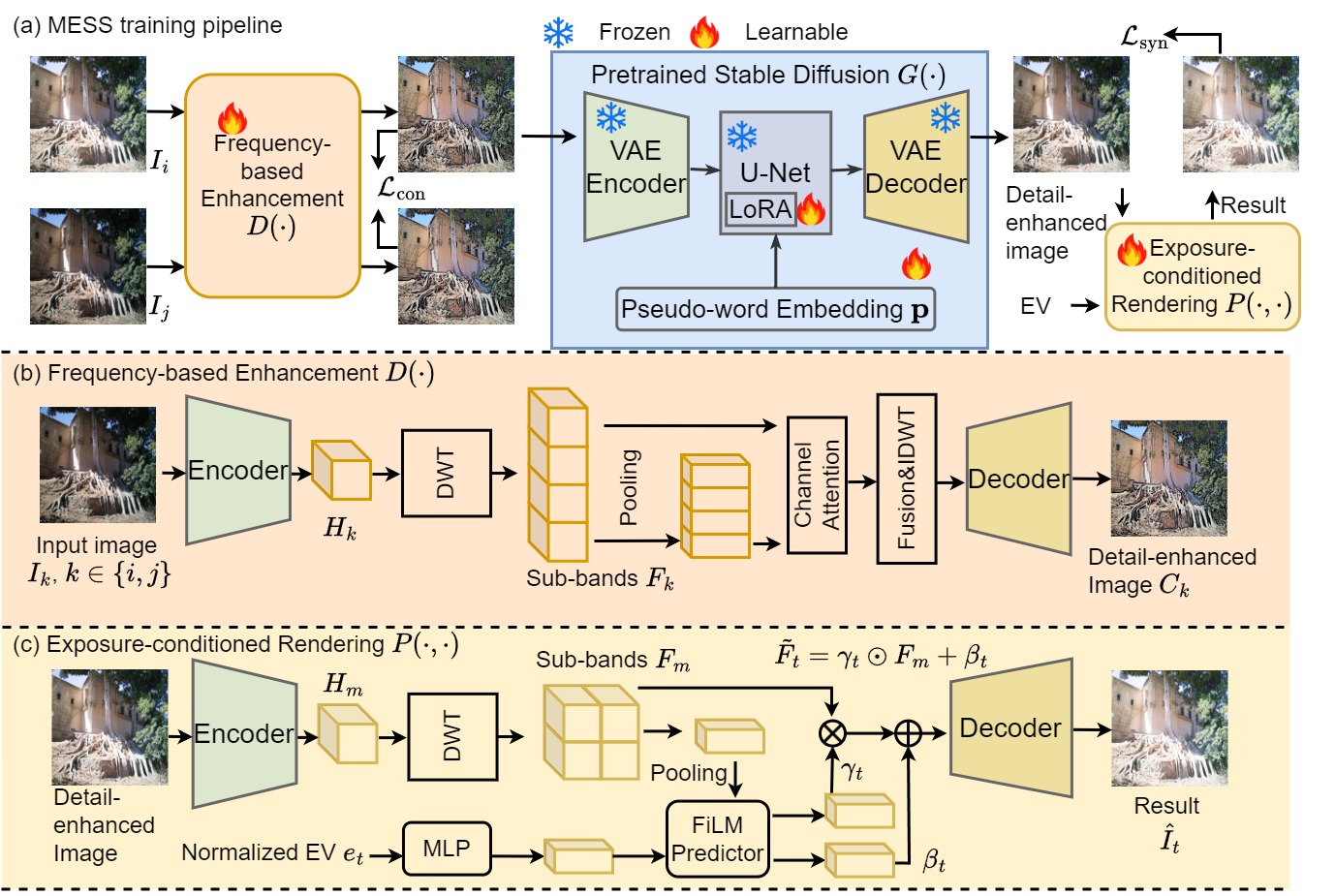}
    \caption{\textbf{MESS Training Framework.} (a) The training pipeline: Given a single RGB image and a target exposure value, MESS generates the corresponding exposure-adjusted image through three modules: frequency-based enhancement, diffusion-based synthesis, and exposure-conditioned rendering. (b) Structure of the frequency-based enhancement module. (c) Structure of the exposure-conditioned rendering module. 
    }
    \label{fig:framework}
    \vspace{-4mm}
\end{figure}

As shown in Fig.~\ref{fig:framework}(a), MESS consists of three sequentially connected modules.
The \textit{frequency-based enhancement module} \(D(\cdot)\) extracts a style-robust scene representation \(C_i = D(I_i)\).
The \textit{diffusion-based detail enhancement module} \(G(\cdot)\) enriches \(C_i\) with plausible structural and textural details, producing an intermediate image \(I_m = G(C_i)\).
The \textit{exposure-conditioned rendering module} \(P(\cdot)\) injects the normalized target exposure value \(e_t\) into \(I_m\) and generates the final exposure-aware RGB output \(\hat{I}_t = P(I_m, e_t)\).
Once the MESS model is trained, it is then applied to high-quality in-the-wild RGB images to synthesize MES. As shown in the bottom right of Fig.~\ref{fig:method_cmp},
for each RGB image, one synthesized exposure variant is sampled as the degraded input, while the original RGB image serves as the target.
The resulting pairs can be used to train a lightweight downstream enhancement network.

\subsection{Frequency-based Enhancement}
\label{subsec:frequency}
\vspace{-2mm}

To reduce rendering-style variations caused by camera ISP pipelines while preserving shared scene content, MESS uses a frequency-based enhancement module \(D(\cdot)\), as illustrated in Fig.~\ref{fig:framework}(b). This design is motivated by the fact that ISP-induced appearance variations, e.g., exposure and contrast changes, are mainly captured by low-frequency responses, whereas high-frequency components preserve stable structures and textures~\cite{li2023uhdfour,huang2022FECNet,wang2023fourllie}.
Given an input image \(I_k\), \(k \in \{i,j\}\), this module first encodes it into feature maps \(H_k\), then decomposes the features into different frequency bands, enhances them, and finally decodes them back to the image domain.

Many frequency-based enhancement methods employ Fast Fourier Transform (FFT) for the feature decomposition. However, FFT lacks spatial localization capability and is limited in modeling local variations~\cite{wang2023fourllie, li2023uhdfour, huang2022FECNet}.
We instead adopt Discrete Wavelet Transform (DWT), in particular the Haar wavelet transform due to its simplicity and reduced coefficient correlation~\cite{porwik2004haar,stankovic2003haar,batziou2023haarunet,zhang2025cwnet}, to perform the decomposition. Haar DWT is applied to each channel of \(H_k\), decomposing it into one low-frequency sub-band and three high-frequency sub-bands, collectively denoted as \(F_k\).
To adaptively fuse the wavelet sub-bands, we apply average pooling and max pooling to \(F_k\) and feed the pooled descriptors into a channel attention block.
The attention-enhanced representation is computed as \(\tilde{F}_k = A_k \odot F_k\), where \(A_k\) denotes the learned channel weights and \(\odot\) denotes channel-wise multiplication. The Inverse DWT (IDWT) is then applied to the processed features, which are finally decoded as the enhanced image representation \(C_k\).

During training, the style-varied renderings \(I_i\) and \(I_j\) are generated from the same RAW image.
We impose the representation consistency loss in \S\ref{subsec:objectives} on \(C_i = D(I_i)\) and \(C_j = D(I_j)\), encouraging \(D(\cdot)\) to produce style-invariant and content-consistent features.

\subsection{Diffusion-based Detail Enhancement}
\label{subsec:diffusion}
\vspace{-2mm}
After frequency-based enhancement, the resulting representation \(C_i\) is passed to a diffusion-based generator \(G\) to synthesize the missing details due to dynamic-range clipping or compression. The intermediate output is \(I_m = G(C_i)\).
We adopt the accelerated Stable Diffusion backbone (SD-Turbo)~\cite{sauer2025SD-Turbo} to perform this task for its strong generative prior and high efficiency.
The diffusion module \(G\) consists of a frozen VAE encoder, a LoRA-adapted U-Net denoising function \(\mathcal{G}_{\theta}\), and a frozen VAE decoder.
Since SD-Turbo is originally a text-to-image model, text conditioning needs to be provided.
Inspired by previous works~\cite{gal2023textual,morelli2023ladivton,lin2024fti4cir,li2025hierarchy}, we introduce a learnable shared pseudo-word embedding \(\mathbf{p}\) to adapt the diffusion prior to the target synthesis task.

Given \(C_i\), we first obtain the latent code \(z = \mathrm{Enc}_{\mathrm{VAE}}(C_i)\) using the frozen VAE encoder.
We then perform denoising as \(z' = \mathcal{G}_{\theta}(z, \mathbf{p})\), where \(\mathcal{G}_{\theta}\) is parameterized by the LoRA-adapted U-Net and \(\mathbf{p}\) is the learnable shared pseudo-word embedding.
The frozen VAE decoder maps \(z'\) to the diffusion output \(I_m = \mathrm{Dec}_{\mathrm{VAE}}(z')\), that is, \(G(C_i) = \mathrm{Dec}_{\mathrm{VAE}}(\mathcal{G}_{\theta}(\mathrm{Enc}_{\mathrm{VAE}}(C_i), \mathbf{p})).\) Following previous diffusion-based super-resolution works~\cite{wu2024osediff,wu2024seesr,sun2025pisasr}, we freeze the VAE and part of the U-Net, optimizing only the LoRA layers and pseudo-word embedding \(\mathbf{p}\).

\subsection{Exposure-conditioned Rendering}
\label{subsec:rendering}

As shown in Fig.~\ref{fig:framework}(c), the exposure-conditioned renderer \(P(\cdot)\) receives the diffusion-enhanced image \(I_m\) and normalized target exposure value \(e_t\) to output the final RGB image \(\hat{I}_t = P(I_m, e_t)\).
It aims to adjust exposure-dependent appearance, including brightness, contrast, and global tone, while preserving the structure and textures recovered by the diffusion module.

Since such appearance changes are mainly reflected in low-frequency components, we adopt the DWT-based modulation design in \S\ref{subsec:frequency}.
Specifically, the renderer first extracts feature maps \(H_m\) from \(I_m\) and applies Haar DWT to obtain wavelet sub-band features \(F_m\).
For each sub-band, global average pooling and max pooling are used to obtain compact content descriptors.
The normalized target exposure value \(e_t \in [0,1]\) is encoded by a lightweight MLP into an exposure embedding.
The content descriptors and exposure embedding are concatenated and fed into a FiLM~\cite{Perez2018FiLM} predictor, which outputs channel-wise affine parameters \(\gamma_t\) and \(\beta_t\).
The exposure-conditioned sub-band features are computed as \(\tilde{F}_t = \gamma_t \odot F_m + \beta_t\), where \(\odot\) denotes channel-wise multiplication, followed by IDWT and decoding to produce \(\hat{I}_t\).
Our formulation provides an efficient exposure-control mechanism: DWT decomposes illumination- and contrast-related responses into multi-resolution sub-bands, while exposure-conditioned affine modulation enables continuous rendering control with respect to \(e_t\).

\subsection{Training Loss}
\label{subsec:objectives}

MESS is trained with two objectives: representation consistency and synthesis reconstruction.
The representation consistency loss is applied to the style-varied renderings \(I_i\) and \(I_j\).
Since they share the same scene content, the enhanced representations \(C_i = D(I_i)\) and \(C_j = D(I_j)\) are encouraged to be close, yielding \(\mathcal{L}_{\mathrm{con}} = \|C_i - C_j\|_1\).
The synthesis reconstruction loss supervises the final synthesized image \(\hat{I}_t\) with the RAW-rendered reference \(I_t^{\star}\) under the same target exposure, which is defined as \(\mathcal{L}_{\mathrm{syn}} = \|\hat{I}_t - I_t^{\star}\|_2^2\).

The final training loss is \(\mathcal{L}_{\mathrm{total}} = \lambda \mathcal{L}_{\mathrm{con}} + \mathcal{L}_{\mathrm{syn}}\), where \(\lambda\) balances the two objectives.

\section{Experiments}
\subsection{Experimental Setup}

We evaluate MESS as a training data generator for generalizable image illumination enhancement.
MESS is first trained with RAW-guided supervision to learn camera-consistent exposure variations.
After training, we apply it to high-quality in-the-wild RGB images to synthesize paired training data for downstream image enhancement tasks (i.e., LLIE, EC and BIE).
Details of RAW datasets, RGB image collection, and downstream paired training data generation are provided in \textbf{Appendix B}.

\textbf{Implementation Details.}
MESS is trained on random $512\times512$ crops for 200K iterations using AdamW
($\beta_1=0.9$, $\beta_2=0.999$) on 8 NVIDIA A100 GPUs.
The total batch size is 64.
The learning rate is initialized as $1\times10^{-4}$ and decayed to $1\times10^{-6}$ with a cosine schedule.
The target EV is uniformly sampled from $[-3,3]$ EV.
The weight \(\lambda\) in the final loss is set to 1.
For downstream image enhancement models, training images are cropped to $512\times512$.
We use AdamW with an initial learning rate of $2\times10^{-4}$, batch size 64, cosine learning rate decay, and an $\ell_1$ reconstruction loss.
Each downstream model is trained for 300K iterations.
Unless otherwise specified, architecture, resolution, optimizer, batch size, schedule, and training iterations are kept fixed across data sources.

\textbf{Evaluation Protocol.}
To comprehensively evaluate the effectiveness of MESS,
we select representative backbone models from each downstream task, and train them on MESS-generated data.
Specifically, we employ FourLLIE~\cite{wang2023fourllie}, RetinexFormer~\cite{Cai2023Retinexformer}, and Multinex~\cite{multinex2026} from LLIE, ExSlot~\cite{jung2025exposure_slot} and CSEC~\cite{Li2024CSEC}from EC, and RAVE~\cite{gaintseva2024RAVE} from BIE.
We then evaluate these models on five commonly used test sets for different image enhancement tasks, including UHDLL~\cite{li2023uhdfour}, LSRW~\cite{Hai2023R2RNet}, SICE~\cite{Cai2018SICE}, MSEC~\cite{Afifi2021Learning}, and BAID~\cite{Lv2022BacklitNet}, to evaluate their generalization performance.

For comparison, we also train these models on three types of datasets to demonstrate the superiority of our MESS-generated data. (1) In-domain datasets, i.e., the associated training sets of the six employed test sets.
(2) Datasets synthesized by RGB-based methods, including Gamma correction~\cite{Lore2017LLNet}, UEC-based synthesis~\cite{Cui2024UEC}, and CSNorm-Fourier~\cite{Yao2023CSNorm}. (3) Datasets synthesized by  RAW-based methods, including MSEC~\cite{Afifi2021Learning} and RCD~\cite{Cui2024UEC}. Note that MSEC is also an in-domain dataset in our experiment.

We report PSNR, SSIM and {LPIPS \cite{zhang2018lpips}} as full-reference (FR) metrics, and ARNIQA~\cite{Agnolucci2024ARNIQA}, MUSIQ~\cite{Ke2021MUSIQ}, MANIQA~\cite{Yang2022MANIQA}, and TOPIQ~\cite{Chen2024TOPIQ} as no-reference (NR) metrics.

\begin{table*}[t]
\centering
\caption{The results of full-reference scores using FourLLIE backbone~\cite{wang2023fourllie}. The model trained on each data source are evaluated on all test sets without fine-tuning. Values in \textcolor{red}{red} and \textcolor{blue}{blue} indicate the best and second-best results, respectively.}
\label{tab:cross_dataset_friqa}
\small
\setlength{\tabcolsep}{3.5pt}
\begin{tabular}{l|c|c|cccccc}
\hline
Training Data & Type & Metric & UHDLL & LSRW & SICE & MSEC & BAID & Average\\
\hline

UHDLL~\cite{li2023uhdfour} & Real & PSNR $\uparrow$ & \textcolor{red}{23.694} & 19.357 & 16.152 & 18.728 & 20.461 & 19.678\\
LSRW~\cite{Hai2023R2RNet} & Real & PSNR $\uparrow$ & 21.383 & \textcolor{red}{20.849} & 17.246 & 18.964 & 19.875 & 19.663\\
SICE~\cite{Cai2018SICE} & Real & PSNR $\uparrow$ & 18.947 & 18.061 & \textcolor{red}{22.385} & 20.329 & 20.103 & 19.965\\
MSEC~\cite{Afifi2021Learning} & Raw-Syn & PSNR $\uparrow$ & 19.348 & 18.214 & 20.367 & \textcolor{red}{22.116} & 20.352 & \textcolor{blue}{20.079}\\
BAID~\cite{Lv2022BacklitNet} & Real & PSNR $\uparrow$ & 17.684 & 16.935 & 18.276 & 18.742 & \textcolor{blue}{21.674} & 18.662\\
RCD~\cite{Cui2024UEC} & Raw-Syn & PSNR $\uparrow$ & 19.702 & 18.242 & 17.478 & 19.924 & 20.159 & 19.101\\

Gamma~\cite{Lore2017LLNet} & RGB-Syn & PSNR $\uparrow$ & 16.845 & 14.692 & 17.248 & 17.681 & 18.136 & 16.920\\
CSNorm-Fourier~\cite{Yao2023CSNorm} & RGB-Syn & PSNR $\uparrow$ & 18.079 & 16.456 & 18.843 & 19.224 & 19.391 & 18.399\\
UEC-based~\cite{Cui2024UEC} & RGB-Syn & PSNR $\uparrow$ & 19.137 & 17.841 & 20.228 & 20.553 & 20.467 & 19.645\\
MESS-generated & RGB-Syn & PSNR $\uparrow$ & \textcolor{blue}{23.416} & \textcolor{blue}{20.514} & \textcolor{blue}{21.769} & \textcolor{blue}{21.982} & \textcolor{red}{21.935} & \textcolor{red}{21.923}\\
\hline
\hline

UHDLL~\cite{li2023uhdfour} & Real & SSIM $\uparrow$ & \textcolor{red}{0.914} & 0.701 & 0.762 & 0.819 & 0.871 & 0.813\\
LSRW~\cite{Hai2023R2RNet} & Real & SSIM $\uparrow$ & 0.897 & \textcolor{red}{0.731} & 0.786 & 0.814 & 0.853 & \textcolor{blue}{0.816}\\
SICE~\cite{Cai2018SICE} & Real & SSIM $\uparrow$ & 0.825 & 0.689 & \textcolor{red}{0.864} & 0.842 & 0.858 & 0.816\\
MSEC~\cite{Afifi2021Learning} & Raw-Syn & SSIM $\uparrow$ & 0.832 & 0.697 & 0.816 & \textcolor{red}{0.881} & 0.855 & 0.816\\
BAID~\cite{Lv2022BacklitNet} & Real & SSIM $\uparrow$ & 0.782 & 0.641 & 0.764 & 0.792 & \textcolor{red}{0.893} & 0.774\\
RCD~\cite{Cui2024UEC} & Raw-Syn & SSIM $\uparrow$ & 0.836 & 0.694 & 0.781 & 0.822 & 0.856 & 0.798\\

Gamma~\cite{Lore2017LLNet} & RGB-Syn & SSIM $\uparrow$ & 0.745 & 0.568 & 0.712 & 0.729 & 0.763 & 0.703\\
CSNorm-Fourier~\cite{Yao2023CSNorm} & RGB-Syn & SSIM $\uparrow$ & 0.798 & 0.614 & 0.767 & 0.781 & 0.816 & 0.755\\
UEC-based~\cite{Cui2024UEC} & RGB-Syn & SSIM $\uparrow$ & 0.833 & 0.659 & 0.804 & 0.827 & 0.852 & 0.795\\
MESS-generated & RGB-Syn & SSIM $\uparrow$ & \textcolor{blue}{0.910} & \textcolor{blue}{0.718} & \textcolor{blue}{0.851} & \textcolor{blue}{0.869} & \textcolor{blue}{0.884} & \textcolor{red}{0.846}\\
\hline
\hline

UHDLL~\cite{li2023uhdfour} & Real & LPIPS $\downarrow$ & \textcolor{red}{0.084} & 0.177 & 0.202 & 0.158 & \textcolor{blue}{0.091} & \textcolor{blue}{0.142}\\
LSRW~\cite{Hai2023R2RNet} & Real & LPIPS $\downarrow$ & 0.113 & \textcolor{red}{0.149} & 0.196 & 0.164 & 0.125 & 0.149\\
SICE~\cite{Cai2018SICE} & Real & LPIPS $\downarrow$ & 0.157 & 0.181 & \textcolor{blue}{0.145} & 0.139 & 0.123 & 0.149\\
MSEC~\cite{Afifi2021Learning} & Raw-Syn & LPIPS $\downarrow$ & 0.142 & 0.178 & 0.166 & \textcolor{red}{0.131} & 0.117 & 0.147\\
BAID~\cite{Lv2022BacklitNet} & Real & LPIPS $\downarrow$ & 0.184 & 0.226 & 0.206 & 0.187 & 0.105 & 0.182\\
RCD~\cite{Cui2024UEC} & Raw-Syn & LPIPS $\downarrow$ & 0.135 & 0.180 & 0.183 & 0.151 & 0.121 & 0.154\\

Gamma~\cite{Lore2017LLNet} & RGB-Syn & LPIPS $\downarrow$ & 0.235 & 0.292 & 0.258 & 0.231 & 0.206 & 0.244\\
CSNorm-Fourier~\cite{Yao2023CSNorm} & RGB-Syn & LPIPS $\downarrow$ & 0.198 & 0.244 & 0.217 & 0.203 & 0.169 & 0.206\\
UEC-based~\cite{Cui2024UEC} & RGB-Syn & LPIPS $\downarrow$ & 0.153 & 0.208 & 0.174 & 0.179 & 0.132 & 0.169\\
MESS-generated & RGB-Syn & LPIPS $\downarrow$ & \textcolor{blue}{0.091} & \textcolor{blue}{0.156} & \textcolor{red}{0.138} & \textcolor{blue}{0.134} & \textcolor{red}{0.087} & \textcolor{red}{0.121}\\
\hline
\end{tabular}
\vspace{-3mm}
\end{table*}

\begin{table*}[t]
\centering
\vspace{-3mm}
\caption{The results of no-reference scores using FourLLIE backbone~\cite{wang2023fourllie}. The model trained on each data source are evaluated on all test sets without fine-tuning. Values in \textcolor{red}{red} and \textcolor{blue}{blue} indicate the best and second-best results, respectively.}
\label{tab:cross_dataset_nriqa}
\small
\setlength{\tabcolsep}{3.5pt}
\begin{tabular}{l|c|c|cccccc}
\hline
Training Data & Type & Metric & UHDLL & LSRW & SICE & MSEC & BAID & Average\\
\hline

UHDLL~\cite{li2023uhdfour} & Real & ARNIQA $\uparrow$ & \textcolor{blue}{0.681} & 0.650 & 0.694 & 0.693 & 0.703 & 0.684\\
LSRW~\cite{Hai2023R2RNet} & Real & ARNIQA $\uparrow$ & 0.662 & \textcolor{red}{0.671} & 0.688 & 0.691 & 0.699 & 0.682\\
SICE~\cite{Cai2018SICE} & Real & ARNIQA $\uparrow$ & 0.666 & 0.630 & \textcolor{blue}{0.720} & 0.711 & 0.693 & 0.684\\
MSEC~\cite{Afifi2021Learning} & Raw-Syn & ARNIQA $\uparrow$ & 0.663 & 0.648 & 0.719 & \textcolor{blue}{0.720} & 0.708 & \textcolor{blue}{0.692}\\
BAID~\cite{Lv2022BacklitNet} & Real & ARNIQA $\uparrow$ & 0.628 & 0.611 & 0.652 & 0.667 & \textcolor{blue}{0.727} & 0.657\\
RCD~\cite{Cui2024UEC} & Raw-Syn & ARNIQA $\uparrow$ & 0.663 & 0.640 & 0.693 & 0.697 & 0.699 & 0.678\\

Gamma~\cite{Lore2017LLNet} & RGB-Syn & ARNIQA $\uparrow$ & 0.625 & 0.604 & 0.662 & 0.673 & 0.685 & 0.650\\
CSNorm-Fourier~\cite{Yao2023CSNorm} & RGB-Syn & ARNIQA $\uparrow$ & 0.631 & 0.611 & 0.676 & 0.681 & 0.695 & 0.659\\
UEC-based~\cite{Cui2024UEC} & RGB-Syn & ARNIQA $\uparrow$ & 0.655 & 0.639 & 0.694 & 0.700 & 0.717 & 0.681\\
MESS-generated & RGB-Syn & ARNIQA $\uparrow$ & \textcolor{red}{0.698} & \textcolor{blue}{0.661} & \textcolor{red}{0.726} & \textcolor{red}{0.721} & \textcolor{red}{0.738} & \textcolor{red}{0.709}\\
\hline
\hline

UHDLL~\cite{li2023uhdfour} & Real & MUSIQ $\uparrow$ & \textcolor{blue}{57.904} & 50.639 & 50.855 & 52.439 & 54.013 & 53.170\\
LSRW~\cite{Hai2023R2RNet} & Real & MUSIQ $\uparrow$ & 56.741 & \textcolor{red}{52.180} & 51.263 & 53.064 & 54.291 & 53.508\\
SICE~\cite{Cai2018SICE} & Real & MUSIQ $\uparrow$ & 55.483 & 50.421 & \textcolor{blue}{54.966} & 54.134 & 53.757 & 53.752\\
MSEC~\cite{Afifi2021Learning} & Raw-Syn & MUSIQ $\uparrow$ & 56.225 & 50.762 & 52.685 & \textcolor{blue}{55.065} & 54.863 & \textcolor{blue}{53.920}\\
BAID~\cite{Lv2022BacklitNet} & Real & MUSIQ $\uparrow$ & 52.876 & 49.214 & 50.936 & 51.872 & \textcolor{blue}{56.218} & 52.223\\
RCD~\cite{Cui2024UEC} & Raw-Syn & MUSIQ $\uparrow$ & 56.191 & 50.623 & 51.706 & 53.255 & 54.001 & 53.155\\

Gamma~\cite{Lore2017LLNet} & RGB-Syn & MUSIQ $\uparrow$ & 53.218 & 48.973 & 51.495 & 52.320 & 53.273 & 51.856\\
CSNorm-Fourier~\cite{Yao2023CSNorm} & RGB-Syn & MUSIQ $\uparrow$ & 54.130 & 49.435 & 52.216 & 53.054 & 53.921 & 52.551\\
UEC-based~\cite{Cui2024UEC} & RGB-Syn & MUSIQ $\uparrow$ & 55.073 & 49.939 & 53.045 & 53.783 & 54.747 & 53.317\\
MESS-generated & RGB-Syn & MUSIQ $\uparrow$ & \textcolor{red}{58.136} & \textcolor{blue}{52.087} & \textcolor{red}{55.672} & \textcolor{red}{55.214} & \textcolor{red}{56.442} & \textcolor{red}{55.510}\\
\hline
\hline

UHDLL~\cite{li2023uhdfour} & Real & MANIQA $\uparrow$ & 0.463 & 0.418 & 0.438 & 0.454 & 0.439 & 0.442\\
LSRW~\cite{Hai2023R2RNet} & Real & MANIQA $\uparrow$ & \textcolor{blue}{0.466} & \textcolor{blue}{0.429} & 0.446 & 0.445 & 0.433 & 0.444\\
SICE~\cite{Cai2018SICE} & Real & MANIQA $\uparrow$ & 0.452 & 0.408 & \textcolor{blue}{0.467} & 0.461 & 0.430 & 0.444\\
MSEC~\cite{Afifi2021Learning} & Raw-Syn & MANIQA $\uparrow$ & 0.451 & 0.412 & 0.462 & \textcolor{blue}{0.476} & 0.449 & \textcolor{blue}{0.450}\\
BAID~\cite{Lv2022BacklitNet} & Real & MANIQA $\uparrow$ & 0.421 & 0.397 & 0.426 & 0.432 & \textcolor{blue}{0.457} & 0.427\\
RCD~\cite{Cui2024UEC} & Raw-Syn & MANIQA $\uparrow$ & 0.454 & 0.411 & 0.441 & 0.451 & 0.436 & 0.439\\

Gamma~\cite{Lore2017LLNet} & RGB-Syn & MANIQA $\uparrow$ & 0.418 & 0.394 & 0.425 & 0.421 & 0.414 & 0.414\\
CSNorm-Fourier~\cite{Yao2023CSNorm} & RGB-Syn & MANIQA $\uparrow$ & 0.436 & 0.402 & 0.437 & 0.440 & 0.424 & 0.428\\
UEC-based~\cite{Cui2024UEC} & RGB-Syn & MANIQA $\uparrow$ & 0.448 & 0.422 & 0.448 & 0.451 & 0.434 & 0.441\\
MESS-generated & RGB-Syn & MANIQA $\uparrow$ & \textcolor{red}{0.478} & \textcolor{red}{0.439} & \textcolor{red}{0.473} & \textcolor{red}{0.482} & \textcolor{red}{0.465} & \textcolor{red}{0.467}\\
\hline
\end{tabular}
\vspace{-5mm}
\end{table*}

\subsection{Main Results}

Due to page limit, in the main paper we can only report the results using the backbone model FourLLIE~\cite{wang2023fourllie}. The results using the other backbone models are provided in \textbf{Appendix C}.

\textbf{Quantitative Results}.
Tables~\ref{tab:cross_dataset_friqa} and~\ref{tab:cross_dataset_nriqa} report full-reference and no-reference results, respectively, from which we see that the training data are critical for model generalization. First, models trained on task-specific paired datasets perform competitively on their corresponding tasks but generalize poorly to other enhancement tasks.
For example, UHDLL and LSRW mainly cover underexposure, while SICE~\cite{Cai2018SICE}, MSEC~\cite{Afifi2021Learning} and BAID~\cite{Lv2022BacklitNet} focus on exposure correction and backlit enhancement, respectively.
Second, existing RGB-based synthesis improves scalability but over-simplifies degradation assumptions.
Gamma correction~\cite{Lore2017LLNet} mainly models global intensity changes, while UEC-based~\cite{Cui2024UEC} and CSNorm-Fourier~\cite{Yao2023CSNorm} lack camera-consistent exposure variations.
Third, RAW-based synthesis improves a little the generalization but remains limited in scalability.
In contrast, MESS-generated data consistently improves both fidelity and perceptual quality across LLIE, EC, and BIE benchmarks, boosting significantly the generalization performance of the lightweight FourLLIE backbone model, which is originally designed for the LLIE task only.

\begin{figure}[t]
\centering
\begin{tabular*}{\linewidth}{@{}p{0.2\linewidth}@{}p{0.2\linewidth}@{}p{0.2\linewidth}@{}p{0.2\linewidth}@{}p{0.2\linewidth}@{}}
\centering\small Input &
\centering\small UHDLL-trained &
\centering\small LSRW-trained &
\centering\small MESS-trained &
\centering\small GT
\end{tabular*}
\vspace{1pt}
\includegraphics[width=1\textwidth]{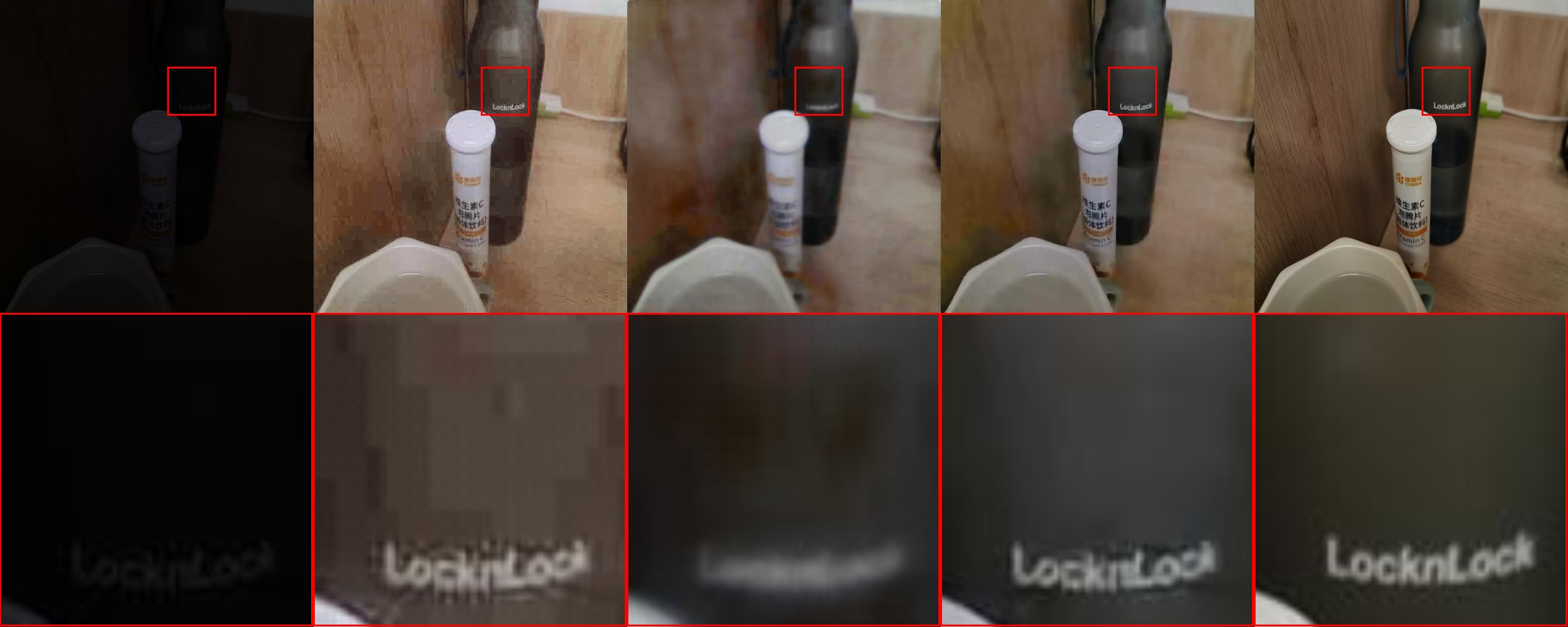}
\vspace{-3mm}
\caption{
Qualitative comparison of the FourLLIE models trained on UHDLL, LSRW and our MESS generated datasets. The test image is from the LSRW dataset. }
\label{fig:qualitative_llie}
\vspace{-3mm}
\end{figure}

\begin{figure}[t]
\centering
\begin{tabular*}{\linewidth}{@{}p{0.2\linewidth}@{}p{0.2\linewidth}@{}p{0.2\linewidth}@{}p{0.2\linewidth}@{}p{0.2\linewidth}@{}}
\centering\small Input &
\centering\small MSEC-trained &
\centering\small UEC-trained &
\centering\small MESS-trained &
\centering\small GT
\end{tabular*}
\vspace{1pt}
\includegraphics[width=1\textwidth]{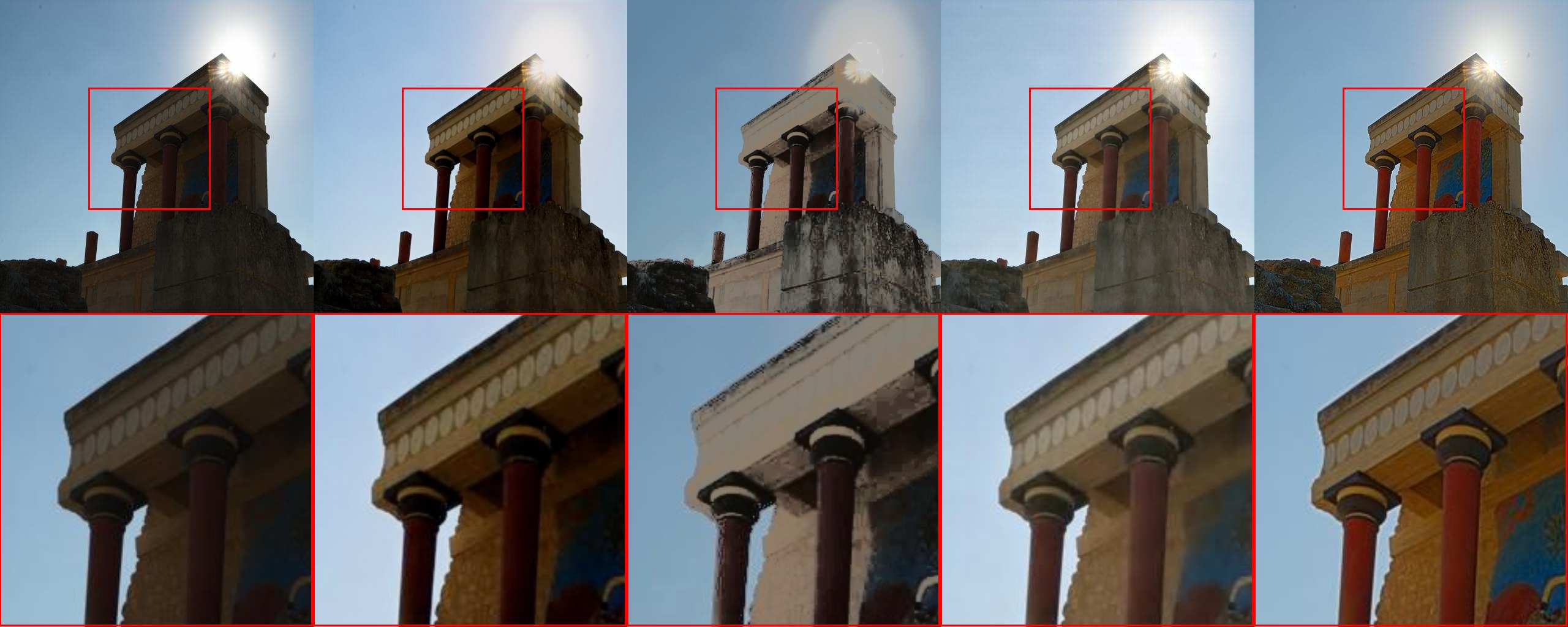}
\vspace{-3mm}
\caption{
Qualitative comparison of the FourLLIE models trained on MSEC, UEC and MESS generated datasets. The test image is from the SICE dataset.
}
\label{fig:qualitative_ec}
\vspace{-3mm}
\end{figure}

\textbf{Qualitative Results}.
Figure~\ref{fig:qualitative_llie} shows qualitative comparisons for the LLIE task under cross-dataset settings.
All models use the same architecture but are trained on different data sources, and are tested on LSRW.
Although UHDLL~\cite{li2023uhdfour} and LSRW~\cite{Hai2023R2RNet} are both real-captured LLIE datasets, models trained on them still exhibit visible artifacts when tested across datasets.
The UHDLL-trained model amplifies JPEG artifacts, while the LSRW-trained model, despite being trained and tested on LSRW, still produces blurred details.
In contrast, the model trained on MESS-generated data avoids these artifacts and preserves sharper structures.
Figure~\ref{fig:qualitative_ec} shows another example of cross-dataset test using the EC task.
All models use the same architecture but are trained on different data sources, and are tested on SICE.
The RAW-based MSEC~\cite{Afifi2021Learning} training data fail to provide effective enhancement in this case.
The RGB-based UEC~\cite{Cui2024UEC} training data improve exposure but lead to unnatural color appearance.
In contrast, MESS-generated data yield more natural exposure correction with better color fidelity and texture preservation.
More visual comparisons can be found in \textbf{Appendix D}.

\textbf{Ablation Study}.
We also perform ablation studies on the frequency-based enhancement module, diffusion-based detail enhancement module, and exposure-conditioned rendering module.
The results verify that each component contributes to the quality of synthesized data and the generalization ability of the downstream model. The details can be found in \textbf{Appendix E}.

\textbf{Effect of Scale of Training Data.}
We evaluate the effect of training data scale by randomly sampling three subsets of MESS, containing 5K, 10K, and 50K pairs, respectively, and comparing them with the full 500K dataset. All settings use the FourLLIE backbone~\cite{wang2023fourllie} and the same downstream training protocol described in the experimental setup. MESS-5K is approximately comparable in size to LSRW~\cite{Hai2023R2RNet} and SICE~\cite{Cai2018SICE}. Table~\ref{tab:cr_scale_summary} reports macro-average results over UHDLL~\cite{li2023uhdfour}, LSRW~\cite{Hai2023R2RNet}, SICE~\cite{Cai2018SICE}, MSEC~\cite{Afifi2021Learning}, and BAID~\cite{Lv2022BacklitNet}. MESS-5K achieves better PSNR, SSIM and LPIPS scores than both the real-data baselines, possibly benefiting from its high-quality ground truth. Further gains with larger training sets highlight the benefits of both data quality and scale.

\begin{table}[t]
\centering
\small
\caption{Dataset-scale comparison using FourLLIE. Scores are macro-averaged over the five test sets. MESS-5K is approximately comparable in size to LSRW and SICE. Values in \textcolor{red}{red} and \textcolor{blue}{blue} indicate the best and second-best results, respectively.}
\label{tab:cr_scale_summary}
\resizebox{\linewidth}{!}{%
\begin{tabular}{lrrrrrr}
\toprule
Training data & PSNR $\uparrow$ & SSIM $\uparrow$ & LPIPS $\downarrow$ & ARNIQA $\uparrow$ & MANIQA $\uparrow$ & MUSIQ $\uparrow$ \\
\midrule
LSRW & 19.663 & 0.816 & 0.149 & 0.682 & 0.444 & 53.508 \\
SICE & 19.965 & 0.816 & 0.149 & 0.684 & 0.444 & 53.752 \\
MESS-5K & 20.220 & 0.822 & 0.143 & 0.686 & 0.451 & 54.050 \\
MESS-10K & 20.590 & 0.830 & 0.137 & 0.692 & 0.456 & 54.410 \\
MESS-50K & \textcolor{blue}{21.199} & \textcolor{blue}{0.839} & \textcolor{blue}{0.129} & \textcolor{blue}{0.701} & \textcolor{blue}{0.463} & \textcolor{blue}{54.988} \\
MESS-500K & \textcolor{red}{21.923} & \textcolor{red}{0.846} & \textcolor{red}{0.121} & \textcolor{red}{0.709} & \textcolor{red}{0.467} & \textcolor{red}{55.510} \\
\bottomrule
\end{tabular}}
\end{table}

\textbf{RAW-only versus RGB Expansion.}
We compare RAW-only training data with matched-size and full-size MESS-RGB datasets under identical downstream architectures and training protocols, using FourLLIE~\cite{wang2023fourllie} as the backbone. The RAW-only dataset contains 39,642 scenes and training pairs, with exposure-perturbed inputs and MEF-generated targets obtained using an approach similar to UNICE~\cite{Cui2025UNICE}. The matched-size MESS-RGB dataset contains the same number of RGB scenes and pairs, using curated high-quality RGB images as targets. Table~\ref{tab:cr_raw_rgb} shows that MESS-RGB dataset improves in all three macro-average full-reference metrics at matched size. The advantages reflect the better target quality of MESS-RGB data over RAW-only MEF targets. Further gains with the full 500K set suggest an additional benefit from increased data scale.

\begin{table}[t]
\centering
\small
\caption{RAW-only and MESS-RGB comparison. Results are macro-averaged over UHDLL, LSRW, SICE, MSEC, and BAID. Values in \textcolor{red}{red} and \textcolor{blue}{blue} indicate the best and second-best results, respectively.}
\label{tab:cr_raw_rgb}
\resizebox{\linewidth}{!}{%
\begin{tabular}{lrrrrr}
\toprule
Training data & Unique scenes & Pairs & PSNR $\uparrow$ & SSIM $\uparrow$ & LPIPS $\downarrow$ \\
\midrule
RAW-only & 39,642 & 39,642 & 20.764 & 0.829 & 0.139 \\
MESS-RGB, matched size & 39,642 & 39,642 & \textcolor{blue}{21.083} & \textcolor{blue}{0.837} & \textcolor{blue}{0.131} \\
MESS-RGB, full set & 500,000 & 500,000 & \textcolor{red}{21.923} & \textcolor{red}{0.846} & \textcolor{red}{0.121} \\
\bottomrule
\end{tabular}}
\end{table}

\begin{figure}[!ht]
\centering
\setlength{\tabcolsep}{2pt}
\begin{tabular}{ccccc}
\small Source RGB & \small $-3$ EV & \small $-1$ EV & \small $+1$ EV & \small $+3$ EV \\
\includegraphics[width=0.185\linewidth]{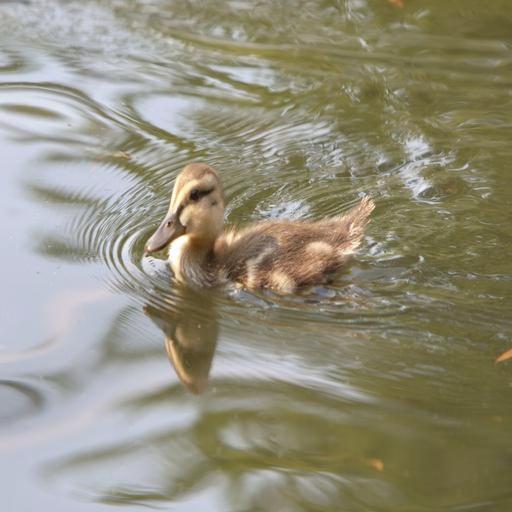} &
\includegraphics[width=0.185\linewidth]{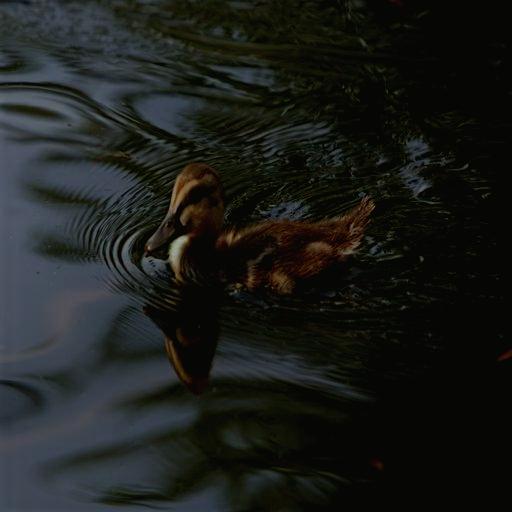} &
\includegraphics[width=0.185\linewidth]{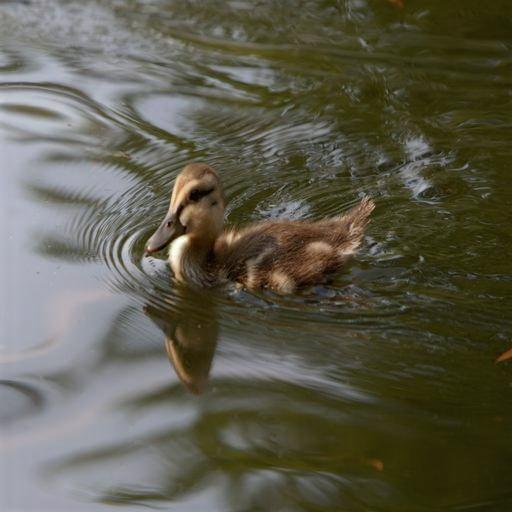} &
\includegraphics[width=0.185\linewidth]{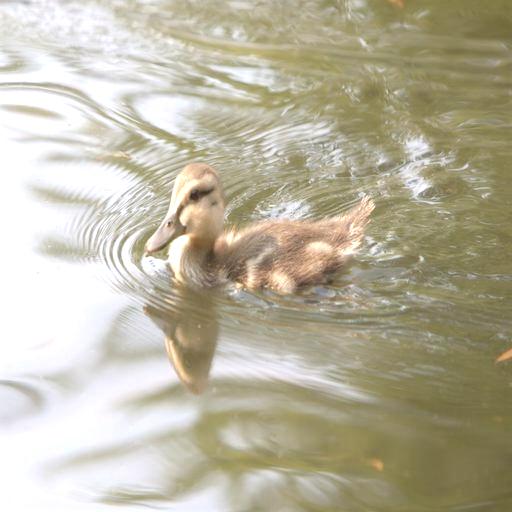} &
\includegraphics[width=0.185\linewidth]{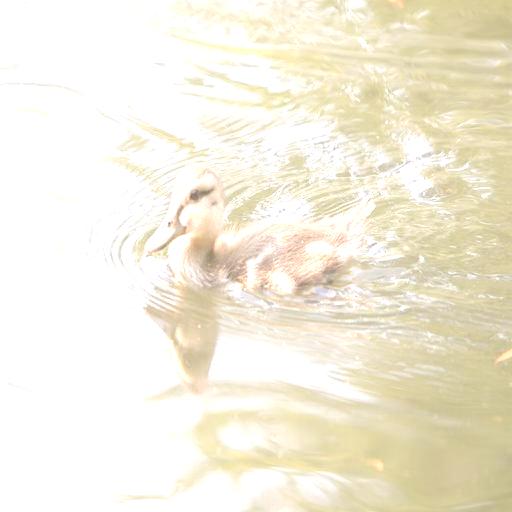} \\
\end{tabular}
\caption{Visualization of MESS synthesis. The source RGB image is followed by synthesized exposure variants from MESS at $-3$, $-1$, $+1$, and $+3$ EV, respectively.}
\label{fig:cr_mes_synthesis}
\end{figure}

\textbf{Visualization of MESS Synthesis Outputs.}
Figure~\ref{fig:cr_mes_synthesis} shows exposure-controlled synthesis from a single in-the-wild RGB image. Varying the target EV produces under- and overexposed variants of the same source image.


\section{Conclusion}

In this paper, we proposed MESS, a scalable data synthesis framework for generalizable image illumination enhancement. MESS integrated frequency-based enhancement, diffusion-based detail compensation, and exposure-conditioned rendering. By transferring RAW-derived exposure variations to large-scale in-the-wild RGB images, MESS combined RAW-rendering realism with RGB-data scalability, enabling realistic and scalable training data synthesis. Using MESS, we constructed 500K training samples from high-quality in-the-wild RGB images, and trained a lightweight model on this dataset, which demonstrated strong generalization performance across diverse real-world illumination enhancement benchmarks. It is anticipated that MESS can advance scalable data synthesis for robust image enhancement in the wild.

{\textbf{Limitations.} Despite the promising results, MESS has two main limitations. First, it currently relies on a fixed set of RAW datasets to derive exposure priors; expanding the RAW data pool could further improve the realism and coverage of the synthesized data. Second, MESS currently selects the best-rendered image based on several NR-IQA metrics; incorporating powerful Vision Language Models (VLMs) into this process could yield more robust data selection.}

\clearpage
{\small
\bibliographystyle{plainnat}
\bibliography{references}
}

\clearpage
\appendix

\begin{center}
\Large \textbf{Appendix}
\end{center}

\vspace{1em}

This appendix provides additional details and results to support the main paper. Sec.~\ref{sec:supp} presents implementation details, including the construction of RAW-guided training data and the network architecture of MESS. Sec.~\ref{sec:supp_rgb_pairs} describes how we use the trained MESS generator to construct the large-scale paired RGB training set. Sec.~\ref{sec:additional_quantitative_results} reports additional quantitative results with other downstream enhancement backbones. Sec.~\ref{sec:supp_qualitative} provides additional qualitative comparisons across low-light enhancement, exposure correction, and backlit enhancement benchmarks. Finally, Sec.~\ref{sec:supp_ablation} provides ablation studies that analyze the contribution of each component in MESS.

\section{Additional Implementation Details}
\label{sec:supp}

\subsection{Details of RAW-guided Training Data Construction}
\label{subsec:supp_raw_data}

This section provides more details for the data construction process described in Sec.~\ref{subsec:framework}. We collect high-bit RAW images from publicly available datasets, including AODRAW~\cite{li2024AODRaw}, FiveK~\cite{Bychkovsky2011FiveK}, HDRP~\cite{hasinoff2016HDRP}, PASCAL-RAW~\cite{omid2014pascalraw}, PPR10K~\cite{jie2021PPR10K}, RAISE~\cite{dang2015raise}, and RAW-NOD~\cite{morawski2022RAWNOD}. Images with severe artifacts, invalid metadata, calibration charts, or extremely poor visual quality are removed before rendering.
For each RAW image \(X\), we first render a seven-exposure sequence by adjusting the exposure value to
\[
    \{-3,-2,-1,0,+1,+2,+3\}\,\mathrm{EV}.
\]
This sequence is used to construct two types of rendered images. We first render exposure-specific references with a camera-consistent ISP setting. The target exposure value is randomly sampled and normalized to \(e_t \in [0,1]\), yielding the reference image \(I_t^{\star}\) for exposure supervision.

We then construct a style-varied rendering pool from the same RAW image.
In addition to directly rendered exposure variants, we apply multiple
multi-exposure fusion (MEF) \cite{mertens2007exposure,goshtasby2005fusion,li2020spdmef}
and tone mapping \cite{reinhard2002photographic,drago2003adaptive,durand2002fast}
algorithms to the seven-exposure sequence.
These operations produce images with diverse tone, contrast, brightness, saturation, and local detail.
During training, two images are randomly sampled from this pool to form \(I_i\) and \(I_j\).

After training MESS on RAW-guided triplets, we apply it to high-quality RGB images collected from in-the-wild datasets. For each RGB image, we sample one or more target exposure values and synthesize exposure-varied outputs. One synthesized image is randomly selected as the input, while the original RGB image is used as the target. This process produces paired data for downstream enhancement training.

The final dataset contains 39,642 groups after manual filtering. Each group contains synthesized exposure-varied images associated with the same high-quality target image. The dataset statistics are summarized in Tab.~\ref{tab:subdataset_stats}, and representative examples are shown in Fig.~\ref{fig:supp_raw_data_examples}.

\afterpage{%
\begin{table}[!t]
\centering
\begin{tabular}{l|cccc}
\toprule
Dataset   & AODRAW & PASCAL-RAW & RAISE   & FiveK  \\
\# Groups & 4396   & 3971       & 6992    & 3651   \\
\midrule
Dataset   & HDRP   & PPR10K     & RAW-NOD & Total  \\
\# Groups & 3155   & 9790       & 7687    & 39642  \\
\bottomrule
\end{tabular}
\caption{Statistics of the constructed training dataset after manual filtering. Each group contains synthesized exposure-varied images associated with the same high-quality target image.}
\label{tab:subdataset_stats}
\end{table}

\begin{figure}[t]
    \centering
    \begin{tabular}{@{}c@{\hspace{0.03\linewidth}}c@{}}
        \makebox[0.48\linewidth][c]{Style-varied rendering} &
        \makebox[0.48\linewidth][c]{Renderings with different EVs}
    \end{tabular}
    \includegraphics[width=1\linewidth]{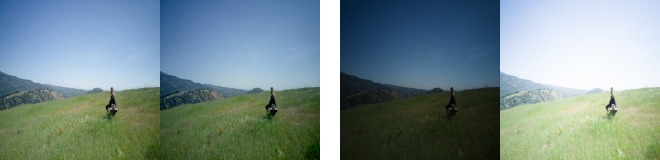}
    \caption{Representative examples of the constructed dataset.
    The style-varied renderings show diverse tones, contrast, and local details, while the exposure-varied renderings correspond to different EV settings of the same scene.}
    \label{fig:supp_raw_data_examples}
\end{figure}
}

\subsection{Network Architecture Details}
\label{supsec:network}

This section describes with more details the architecture of the proposed MESS network. As shown in Figure~\ref{fig:method_cmp} in the main paper,
MESS consists of three sequential modules: the frequency-based enhancement module \(D(\cdot)\), the diffusion-based detail enhancement module \(G(\cdot)\), and the exposure-conditioned rendering module \(P(\cdot)\).
All convolutional layers use reflection padding unless otherwise specified.
LeakyReLU with a negative slope of 0.2 is used in all convolutional blocks.

\subsubsection{Frequency-based Enhancement Module}
\label{supsec:network_frequency}

The frequency-based enhancement module \(D(\cdot)\) adopts an encoder--wavelet attention--decoder structure.
Given an RGB input image \(I_k \in \mathbb{R}^{3 \times H \times W}\), where \(k \in \{i,j\}\), the encoder \(E_D(\cdot)\) maps it to a 64-channel feature map \(H_k \in \mathbb{R}^{64 \times H \times W}\).

The encoder consists of one \(3 \times 3\) convolutional layer followed by four residual blocks.
The first convolution maps the RGB input to 64 channels.
Each residual block contains two \(3 \times 3\) convolutional layers with 64 channels and a residual skip connection.
The encoder does not use spatial downsampling.

A single-level Haar DWT is applied to \(H_k\).
For each feature channel, the transform produces one low-frequency sub-band and three high-frequency sub-bands, denoted by \(LL\), \(LH\), \(HL\), and \(HH\).
The four sub-bands are concatenated along the channel dimension, resulting in a wavelet feature \(F_k \in \mathbb{R}^{256 \times H/2 \times W/2}\).

A wavelet-domain channel attention block adaptively reweights the concatenated sub-band feature.
Specifically, global average pooling and global max pooling are applied to \(F_k\), producing two 256-dimensional channel descriptors.
The two descriptors are passed through a shared two-layer MLP with hidden dimension 64 and output dimension 256.
The outputs of the two branches are summed and activated by a sigmoid function to obtain the channel attention weights \(A_k\).
The attended wavelet feature is computed as \(\tilde{F}_k = A_k \odot F_k\), where \(A_k\) is broadcast along the spatial dimensions.

The attended wavelet feature is transformed back by inverse Haar DWT, producing \(\tilde{H}_k \in \mathbb{R}^{64 \times H \times W}\).
The decoder \(Q_D(\cdot)\) then maps \(\tilde{H}_k\) to the enhanced representation \(C_k\).
The decoder contains four residual blocks with 64 channels and a final \(3 \times 3\) convolutional layer that maps 64 channels to 3 channels.
A sigmoid activation is applied at the end to constrain the output to \([0,1]\).

\subsubsection{Diffusion-based Detail Enhancement Module}
\label{supsec:network_diffusion}

The diffusion-based detail enhancement module \(G(\cdot)\) is built on SD-Turbo and operates in the latent space.
It consists of a frozen VAE encoder, a LoRA-adapted U-Net, a learnable pseudo-word embedding, and a frozen VAE decoder.

Given the enhanced representation \(C_i\), we first normalize it from \([0,1]\) to \([-1,1]\) and feed it into the frozen VAE encoder to obtain the latent code \(z\).
The latent code is then processed by the SD-Turbo U-Net conditioned on a learnable pseudo-word embedding \(\mathbf{p}\).
The output latent \(z'\) is decoded by the frozen VAE decoder and rescaled to \([0,1]\), producing the intermediate image \(I_m\).

The pseudo-word embedding \(\mathbf{p}\) consists of a single learnable token and is shared across all training samples.
No hand-crafted text prompt is used.

LoRA adapters are inserted into the attention projection layers of the U-Net, including the query, key, value, and output projections in both self-attention and cross-attention blocks.
For a frozen projection matrix \(W\), the LoRA-adapted projection is defined as
\begin{equation}
    y = Wx + \frac{\alpha}{r}BAx,
\end{equation}
where \(A\) and \(B\) are low-rank trainable matrices.
We use LoRA rank \(r=8\) and scaling factor \(\alpha=8\).
Only the LoRA parameters and the pseudo-word embedding are learnable in this module, while the VAE encoder, VAE decoder, and original U-Net weights remain frozen.

\subsubsection{Exposure-conditioned Rendering Module}
\label{supsec:network_rendering}

The exposure-conditioned rendering module \(P(\cdot)\) contains a convolutional encoder, a wavelet-domain FiLM modulation block, and a decoder.
Given the diffusion-enhanced image \(I_m \in \mathbb{R}^{3 \times H \times W}\), the encoder \(E_P(\cdot)\) extracts a 64-channel feature map \(H_m \in \mathbb{R}^{64 \times H \times W}\).

The encoder \(E_P(\cdot)\) has the same structure as \(E_D(\cdot)\), consisting of one \(3 \times 3\) convolutional layer followed by four residual blocks.
The first convolution maps the RGB input to 64 channels, and each residual block contains two \(3 \times 3\) convolutional layers with a residual skip connection.

A single-level Haar DWT is applied to \(H_m\).
The resulting \(LL\), \(LH\), \(HL\), and \(HH\) sub-bands are concatenated along the channel dimension, producing a wavelet feature \(F_m \in \mathbb{R}^{256 \times H/2 \times W/2}\).

The scalar target exposure value \(e_t \in [0,1]\) is encoded by a two-layer MLP.
The exposure encoder maps the one-dimensional input to a 128-dimensional embedding through a 64-dimensional hidden layer, with a ReLU activation after the first linear layer.

To predict content-adaptive modulation parameters, global average pooling and global max pooling are applied to \(F_m\), producing two 256-dimensional descriptors.
These descriptors are concatenated with the 128-dimensional exposure embedding, resulting in a 640-dimensional vector.
A two-layer FiLM predictor maps this vector to a 512-dimensional output, which is split into two 256-dimensional vectors, \(\gamma_t\) and \(\beta_t\).
The wavelet feature is modulated by
\begin{equation}
    \tilde{F}_t = (1 + \gamma_t) \odot F_m + \beta_t,
\end{equation}
where \(\gamma_t\) and \(\beta_t\) are broadcast along the spatial dimensions.
The residual form \(1+\gamma_t\) is used to stabilize affine modulation.

The modulated feature \(\tilde{F}_t\) is transformed back by inverse Haar DWT, yielding a 64-channel feature map.
The decoder \(Q_P(\cdot)\) contains four residual blocks with 64 channels and a final \(3 \times 3\) convolutional layer that maps 64 channels to 3 channels.
A sigmoid activation is applied to produce the final RGB output \(\hat{I}_t \in [0,1]^{3 \times H \times W}\).

\section{Construction of Large-scale Paired RGB Training Data}
\label{sec:supp_rgb_pairs}

After training on RAW-guided triplets, MESS is used as an exposure-variation generator to construct paired RGB data at scale.
Given a selected high-quality RGB image \(I_{\mathrm{hq}}\), we treat it as the normal-exposure reference and synthesize an exposure-degraded counterpart \(\tilde{I}\) by feeding \(I_{\mathrm{hq}}\) and a sampled exposure condition into MESS.
The resulting pair \((\tilde{I}, I_{\mathrm{hq}})\) is used for downstream enhancement training, where \(\tilde{I}\) serves as the degraded input and \(I_{\mathrm{hq}}\) serves as the restoration target.
Although MESS can generate multiple exposure variants for each source image, we store one synthesized degraded image for each target image in the final downstream training set, resulting in 500K paired RGB samples.

\subsection{Source Image Selection}

We start from approximately 20 million in-the-wild RGB images collected from public image sources, including Unsplash~\cite{UnsplashDataset}, SA-1B~\cite{Kirillov2023SegmentAnything}, and Open Images~\cite{Kuznetsova2020OpenImages}.
Before quality ranking, we remove images that are unsuitable as restoration targets, such as images with obvious motion blur, strong compression artifacts, severe sensor noise, extremely low contrast, or large saturated or under-exposed regions.
Near-duplicate images are also removed using perceptual hashing and visual feature similarity to improve scene diversity.
For the remaining candidates, we evaluate perceptual quality using multiple no-reference IQA models, including ARNIQA~\cite{Agnolucci2024ARNIQA}, MUSIQ~\cite{Ke2021MUSIQ}, MANIQA~\cite{Yang2022MANIQA}, and TOPIQ~\cite{Chen2024TOPIQ}.
Since different IQA models have different score ranges and preference directions, their outputs are normalized before computing an ensemble quality score.
The top 500K images according to the ensemble score are retained as high-quality RGB targets.
This strategy reduces the bias of any single quality metric and favors images that are consistently sharp, natural, well-exposed, and visually pleasing.

\subsection{Exposure-degraded Input Synthesis}

For each selected target image, we sample one exposure condition within the same exposure range used for MESS training, corresponding to \([-3,+3]\) EV before normalization.
Exposure values close to normal exposure are avoided, since they usually produce trivial pairs with limited restoration difficulty.
Under-exposure and over-exposure cases are synthesized with equal probability.
To emphasize challenging degradations, the sampling distribution is biased toward larger exposure shifts, while a smaller portion of moderate exposure shifts is retained to improve training stability.
After synthesis, each generated pair is checked by lightweight validity tests.
We reject a synthesized input if it is nearly identical to the target, contains excessive black or saturated pixels, or shows visible structural artifacts produced by MESS.
For rejected samples, a new exposure condition is sampled and the synthesis process is repeated.

\subsection{Final Paired Dataset}

The final RGB training set contains 500K pairs.
Each pair consists of one MESS-synthesized exposure-degraded image and the corresponding high-quality RGB target.
Because the degraded images are generated from diverse in-the-wild scenes and cover both under-exposure and over-exposure conditions, the resulting dataset provides broad scene and degradation diversity for training a generalizable enhancement network.

\section{Additional Results with Other Downstream Backbones}
\label{sec:additional_quantitative_results}

In the main paper, we report the results using FourLLIE~\cite{wang2023fourllie} as the representative downstream backbone due to space limitation.
In this section, we provide additional experiments with more downstream image enhancement models to further verify the generality of MESS-generated training data.
Specifically, we evaluate RetinexFormer~\cite{Cai2023Retinexformer} and Multinex~\cite{multinex2026} for low-light image enhancement, ExSLOT~\cite{jung2025exposure_slot} and CSEC~\cite{Li2024CSEC} for exposure correction, and RAVE~\cite{gaintseva2024RAVE} for backlit image enhancement.

For each downstream architecture, we keep the network architecture, training resolution, optimizer, batch size, learning rate schedule, loss function, and number of training iterations fixed.
Only the training data source is changed.
All trained models are evaluated on the same five test sets, including UHDLL~\cite{li2023uhdfour}, LSRW~\cite{Hai2023R2RNet}, SICE~\cite{Cai2018SICE}, MSEC~\cite{Afifi2021Learning}, and BAID~\cite{Lv2022BacklitNet}.
We report full-reference metrics, including PSNR, SSIM, and LPIPS~\cite{zhang2018lpips}, and no-reference metrics, including ARNIQA~\cite{Agnolucci2024ARNIQA}, MUSIQ~\cite{Ke2021MUSIQ}, and MANIQA~\cite{Yang2022MANIQA}.

\subsection{Results with Low-Light Image Enhancement Backbones}

\paragraph{Evaluation with etinexFormer.}
Tables~\ref{tab:retinexformer_friqa} and~\ref{tab:retinexformer_nriqa} show the full-reference and no-reference results using RetinexFormer~\cite{Cai2023Retinexformer} as the downstream backbone.
The results demonstrate that MESS-generated data consistently provides strong cross-dataset generalization.
Compared with task-specific paired datasets, MESS-generated data achieves more balanced performance across low-light, exposure correction, and backlit enhancement benchmarks.
Compared with RGB-based synthetic datasets, MESS-generated data yields better perceptual quality and stronger robustness to diverse illumination degradations.

\paragraph{Evaluation with Multinex.}
Tables~\ref{tab:multinex_friqa} and~\ref{tab:multinex_nriqa} show the results using Multinex~\cite{multinex2026} as the downstream backbone.
MESS-generated data again achieves competitive or superior performance across different test sets.
This indicates that the benefit of MESS is not tied to a specific LLIE architecture, but comes from the improved diversity and camera-consistent exposure supervision of the generated data.

\subsection{Results with Exposure Correction Backbones}

\paragraph{Evaluation with ExSLOT.}
Tables~\ref{tab:exslot_friqa} and~\ref{tab:exslot_nriqa} report the results using ExSLOT~\cite{jung2025exposure_slot} as the downstream backbone.
Only the training data source is changed, while the backbone and training setting remain fixed.
The results verify whether MESS-generated data can improve the generalization ability of an exposure correction architecture across LLIE, EC, and BIE benchmarks.

\paragraph{Evaluation with CSEC.}
Tables~\ref{tab:csec_friqa} and~\ref{tab:csec_nriqa} report the results using CSEC~\cite{Li2024CSEC} as the downstream backbone.
Similar to other backbones, the architecture and training protocol are fixed, and only the data source is changed.
These experiments further examine whether MESS-generated training data provides architecture-agnostic improvements.

\subsection{Results with Backlit Image Enhancement Backbone}

\paragraph{Evaluation with RAVE.}
Tables~\ref{tab:rave_friqa} and~\ref{tab:rave_nriqa} report the results using RAVE~\cite{gaintseva2024RAVE} as the downstream backbone.
The same RAVE architecture is trained with different data sources and evaluated on all test sets without fine-tuning.
This setting examines whether MESS-generated data can also benefit a backlit image enhancement architecture under cross-task evaluation.

\section{Additional Qualitative Results}
\label{sec:supp_qualitative}

\begin{figure}[t!]
\centering
\begin{tabular*}{\linewidth}{@{}p{0.2\linewidth}@{}p{0.2\linewidth}@{}p{0.2\linewidth}@{}p{0.2\linewidth}@{}p{0.2\linewidth}@{}}
\centering\small Input &
\centering\small UHDLL-trained &
\centering\small LSRW-trained &
\centering\small MESS-trained &
\centering\small GT
\end{tabular*}
\vspace{1pt}
\includegraphics[width=1\textwidth]{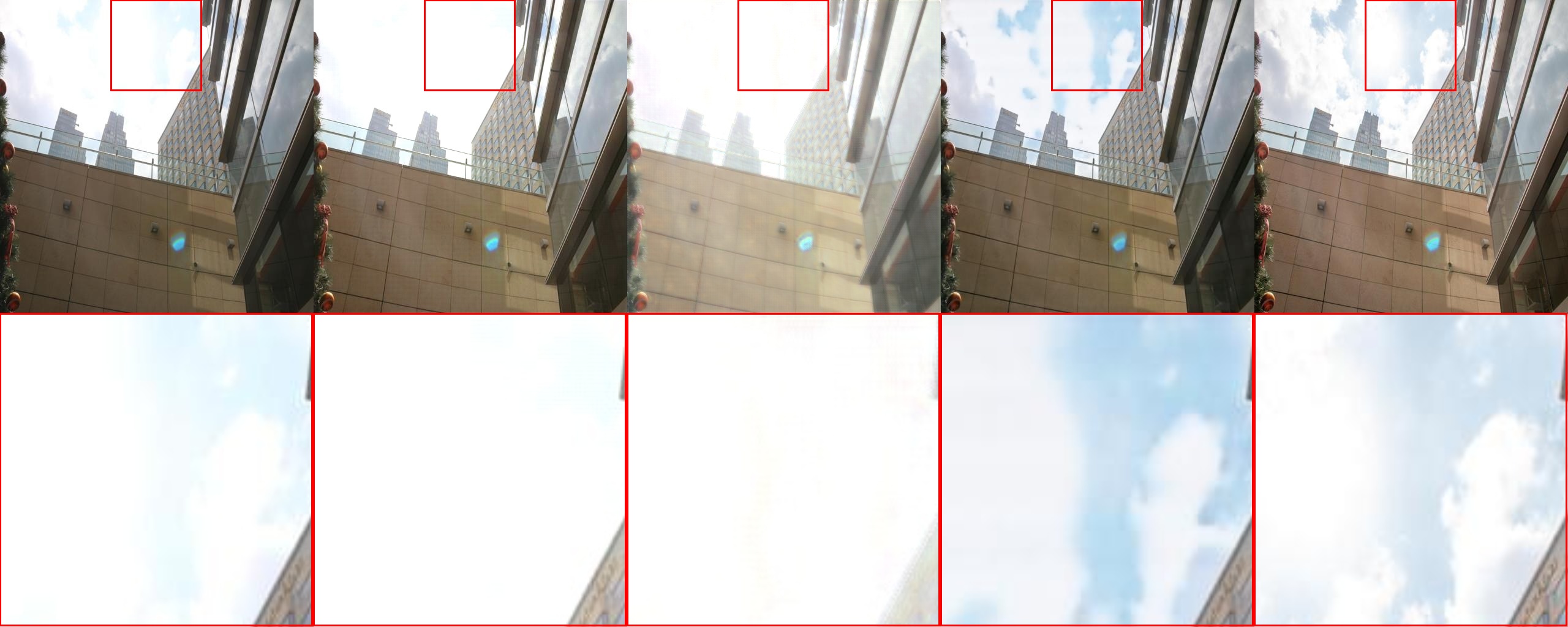}
\includegraphics[width=1\textwidth]{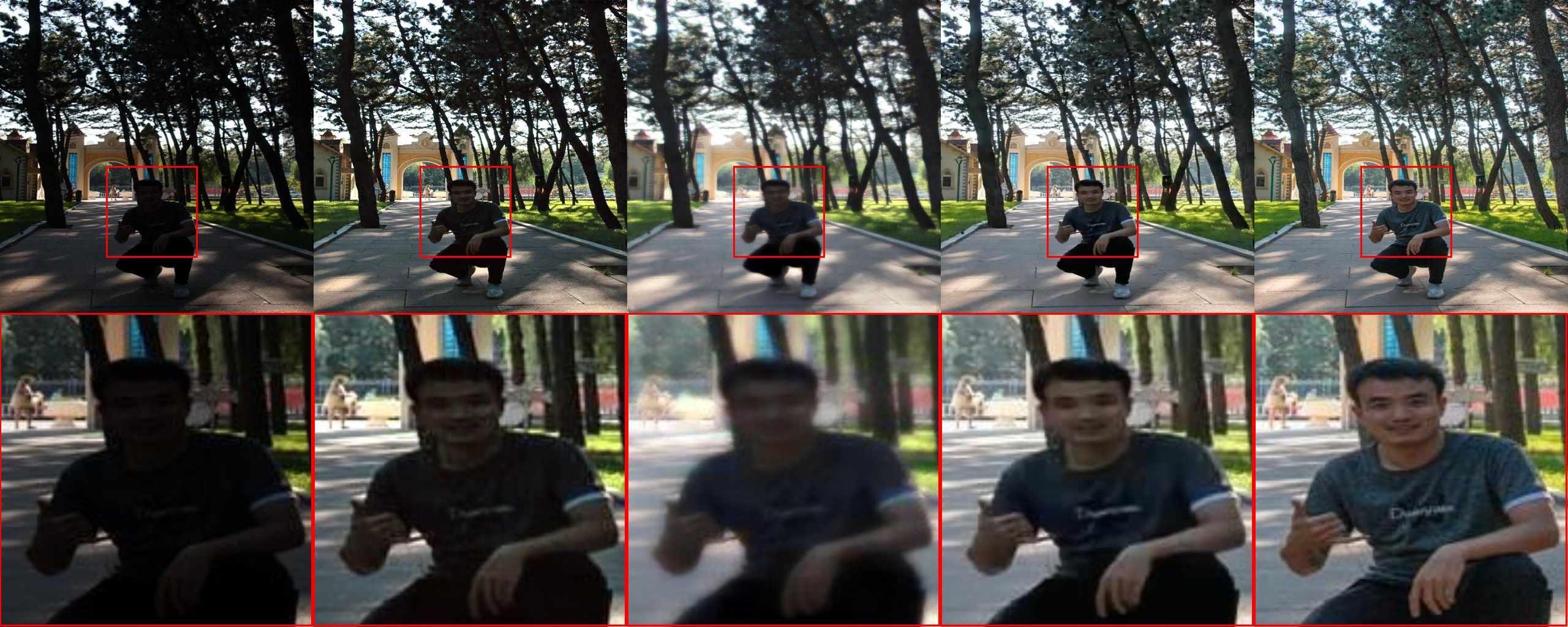}
\includegraphics[width=1\textwidth]{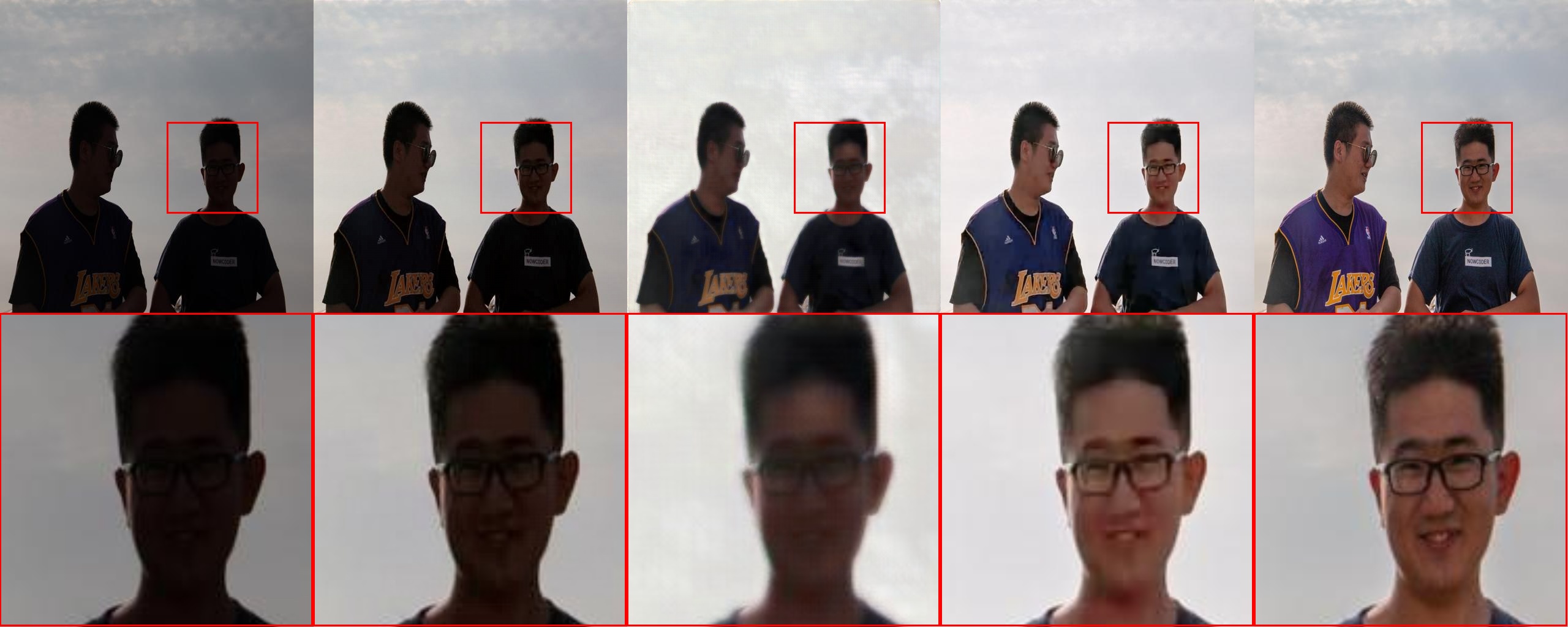}
\caption{
Qualitative comparison of the FourLLIE models trained on UHDLL, LSRW, and our MESS-generated datasets. The test images are from the BAID dataset.
}
\label{fig:qualitative_bie_supp1}
\end{figure}

\begin{figure}[t]
\centering
\begin{tabular*}{\linewidth}{@{}p{0.2\linewidth}@{}p{0.2\linewidth}@{}p{0.2\linewidth}@{}p{0.2\linewidth}@{}p{0.2\linewidth}@{}}
\centering\small Input &
\centering\small UHDLL-trained &
\centering\small LSRW-trained &
\centering\small MESS-trained &
\centering\small GT
\end{tabular*}
\vspace{1pt}
\includegraphics[width=1\textwidth]{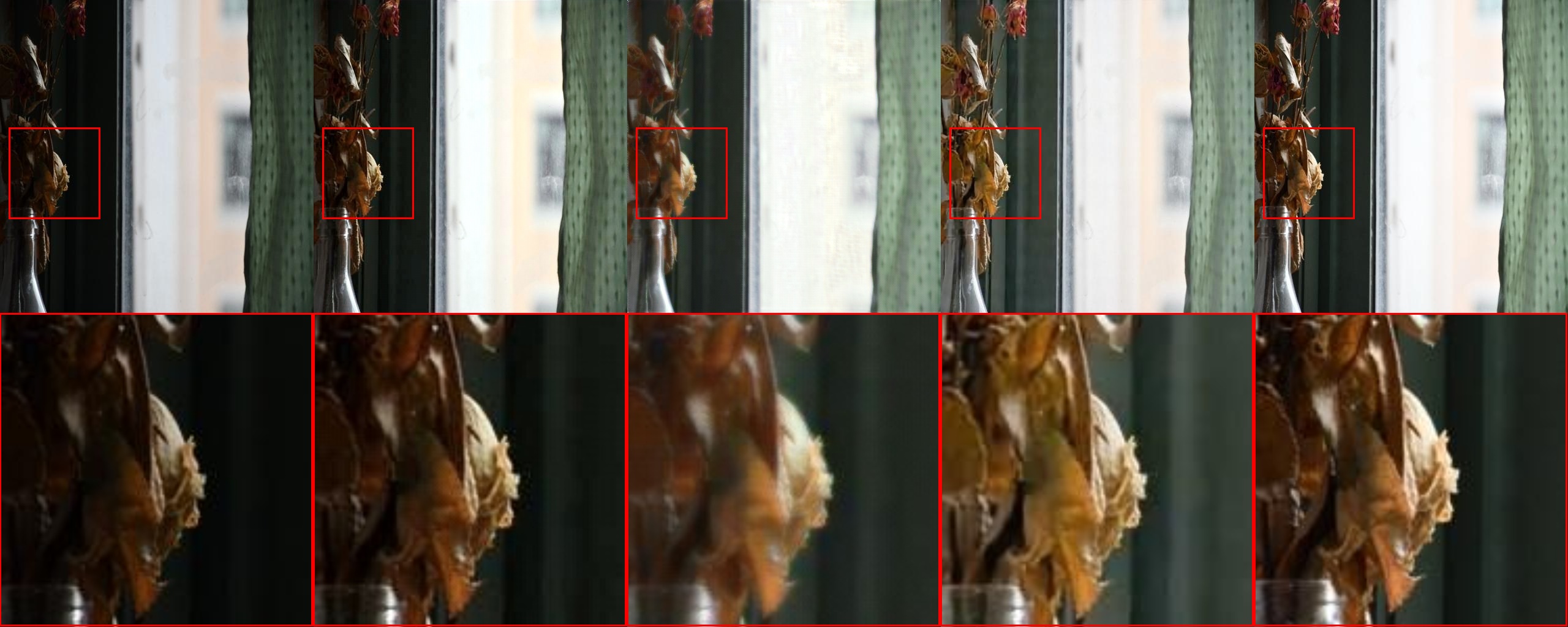}
\includegraphics[width=1\textwidth]{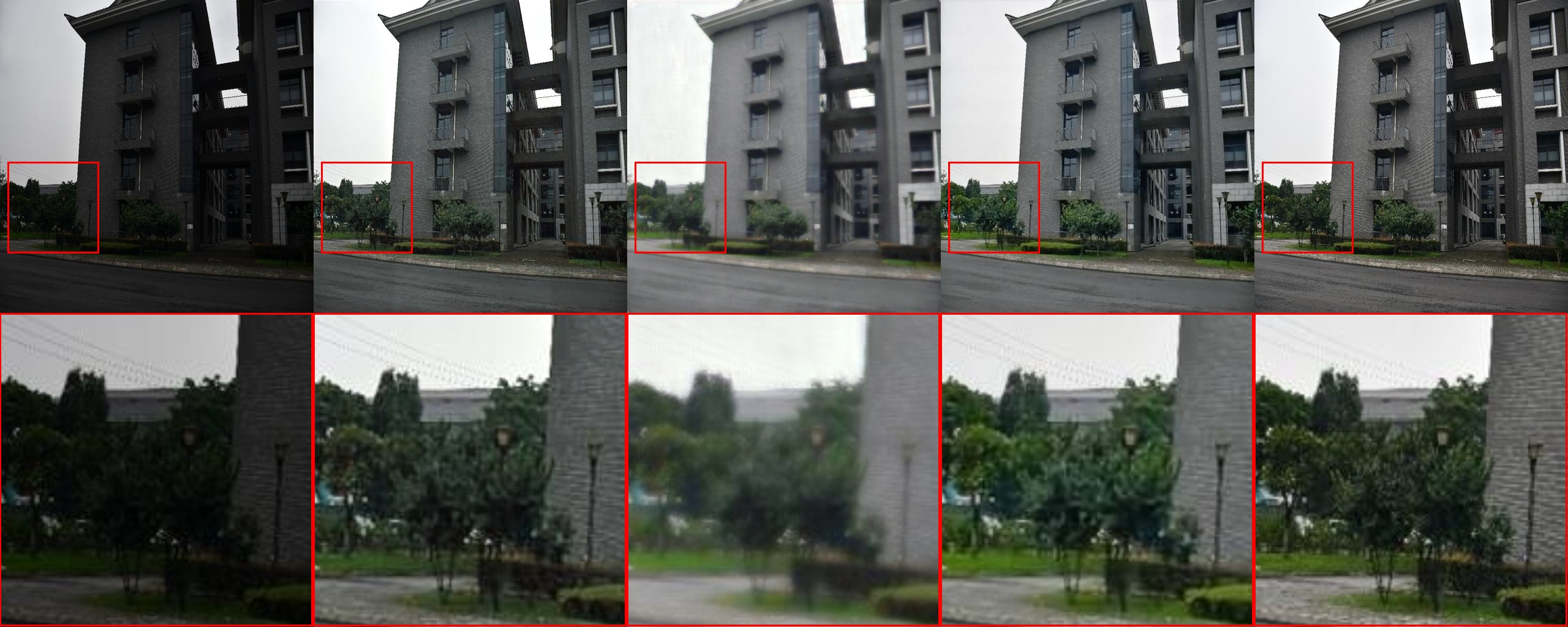}
\includegraphics[width=1\textwidth]{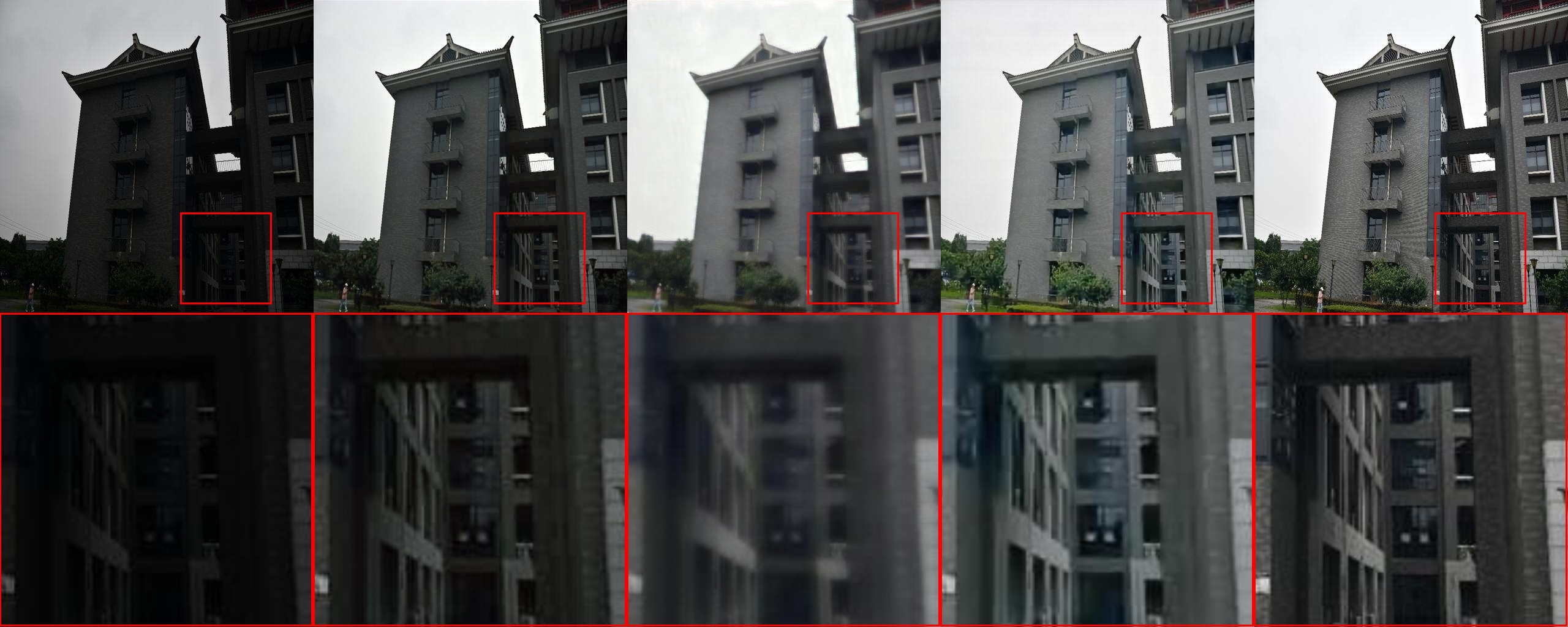}
\caption{
Qualitative comparison of the FourLLIE models trained on UHDLL, LSRW, and our MESS-generated datasets. The test images are from the LSRW dataset.
}
\label{fig:qualitative_llie_supp1}
\end{figure}

\begin{figure}[t]
\centering
\begin{tabular*}{\linewidth}{@{}p{0.2\linewidth}@{}p{0.2\linewidth}@{}p{0.2\linewidth}@{}p{0.2\linewidth}@{}p{0.2\linewidth}@{}}
\centering\small Input &
\centering\small UHDLL-trained &
\centering\small LSRW-trained &
\centering\small MESS-trained &
\centering\small GT
\end{tabular*}
\vspace{1pt}
\includegraphics[width=1\textwidth]{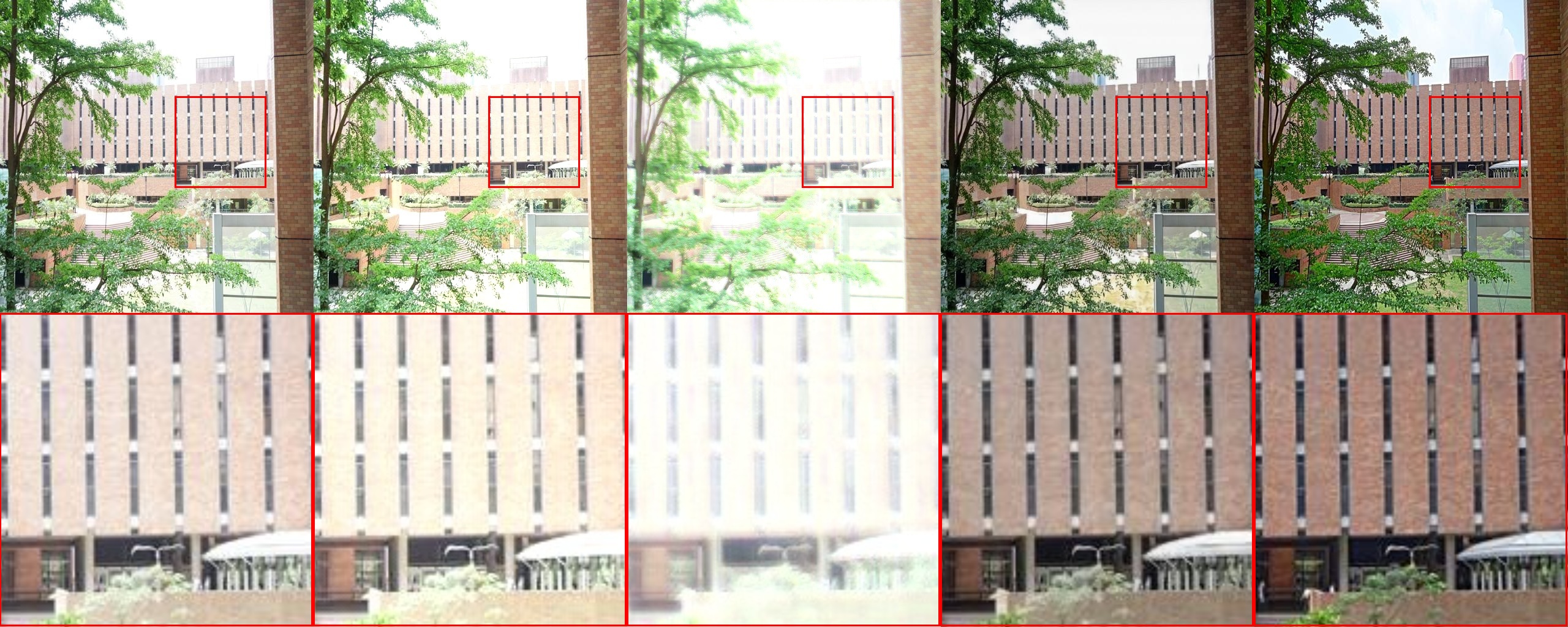}
\includegraphics[width=1\textwidth]{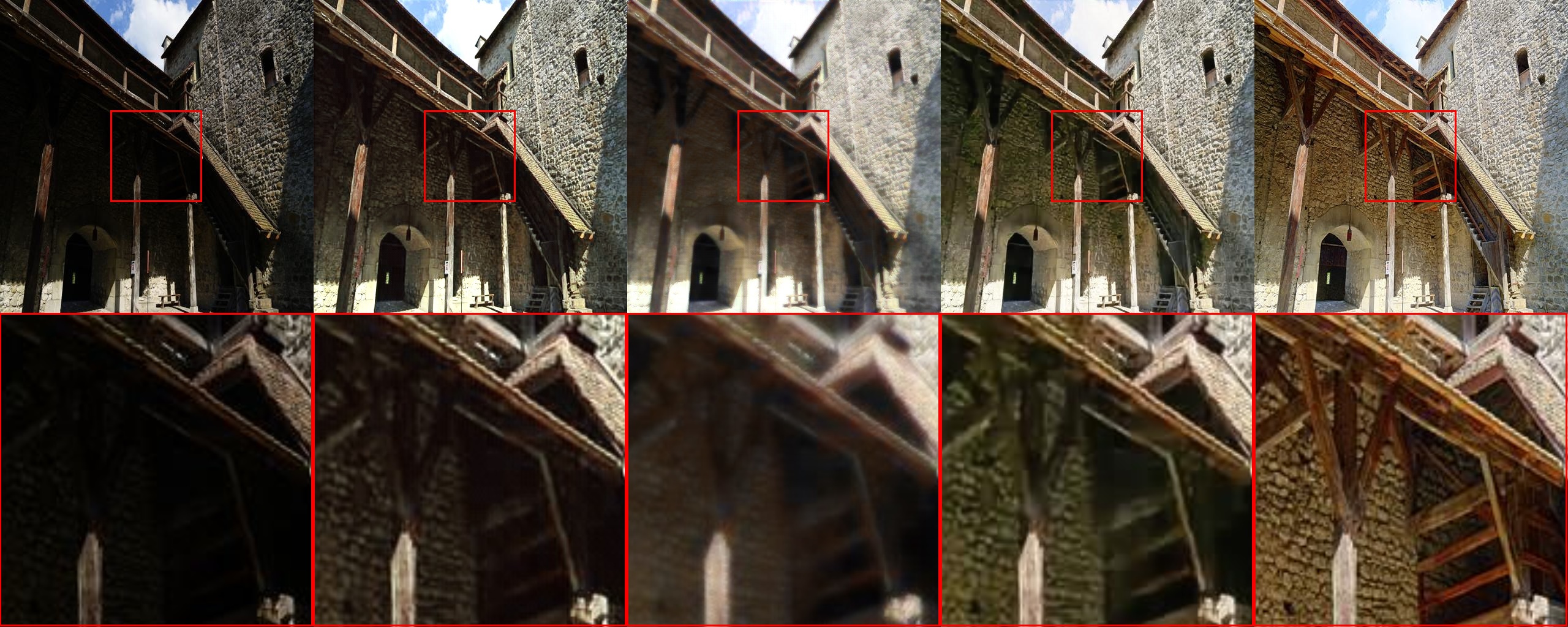}
\includegraphics[width=1\textwidth]{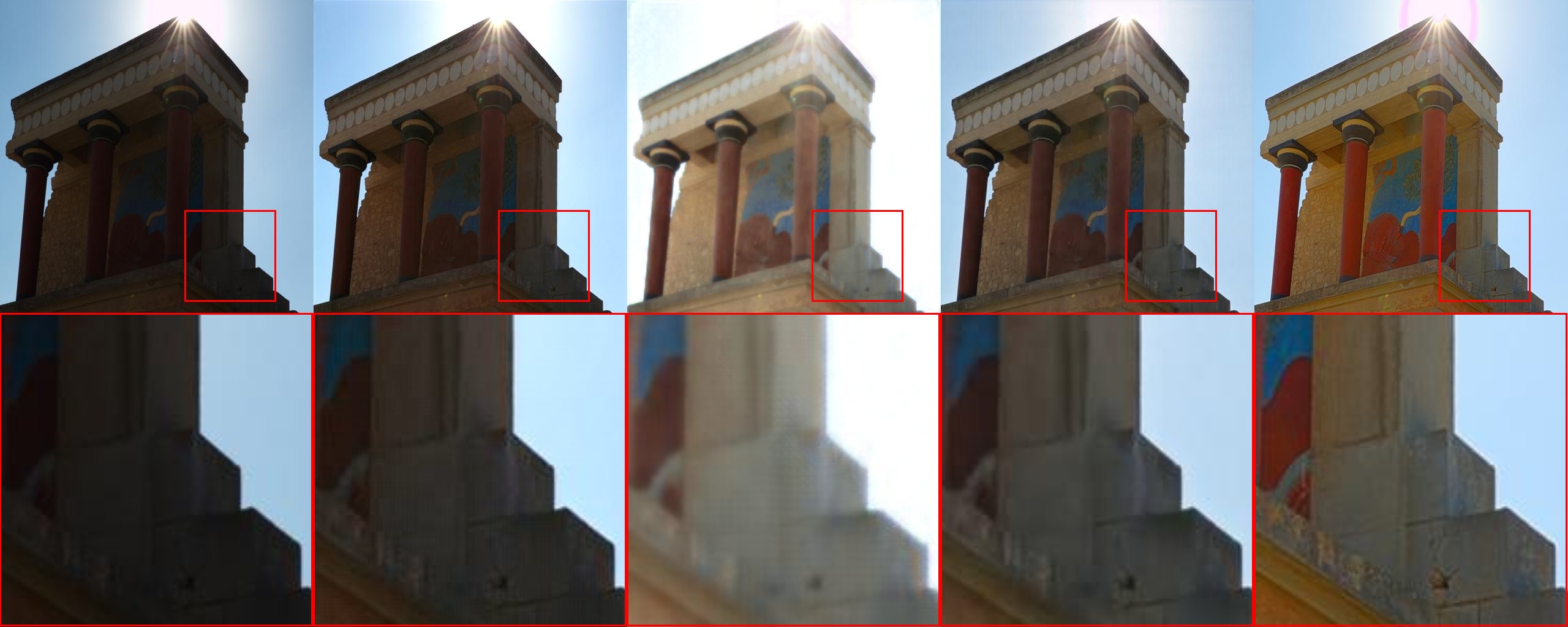}
\caption{
Qualitative comparison of the FourLLIE models trained on UHDLL, LSRW, and our MESS-generated datasets. The test images are from the SICE dataset.
}
\label{fig:qualitative_ec_supp1}
\end{figure}

In this section, we provide additional qualitative results to further analyze the generalization ability of MESS-generated training data.
Figs.~\ref{fig:qualitative_bie_supp1}, \ref{fig:qualitative_llie_supp1}, and \ref{fig:qualitative_ec_supp1} compare MESS with models trained on real-captured task-specific datasets, i.e., UHDLL and LSRW.
These results show that although real paired datasets are physically valid, their limited scene and degradation diversity makes the trained models less robust under cross-dataset and cross-task evaluation.
In particular, models trained on UHDLL or LSRW often over-brighten sky and building regions, desaturate vegetation, blur window or facade structures, and amplify local artifacts.

\begin{figure}[t]
\centering
\begin{tabular*}{\linewidth}{@{}p{0.2\linewidth}@{}p{0.2\linewidth}@{}p{0.2\linewidth}@{}p{0.2\linewidth}@{}p{0.2\linewidth}@{}}
\centering\small Input &
\centering\small MSEC-trained &
\centering\small UEC-trained &
\centering\small MESS-trained &
\centering\small GT
\end{tabular*}
\vspace{1pt}
\includegraphics[width=1\textwidth]{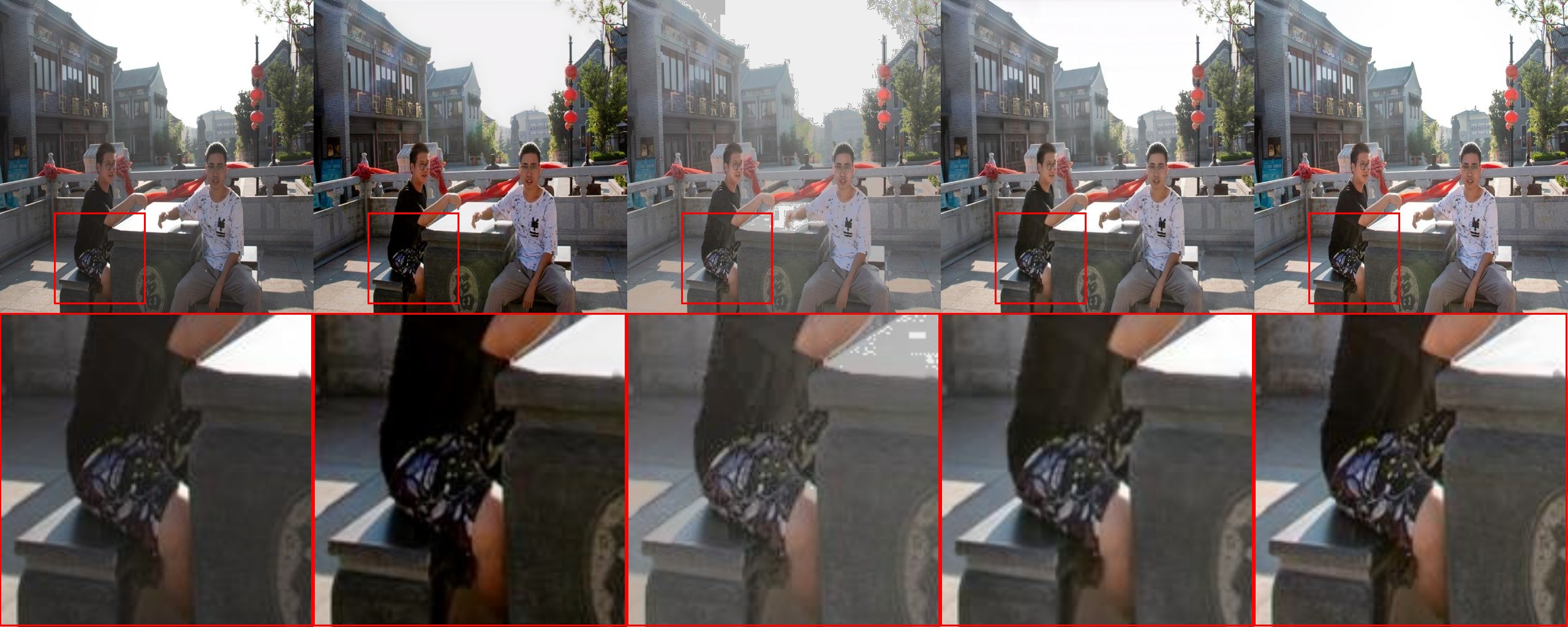}
\includegraphics[width=1\textwidth]{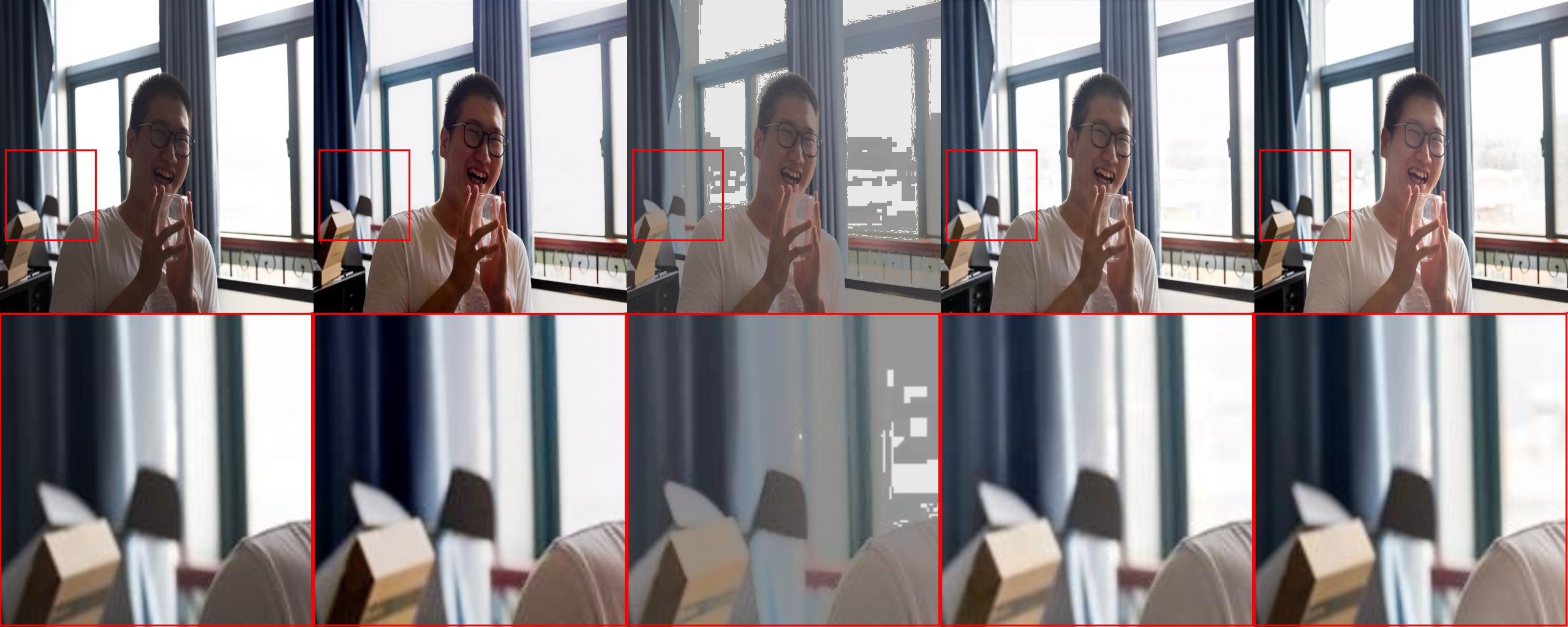}
\includegraphics[width=1\textwidth]{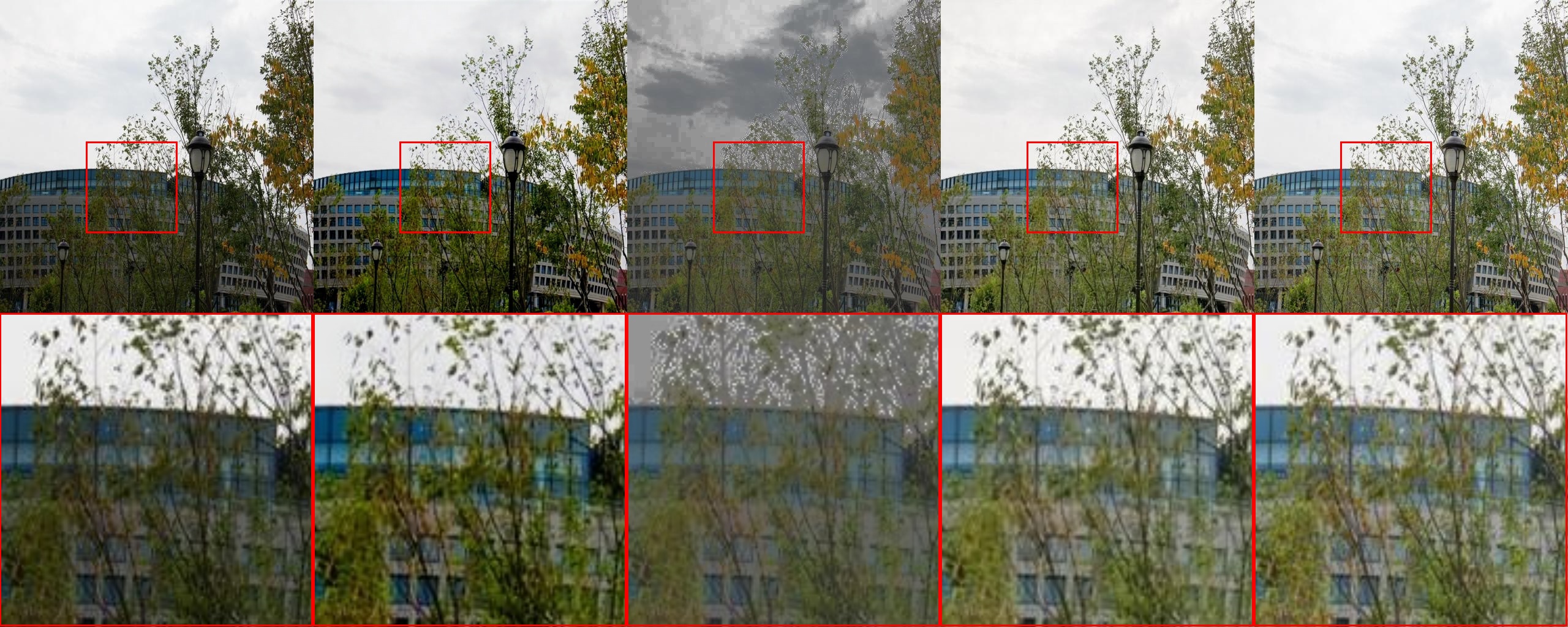}
\caption{
Qualitative comparison of the FourLLIE models trained on MSEC, UEC, and MESS-generated datasets. The test images are from the BAID dataset.
}
\label{fig:qualitative_bie_supp2}
\end{figure}

\begin{figure}[t]
\centering
\begin{tabular*}{\linewidth}{@{}p{0.2\linewidth}@{}p{0.2\linewidth}@{}p{0.2\linewidth}@{}p{0.2\linewidth}@{}p{0.2\linewidth}@{}}
\centering\small Input &
\centering\small MSEC-trained &
\centering\small UEC-trained &
\centering\small MESS-trained &
\centering\small GT
\end{tabular*}
\vspace{1pt}
\includegraphics[width=1\textwidth]{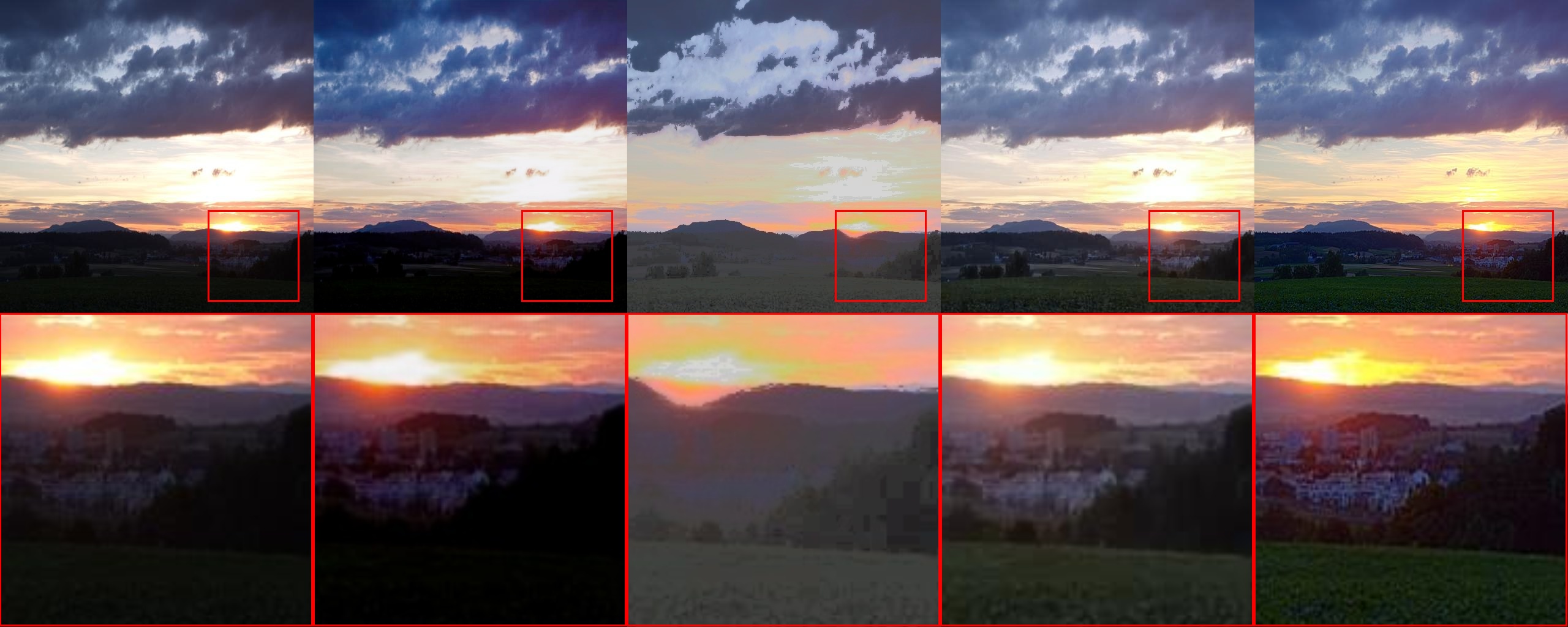}
\includegraphics[width=1\textwidth]{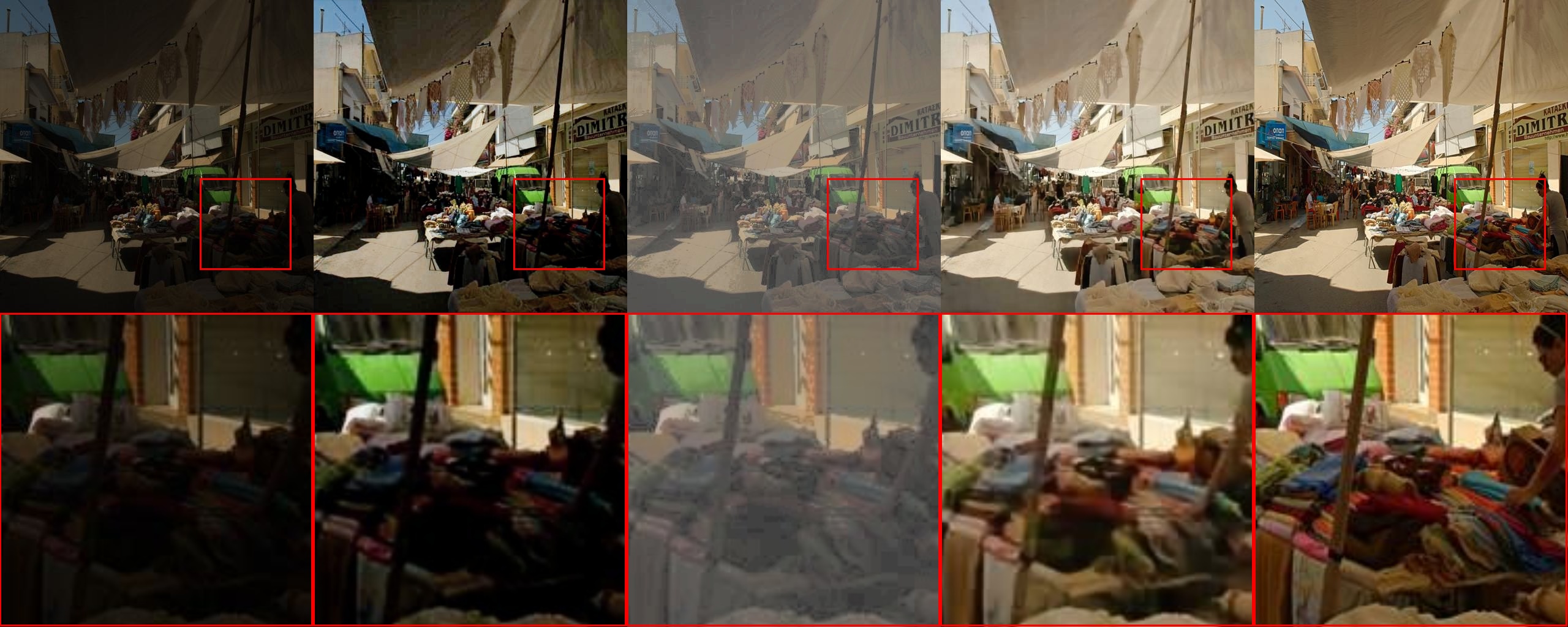}
\includegraphics[width=1\textwidth]{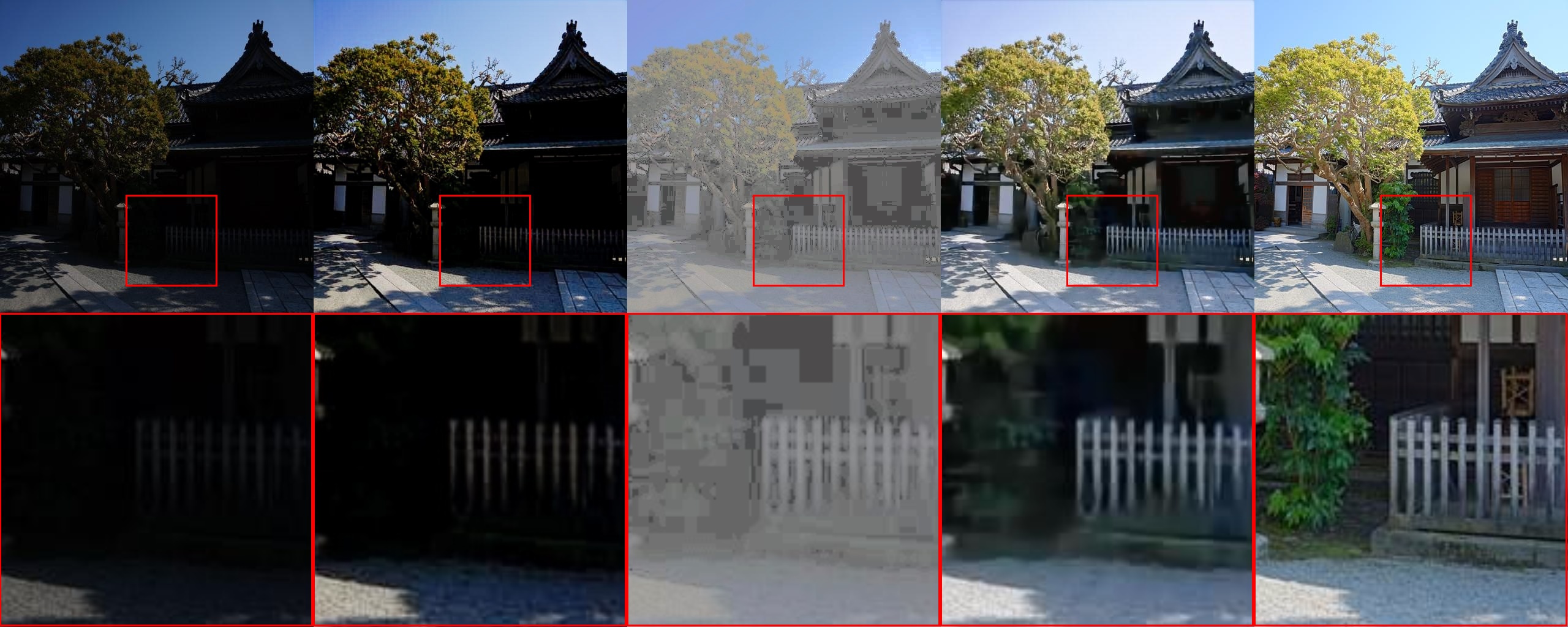}
\caption{
Qualitative comparison of the FourLLIE models trained on MSEC, UEC, and MESS-generated datasets. The test images are from the SICE dataset.
}
\label{fig:qualitative_ec_supp2}
\end{figure}

\begin{figure}[t]
\centering
\begin{tabular*}{\linewidth}{@{}p{0.2\linewidth}@{}p{0.2\linewidth}@{}p{0.2\linewidth}@{}p{0.2\linewidth}@{}p{0.2\linewidth}@{}}
\centering\small Input &
\centering\small MSEC-trained &
\centering\small UEC-trained &
\centering\small MESS-trained &
\centering\small GT
\end{tabular*}
\vspace{1pt}
\includegraphics[width=1\textwidth]{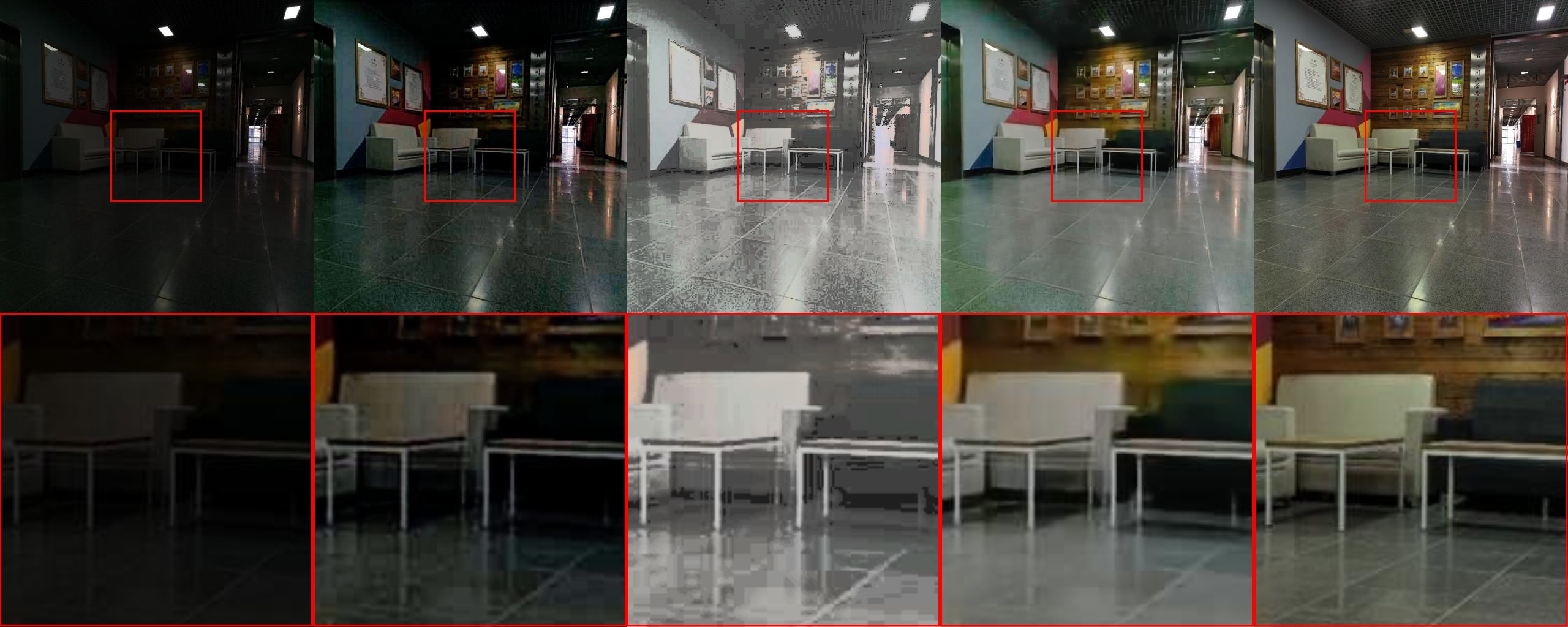}
\includegraphics[width=1\textwidth]{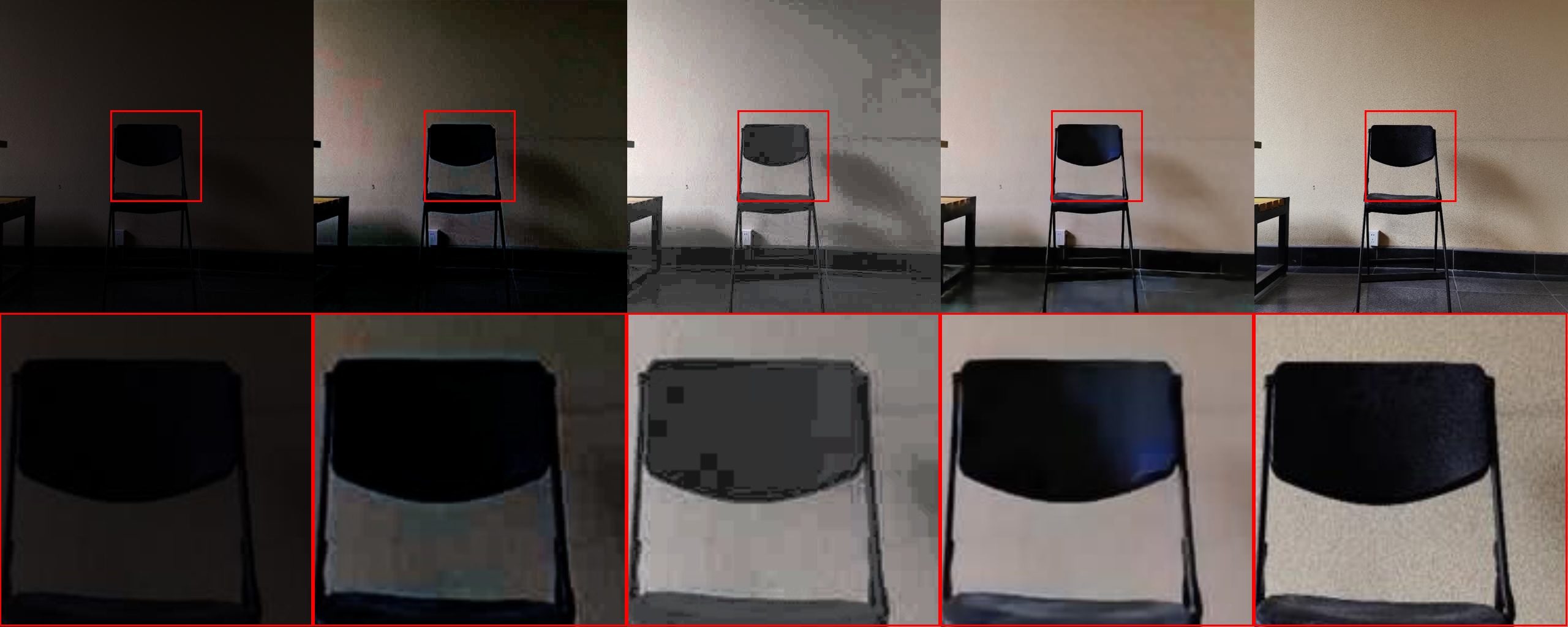}
\includegraphics[width=1\textwidth]{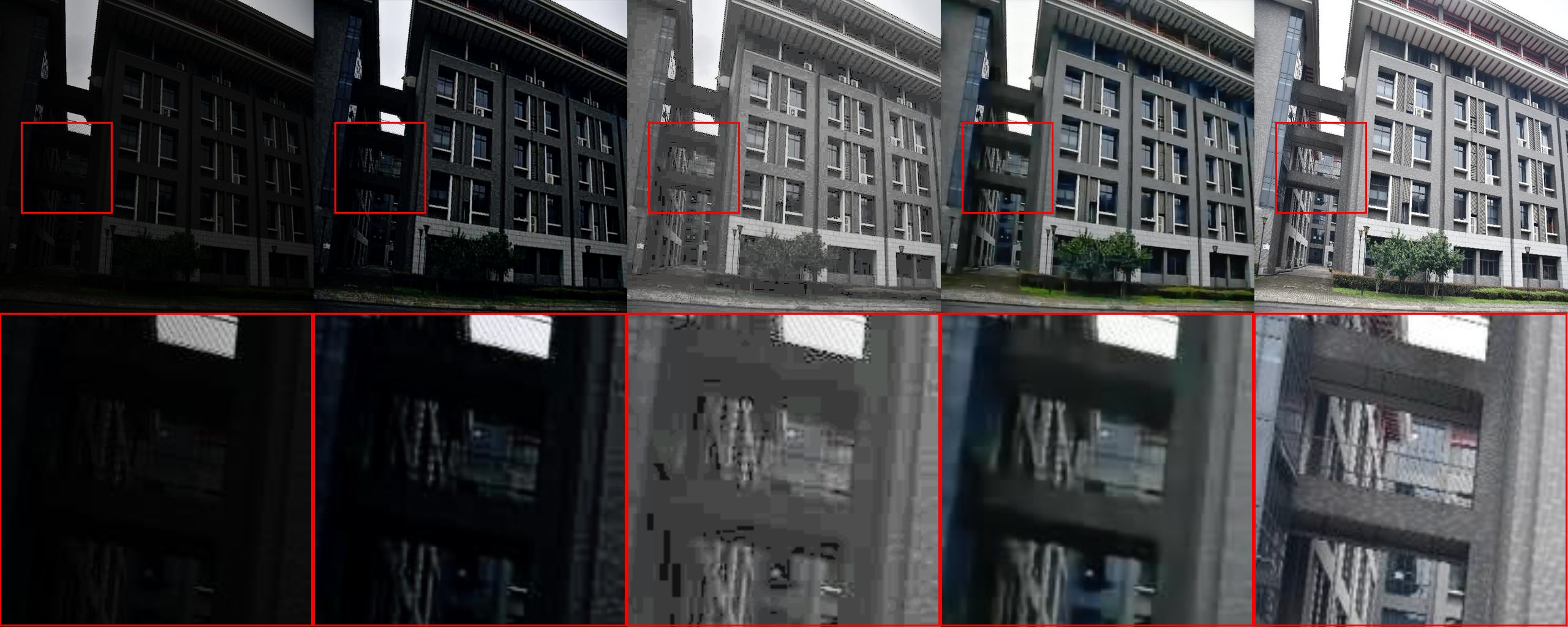}
\caption{
Qualitative comparison of the FourLLIE models trained on MSEC, UEC, and MESS-generated datasets. The test images are from the LSRW dataset.
}
\label{fig:qualitative_llie_supp2}
\end{figure}

Figs.~\ref{fig:qualitative_bie_supp2}, \ref{fig:qualitative_ec_supp2}, and \ref{fig:qualitative_llie_supp2} compare MESS with different data synthesis strategies, including RAW-based rendering with MSEC and RGB-based rendering with UEC.
As shown in these qualitative comparisons, the MESS-trained model achieves more stable visual quality, more natural color reproduction, and better preservation of object structures across low-light enhancement, exposure correction, and backlit enhancement scenarios.

\section{Ablation Study}
\label{sec:supp_ablation}

We conduct ablation studies to evaluate the contribution of each component in MESS.
All variants use the same FourLLIE backbone to synthesize paired training data, with identical downstream training and evaluation protocols.
The results are averaged over UHDLL, LSRW, SICE, MSEC, and BAID.

As shown in Table~\ref{tab:ablation_main}, removing any major component consistently degrades performance.
The representation consistency loss improves the robustness of the learned scene representation against rendering-style variations.
The frequency-based enhancement module helps separate structural information from exposure-related appearance changes, leading to more reliable synthesized training pairs.
The diffusion-based detail enhancement module mainly benefits perceptual quality by compensating for missing local details.
The exposure-conditioned rendering module is crucial for generating controllable exposure variants; without explicit exposure conditioning, the synthesized pairs contain less accurate exposure relationships and lead to weaker downstream generalization.
Replacing DWT with FFT performs better than removing frequency modeling, but remains inferior to the full model, confirming the advantage of spatially localized wavelet representations.

Table~\ref{tab:ablation_renderer} compares different exposure injection strategies.
The renderer without exposure condition performs the worst, while directly concatenating the exposure value with RGB features brings only limited improvement.
RGB-domain FiLM is more effective, but still lacks explicit frequency decomposition.
DWT-FiLM further improves the results by modulating multi-resolution frequency components.
With additional content descriptors, the full renderer achieves the best performance across all metrics, demonstrating the effectiveness of content-adaptive exposure modulation for realistic exposure synthesis.

\begin{table*}[t]
\centering
\caption{Full-reference results using RetinexFormer~\cite{Cai2023Retinexformer} backbone. The model trained on each data source is evaluated on all test sets without fine-tuning. Values in \textcolor{red}{red} and \textcolor{blue}{blue} indicate the best and second-best results, respectively.}
\label{tab:retinexformer_friqa}
\small
\setlength{\tabcolsep}{4pt}
\begin{tabular}{l|c|c|ccccc}
\hline
Training Data & Type & Metric & UHDLL & LSRW & SICE & MSEC & BAID\\
\hline

UHDLL~\cite{li2023uhdfour} & Real & PSNR $\uparrow$ & \textcolor{red}{22.509} & 19.024 & 15.687 & 18.745 & 20.025\\
LSRW~\cite{Hai2023R2RNet} & Real & PSNR $\uparrow$ & 20.175 & \textcolor{red}{20.496} & 16.745 & 18.980 & 19.440\\
SICE~\cite{Cai2018SICE} & Real & PSNR $\uparrow$ & 17.770 & 17.711 & \textcolor{red}{21.899} & 20.300 & 19.645\\
MSEC~\cite{Afifi2021Learning} & Raw-Syn & PSNR $\uparrow$ & 18.162 & 17.890 & 19.910 & \textcolor{blue}{22.809} & 19.910\\
BAID~\cite{Lv2022BacklitNet} & Real & PSNR $\uparrow$ & 16.508 & 16.591 & 17.782 & 18.750 & \textcolor{red}{21.471}\\
RCD~\cite{Cui2024UEC} & Raw-Syn & PSNR $\uparrow$ & 18.701 & 17.855 & 18.641 & 20.768 & 20.243\\

Gamma~\cite{Lore2017LLNet} & RGB-Syn & PSNR $\uparrow$ & 15.659 & 14.311 & 16.759 & 17.671 & 17.666\\
CSNorm-Fourier~\cite{Yao2023CSNorm} & RGB-Syn & PSNR $\uparrow$ & 16.885 & 16.090 & 18.370 & 19.209 & 18.950\\
UEC-based~\cite{Cui2024UEC} & RGB-Syn & PSNR $\uparrow$ & 17.976 & 17.473 & 19.731 & 20.547 & 19.981\\
MESS-generated & RGB-Syn & PSNR $\uparrow$ & \textcolor{blue}{22.231} & \textcolor{blue}{20.161} & \textcolor{blue}{21.591} & \textcolor{red}{22.936} & \textcolor{blue}{21.210}\\
\hline
\hline

UHDLL~\cite{li2023uhdfour} & Real & SSIM $\uparrow$ & \textcolor{red}{0.901} & 0.715 & 0.763 & 0.819 & 0.892\\
LSRW~\cite{Hai2023R2RNet} & Real & SSIM $\uparrow$ & 0.876 & \textcolor{red}{0.746} & 0.784 & 0.816 & 0.878\\
SICE~\cite{Cai2018SICE} & Real & SSIM $\uparrow$ & 0.812 & 0.702 & \textcolor{red}{0.864} & 0.841 & 0.881\\
MSEC~\cite{Afifi2021Learning} & Raw-Syn & SSIM $\uparrow$ & 0.821 & 0.714 & 0.816 & \textcolor{red}{0.886} & 0.878\\
BAID~\cite{Lv2022BacklitNet} & Real & SSIM $\uparrow$ & 0.771 & 0.657 & 0.764 & 0.793 & \textcolor{red}{0.917}\\
RCD~\cite{Cui2024UEC} & Raw-Syn & SSIM $\uparrow$ & 0.824 & 0.704 & 0.792 & 0.852 & 0.884\\

Gamma~\cite{Lore2017LLNet} & RGB-Syn & SSIM $\uparrow$ & 0.731 & 0.584 & 0.714 & 0.728 & 0.789\\
CSNorm-Fourier~\cite{Yao2023CSNorm} & RGB-Syn & SSIM $\uparrow$ & 0.783 & 0.630 & 0.766 & 0.783 & 0.838\\
UEC-based~\cite{Cui2024UEC} & RGB-Syn & SSIM $\uparrow$ & 0.822 & 0.672 & 0.803 & 0.827 & 0.876\\
MESS-generated & RGB-Syn & SSIM $\uparrow$ & \textcolor{blue}{0.897} & \textcolor{blue}{0.733} & \textcolor{blue}{0.851} & \textcolor{blue}{0.882} & \textcolor{blue}{0.908}\\
\hline
\hline

UHDLL~\cite{li2023uhdfour} & Real & LPIPS $\downarrow$ & \textcolor{red}{0.068} & 0.172 & 0.192 & 0.158 & 0.083\\
LSRW~\cite{Hai2023R2RNet} & Real & LPIPS $\downarrow$ & 0.095 & \textcolor{red}{0.143} & 0.189 & 0.164 & 0.103\\
SICE~\cite{Cai2018SICE} & Real & LPIPS $\downarrow$ & 0.141 & 0.176 & \textcolor{red}{0.130} & 0.143 & 0.101\\
MSEC~\cite{Afifi2021Learning} & Raw-Syn & LPIPS $\downarrow$ & 0.124 & 0.172 & 0.159 & \textcolor{blue}{0.132} & 0.094\\
BAID~\cite{Lv2022BacklitNet} & Real & LPIPS $\downarrow$ & 0.168 & 0.222 & 0.199 & 0.188 & \textcolor{red}{0.066}\\
RCD~\cite{Cui2024UEC} & Raw-Syn & LPIPS $\downarrow$ & 0.122 & 0.185 & 0.178 & 0.149 & 0.082\\

Gamma~\cite{Lore2017LLNet} & RGB-Syn & LPIPS $\downarrow$ & 0.219 & 0.286 & 0.250 & 0.202 & 0.148\\
CSNorm-Fourier~\cite{Yao2023CSNorm} & RGB-Syn & LPIPS $\downarrow$ & 0.181 & 0.238 & 0.208 & 0.230 & 0.185\\
UEC-based~\cite{Cui2024UEC} & RGB-Syn & LPIPS $\downarrow$ & 0.137 & 0.200 & 0.167 & 0.180 & 0.110\\
MESS-generated & RGB-Syn & LPIPS $\downarrow$ & \textcolor{blue}{0.075} & \textcolor{blue}{0.150} & \textcolor{blue}{0.137} & \textcolor{red}{0.129} & \textcolor{blue}{0.070}\\
\hline
\end{tabular}
\end{table*}

\begin{table*}[t]
\centering
\caption{No-reference results using RetinexFormer~\cite{Cai2023Retinexformer} backbone. The model trained on each data source is evaluated on all test sets without fine-tuning. Values in \textcolor{red}{red} and \textcolor{blue}{blue} indicate the best and second-best results, respectively.}
\label{tab:retinexformer_nriqa}
\small
\setlength{\tabcolsep}{4pt}
\begin{tabular}{l|c|c|ccccc}
\hline
Training Data & Type & Metric & UHDLL & LSRW & SICE & MSEC & BAID\\
\hline

UHDLL~\cite{li2023uhdfour} & Real & ARNIQA $\uparrow$ & \textcolor{blue}{0.667} & 0.657 & 0.693 & 0.694 & 0.680\\
LSRW~\cite{Hai2023R2RNet} & Real & ARNIQA $\uparrow$ & 0.648 & \textcolor{red}{0.685} & 0.685 & 0.692 & 0.678\\
SICE~\cite{Cai2018SICE} & Real & ARNIQA $\uparrow$ & 0.651 & 0.643 & \textcolor{blue}{0.717} & 0.711 & 0.671\\
MSEC~\cite{Afifi2021Learning} & Raw-Syn & ARNIQA $\uparrow$ & 0.647 & 0.658 & 0.707 & \textcolor{blue}{0.713} & 0.684\\
BAID~\cite{Lv2022BacklitNet} & Real & ARNIQA $\uparrow$ & 0.613 & 0.626 & 0.650 & 0.666 & \textcolor{blue}{0.705}\\
RCD~\cite{Cui2024UEC} & Raw-Syn & ARNIQA $\uparrow$ & 0.641 & 0.651 & 0.691 & 0.708 & 0.681\\

Gamma~\cite{Lore2017LLNet} & RGB-Syn & ARNIQA $\uparrow$ & 0.611 & 0.616 & 0.658 & 0.673 & 0.662\\
CSNorm-Fourier~\cite{Yao2023CSNorm} & RGB-Syn & ARNIQA $\uparrow$ & 0.616 & 0.624 & 0.671 & 0.682 & 0.672\\
UEC-based~\cite{Cui2024UEC} & RGB-Syn & ARNIQA $\uparrow$ & 0.643 & 0.653 & 0.690 & 0.699 & 0.695\\
MESS-generated & RGB-Syn & ARNIQA $\uparrow$ & \textcolor{red}{0.684} & \textcolor{blue}{0.675} & \textcolor{red}{0.723} & \textcolor{red}{0.720} & \textcolor{red}{0.716}\\
\hline
\hline

UHDLL~\cite{li2023uhdfour} & Real & MUSIQ $\uparrow$ & \textcolor{blue}{57.696} & 50.842 & 50.615 & 52.444 & 53.977\\
LSRW~\cite{Hai2023R2RNet} & Real & MUSIQ $\uparrow$ & 56.335 & \textcolor{red}{52.359} & 50.991 & 53.050 & 54.254\\
SICE~\cite{Cai2018SICE} & Real & MUSIQ $\uparrow$ & 55.283 & 50.596 & \textcolor{blue}{54.710} & 54.120 & 53.720\\
MSEC~\cite{Afifi2021Learning} & Raw-Syn & MUSIQ $\uparrow$ & 55.990 & 50.936 & 52.401 & \textcolor{blue}{54.951} & 53.831\\
BAID~\cite{Lv2022BacklitNet} & Real & MUSIQ $\uparrow$ & 52.681 & 49.143 & 50.675 & 51.855 & \textcolor{blue}{56.205}\\
RCD~\cite{Cui2024UEC} & Raw-Syn & MUSIQ $\uparrow$ & 55.661 & 50.558 & 51.696 & 53.987 & 54.331\\

Gamma~\cite{Lore2017LLNet} & RGB-Syn & MUSIQ $\uparrow$ & 53.000 & 49.417 & 51.267 & 52.325 & 53.268\\
CSNorm-Fourier~\cite{Yao2023CSNorm} & RGB-Syn & MUSIQ $\uparrow$ & 53.921 & 49.604 & 51.952 & 53.055 & 53.892\\
UEC-based~\cite{Cui2024UEC} & RGB-Syn & MUSIQ $\uparrow$ & 54.843 & 50.089 & 52.785 & 53.791 & 54.757\\
MESS-generated & RGB-Syn & MUSIQ $\uparrow$ & \textcolor{red}{57.928} & \textcolor{blue}{52.266} & \textcolor{red}{55.416} & \textcolor{red}{55.112} & \textcolor{red}{56.429}\\
\hline
\hline

UHDLL~\cite{li2023uhdfour} & Real & MANIQA $\uparrow$ & \textcolor{blue}{0.465} & 0.407 & 0.432 & 0.455 & 0.422\\
LSRW~\cite{Hai2023R2RNet} & Real & MANIQA $\uparrow$ & 0.461 & \textcolor{blue}{0.419} & 0.437 & 0.445 & 0.418\\
SICE~\cite{Cai2018SICE} & Real & MANIQA $\uparrow$ & 0.449 & 0.398 & \textcolor{blue}{0.460} & 0.460 & 0.414\\
MSEC~\cite{Afifi2021Learning} & Raw-Syn & MANIQA $\uparrow$ & 0.450 & 0.402 & 0.455 & \textcolor{red}{0.466} & 0.425\\
BAID~\cite{Lv2022BacklitNet} & Real & MANIQA $\uparrow$ & 0.419 & 0.387 & 0.417 & 0.431 & \textcolor{blue}{0.440}\\
RCD~\cite{Cui2024UEC} & Raw-Syn & MANIQA $\uparrow$ & 0.443 & 0.404 & 0.445 & \textcolor{blue}{0.464} & 0.422\\

Gamma~\cite{Lore2017LLNet} & RGB-Syn & MANIQA $\uparrow$ & 0.418 & 0.383 & 0.420 & 0.419 & 0.396\\
CSNorm-Fourier~\cite{Yao2023CSNorm} & RGB-Syn & MANIQA $\uparrow$ & 0.436 & 0.392 & 0.431 & 0.439 & 0.409\\
UEC-based~\cite{Cui2024UEC} & RGB-Syn & MANIQA $\uparrow$ & 0.448 & 0.412 & 0.439 & 0.450 & 0.415\\
MESS-generated & RGB-Syn & MANIQA $\uparrow$ & \textcolor{red}{0.477} & \textcolor{red}{0.429} & \textcolor{red}{0.466} & 0.462 & \textcolor{red}{0.448}\\
\hline
\end{tabular}
\end{table*}

\begin{table*}[t]
\centering
\caption{Full-reference results using Multinex~\cite{multinex2026} backbone. The model trained on each data source is evaluated on all test sets without fine-tuning. Values in \textcolor{red}{red} and \textcolor{blue}{blue} indicate the best and second-best results, respectively.}
\label{tab:multinex_friqa}
\small
\setlength{\tabcolsep}{4pt}
\begin{tabular}{l|c|c|ccccc}
\hline
Training Data & Type & Metric & UHDLL & LSRW & SICE & MSEC & BAID\\
\hline

UHDLL~\cite{li2023uhdfour} & Real & PSNR $\uparrow$ & \textcolor{red}{20.622} & 17.579 & 14.033 & 18.743 & 18.473\\
LSRW~\cite{Hai2023R2RNet} & Real & PSNR $\uparrow$ & 18.295 & \textcolor{red}{19.061} & 15.103 & 18.971 & 17.878\\
SICE~\cite{Cai2018SICE} & Real & PSNR $\uparrow$ & 15.878 & 16.274 & \textcolor{red}{20.073} & 20.302 & 18.084\\
MSEC~\cite{Afifi2021Learning} & Raw-Syn & PSNR $\uparrow$ & 16.274 & 16.448 & 18.265 & \textcolor{red}{22.322} & 18.345\\
BAID~\cite{Lv2022BacklitNet} & Real & PSNR $\uparrow$ & 14.611 & 15.146 & 16.149 & 18.748 & \textcolor{red}{19.918}\\
RCD~\cite{Cui2024UEC} & Raw-Syn & PSNR $\uparrow$ & 16.818 & 16.417 & 16.995 & 20.769 & 18.685\\

Gamma~\cite{Lore2017LLNet} & RGB-Syn & PSNR $\uparrow$ & 13.768 & 12.886 & 15.123 & 17.669 & 16.110\\
CSNorm-Fourier~\cite{Yao2023CSNorm} & RGB-Syn & PSNR $\uparrow$ & 14.994 & 14.662 & 16.731 & 19.209 & 17.382\\
UEC-based~\cite{Cui2024UEC} & RGB-Syn & PSNR $\uparrow$ & 16.087 & 16.038 & 18.092 & 20.551 & 18.433\\
MESS-generated & RGB-Syn & PSNR $\uparrow$ & \textcolor{blue}{20.343} & \textcolor{blue}{18.727} & \textcolor{blue}{19.950} & \textcolor{blue}{22.181} & \textcolor{blue}{19.657}\\
\hline
\hline

UHDLL~\cite{li2023uhdfour} & Real & SSIM $\uparrow$ & \textcolor{red}{0.868} & 0.689 & 0.734 & 0.819 & 0.869\\
LSRW~\cite{Hai2023R2RNet} & Real & SSIM $\uparrow$ & 0.844 & \textcolor{red}{0.720} & 0.756 & 0.815 & 0.855\\
SICE~\cite{Cai2018SICE} & Real & SSIM $\uparrow$ & 0.778 & 0.676 & \textcolor{blue}{0.827} & 0.841 & 0.859\\
MSEC~\cite{Afifi2021Learning} & Raw-Syn & SSIM $\uparrow$ & 0.787 & 0.688 & 0.787 & \textcolor{blue}{0.875} & 0.855\\
BAID~\cite{Lv2022BacklitNet} & Real & SSIM $\uparrow$ & 0.738 & 0.631 & 0.736 & 0.792 & \textcolor{red}{0.894}\\
RCD~\cite{Cui2024UEC} & Raw-Syn & SSIM $\uparrow$ & 0.799 & 0.674 & 0.762 & 0.857 & 0.863\\

Gamma~\cite{Lore2017LLNet} & RGB-Syn & SSIM $\uparrow$ & 0.699 & 0.557 & 0.685 & 0.744 & 0.765\\
CSNorm-Fourier~\cite{Yao2023CSNorm} & RGB-Syn & SSIM $\uparrow$ & 0.751 & 0.603 & 0.738 & 0.766 & 0.816\\
UEC-based~\cite{Cui2024UEC} & RGB-Syn & SSIM $\uparrow$ & 0.788 & 0.646 & 0.776 & 0.827 & 0.853\\
MESS-generated & RGB-Syn & SSIM $\uparrow$ & \textcolor{blue}{0.864} & \textcolor{blue}{0.707} & \textcolor{red}{0.832} & \textcolor{red}{0.879} & \textcolor{blue}{0.885}\\
\hline
\hline

UHDLL~\cite{li2023uhdfour} & Real & LPIPS $\downarrow$ & \textcolor{red}{0.118} & 0.200 & 0.224 & 0.159 & 0.124\\
LSRW~\cite{Hai2023R2RNet} & Real & LPIPS $\downarrow$ & 0.146 & \textcolor{red}{0.170} & 0.221 & 0.164 & 0.143\\
SICE~\cite{Cai2018SICE} & Real & LPIPS $\downarrow$ & 0.191 & 0.203 & \textcolor{red}{0.162} & 0.143 & 0.142\\
MSEC~\cite{Afifi2021Learning} & Raw-Syn & LPIPS $\downarrow$ & 0.175 & 0.200 & 0.190 & \textcolor{blue}{0.135} & 0.135\\
BAID~\cite{Lv2022BacklitNet} & Real & LPIPS $\downarrow$ & 0.218 & 0.248 & 0.231 & 0.188 & \textcolor{red}{0.106}\\
RCD~\cite{Cui2024UEC} & Raw-Syn & LPIPS $\downarrow$ & 0.179 & 0.216 & 0.206 & 0.144 & 0.125\\

Gamma~\cite{Lore2017LLNet} & RGB-Syn & LPIPS $\downarrow$ & 0.269 & 0.314 & 0.282 & 0.210 & 0.200\\
CSNorm-Fourier~\cite{Yao2023CSNorm} & RGB-Syn & LPIPS $\downarrow$ & 0.231 & 0.266 & 0.240 & 0.222 & 0.214\\
UEC-based~\cite{Cui2024UEC} & RGB-Syn & LPIPS $\downarrow$ & 0.187 & 0.229 & 0.199 & 0.179 & 0.151\\
MESS-generated & RGB-Syn & LPIPS $\downarrow$ & \textcolor{blue}{0.125} & \textcolor{blue}{0.177} & \textcolor{blue}{0.169} & \textcolor{red}{0.132} & \textcolor{blue}{0.111}\\
\hline
\end{tabular}
\end{table*}

\begin{table*}[t]
\centering
\caption{No-reference results using Multinex~\cite{multinex2026} backbone. The model trained on each data source is evaluated on all test sets without fine-tuning. Values in \textcolor{red}{red} and \textcolor{blue}{blue} indicate the best and second-best results, respectively.}
\label{tab:multinex_nriqa}
\small
\setlength{\tabcolsep}{4pt}
\begin{tabular}{l|c|c|ccccc}
\hline
Training Data & Type & Metric & UHDLL & LSRW & SICE & MSEC & BAID\\
\hline

UHDLL~\cite{li2023uhdfour} & Real & ARNIQA $\uparrow$ & \textcolor{blue}{0.614} & 0.622 & 0.672 & 0.692 & 0.646\\
LSRW~\cite{Hai2023R2RNet} & Real & ARNIQA $\uparrow$ & 0.595 & \textcolor{red}{0.645} & 0.666 & 0.689 & 0.639\\
SICE~\cite{Cai2018SICE} & Real & ARNIQA $\uparrow$ & 0.598 & 0.605 & \textcolor{blue}{0.699} & \textcolor{red}{0.710} & 0.637\\
MSEC~\cite{Afifi2021Learning} & Raw-Syn & ARNIQA $\uparrow$ & 0.612 & 0.618 & 0.690 & 0.702 & 0.647\\
BAID~\cite{Lv2022BacklitNet} & Real & ARNIQA $\uparrow$ & 0.559 & 0.585 & 0.631 & 0.675 & \textcolor{blue}{0.669}\\
RCD~\cite{Cui2024UEC} & Raw-Syn & ARNIQA $\uparrow$ & 0.602 & 0.615 & 0.672 & 0.703 & 0.653\\

Gamma~\cite{Lore2017LLNet} & RGB-Syn & ARNIQA $\uparrow$ & 0.560 & 0.577 & 0.641 & 0.666 & 0.627\\
CSNorm-Fourier~\cite{Yao2023CSNorm} & RGB-Syn & ARNIQA $\uparrow$ & 0.564 & 0.585 & 0.656 & 0.680 & 0.636\\
UEC-based~\cite{Cui2024UEC} & RGB-Syn & ARNIQA $\uparrow$ & 0.586 & 0.614 & 0.673 & 0.701 & 0.659\\
MESS-generated & RGB-Syn & ARNIQA $\uparrow$ & \textcolor{red}{0.631} & \textcolor{blue}{0.635} & \textcolor{red}{0.705} & \textcolor{blue}{0.708} & \textcolor{red}{0.680}\\
\hline
\hline

UHDLL~\cite{li2023uhdfour} & Real & MUSIQ $\uparrow$ & \textcolor{blue}{58.031} & 51.432 & 55.099 & 52.412 & 58.954\\
LSRW~\cite{Hai2023R2RNet} & Real & MUSIQ $\uparrow$ & 56.670 & \textcolor{red}{52.985} & 55.401 & 53.058 & 59.207\\
SICE~\cite{Cai2018SICE} & Real & MUSIQ $\uparrow$ & 55.588 & 51.224 & \textcolor{blue}{59.133} & 54.147 & 58.732\\
MSEC~\cite{Afifi2021Learning} & Raw-Syn & MUSIQ $\uparrow$ & 56.328 & 51.563 & 56.852 & \textcolor{red}{54.650} & 58.789\\
BAID~\cite{Lv2022BacklitNet} & Real & MUSIQ $\uparrow$ & 53.023 & 50.013 & 55.005 & 51.868 & \textcolor{blue}{61.163}\\
RCD~\cite{Cui2024UEC} & Raw-Syn & MUSIQ $\uparrow$ & 56.001 & 51.229 & 56.137 & 53.982 & 59.295\\

Gamma~\cite{Lore2017LLNet} & RGB-Syn & MUSIQ $\uparrow$ & 53.318 & 49.751 & 55.660 & 52.316 & 58.237\\
CSNorm-Fourier~\cite{Yao2023CSNorm} & RGB-Syn & MUSIQ $\uparrow$ & 54.259 & 50.246 & 56.391 & 53.079 & 58.858\\
UEC-based~\cite{Cui2024UEC} & RGB-Syn & MUSIQ $\uparrow$ & 55.227 & 50.753 & 57.241 & 53.786 & 59.709\\
MESS-generated & RGB-Syn & MUSIQ $\uparrow$ & \textcolor{red}{58.263} & \textcolor{blue}{52.892} & \textcolor{red}{59.839} & \textcolor{blue}{54.507} & \textcolor{red}{61.387}\\
\hline
\hline

UHDLL~\cite{li2023uhdfour} & Real & MANIQA $\uparrow$ & \textcolor{blue}{0.417} & 0.372 & 0.407 & 0.454 & 0.377\\
LSRW~\cite{Hai2023R2RNet} & Real & MANIQA $\uparrow$ & 0.412 & \textcolor{blue}{0.382} & 0.416 & 0.446 & 0.368\\
SICE~\cite{Cai2018SICE} & Real & MANIQA $\uparrow$ & 0.402 & 0.360 & \textcolor{blue}{0.437} & 0.460 & 0.368\\
MSEC~\cite{Afifi2021Learning} & Raw-Syn & MANIQA $\uparrow$ & 0.400 & 0.367 & 0.431 & 0.455 & 0.378\\
BAID~\cite{Lv2022BacklitNet} & Real & MANIQA $\uparrow$ & 0.370 & 0.351 & 0.396 & 0.434 & \textcolor{red}{0.402}\\
RCD~\cite{Cui2024UEC} & Raw-Syn & MANIQA $\uparrow$ & 0.397 & 0.361 & 0.412 & \textcolor{blue}{0.461} & 0.387\\

Gamma~\cite{Lore2017LLNet} & RGB-Syn & MANIQA $\uparrow$ & 0.368 & 0.346 & 0.396 & 0.421 & 0.353\\
CSNorm-Fourier~\cite{Yao2023CSNorm} & RGB-Syn & MANIQA $\uparrow$ & 0.385 & 0.353 & 0.406 & 0.441 & 0.362\\
UEC-based~\cite{Cui2024UEC} & RGB-Syn & MANIQA $\uparrow$ & 0.395 & 0.375 & 0.418 & 0.451 & 0.370\\
MESS-generated & RGB-Syn & MANIQA $\uparrow$ & \textcolor{red}{0.427} & \textcolor{red}{0.392} & \textcolor{red}{0.442} & \textcolor{red}{0.462} & \textcolor{blue}{0.394}\\
\hline
\end{tabular}
\end{table*}

\begin{table*}[t]
\centering
\caption{Full-reference results using ExSLOT~\cite{jung2025exposure_slot} backbone. The model trained on each data source is evaluated on all test sets without fine-tuning. Values in \textcolor{red}{red} and \textcolor{blue}{blue} indicate the best and second-best results, respectively.}
\label{tab:exslot_friqa}
\small
\setlength{\tabcolsep}{4pt}
\begin{tabular}{l|c|c|ccccc}
\hline
Training Data & Type & Metric & UHDLL & LSRW & SICE & MSEC & BAID\\
\hline

UHDLL~\cite{li2023uhdfour} & Real & PSNR $\uparrow$ & \textcolor{red}{19.254} & 16.508 & 12.332 & 17.656 & 16.346\\
LSRW~\cite{Hai2023R2RNet} & Real & PSNR $\uparrow$ & 16.957 & \textcolor{red}{17.977} & 13.427 & 18.935 & 15.748\\
SICE~\cite{Cai2018SICE} & Real & PSNR $\uparrow$ & 14.518 & 15.173 & \textcolor{blue}{17.934} & 20.345 & 16.013\\
MSEC~\cite{Afifi2021Learning} & Raw-Syn & PSNR $\uparrow$ & 14.927 & 15.361 & 16.544 & \textcolor{blue}{23.200} & 16.222\\
BAID~\cite{Lv2022BacklitNet} & Real & PSNR $\uparrow$ & 13.270 & 14.047 & 14.439 & 18.762 & \textcolor{red}{17.823}\\
RCD~\cite{Cui2024UEC} & Raw-Syn & PSNR $\uparrow$ & 15.469 & 15.322 & 15.283 & 20.558 & 16.564\\

Gamma~\cite{Lore2017LLNet} & RGB-Syn & PSNR $\uparrow$ & 12.381 & 11.825 & 13.404 & 18.721 & 14.002\\
CSNorm-Fourier~\cite{Yao2023CSNorm} & RGB-Syn & PSNR $\uparrow$ & 13.646 & 13.585 & 14.983 & 19.243 & 15.261\\
UEC-based~\cite{Cui2024UEC} & RGB-Syn & PSNR $\uparrow$ & 14.697 & 14.989 & 16.393 & 20.578 & 16.332\\
MESS-generated & RGB-Syn & PSNR $\uparrow$ & \textcolor{blue}{18.976} & \textcolor{blue}{17.642} & \textcolor{red}{18.550} & \textcolor{red}{23.329} & \textcolor{blue}{17.562}\\
\hline
\hline

UHDLL~\cite{li2023uhdfour} & Real & SSIM $\uparrow$ & \textcolor{red}{0.859} & 0.673 & 0.660 & 0.821 & 0.857\\
LSRW~\cite{Hai2023R2RNet} & Real & SSIM $\uparrow$ & 0.837 & \textcolor{red}{0.702} & 0.733 & 0.813 & 0.843\\
SICE~\cite{Cai2018SICE} & Real & SSIM $\uparrow$ & 0.769 & 0.660 & \textcolor{red}{0.811} & 0.841 & 0.848\\
MSEC~\cite{Afifi2021Learning} & Raw-Syn & SSIM $\uparrow$ & 0.776 & 0.666 & 0.762 & \textcolor{blue}{0.890} & 0.842\\
BAID~\cite{Lv2022BacklitNet} & Real & SSIM $\uparrow$ & 0.725 & 0.613 & 0.711 & 0.793 & \textcolor{red}{0.882}\\
RCD~\cite{Cui2024UEC} & Raw-Syn & SSIM $\uparrow$ & 0.782 & 0.654 & 0.735 & 0.856 & 0.856\\

Gamma~\cite{Lore2017LLNet} & RGB-Syn & SSIM $\uparrow$ & 0.689 & 0.539 & 0.708 & 0.729 & 0.752\\
CSNorm-Fourier~\cite{Yao2023CSNorm} & RGB-Syn & SSIM $\uparrow$ & 0.744 & 0.585 & 0.714 & 0.782 & 0.804\\
UEC-based~\cite{Cui2024UEC} & RGB-Syn & SSIM $\uparrow$ & 0.779 & 0.631 & 0.752 & 0.828 & 0.841\\
MESS-generated & RGB-Syn & SSIM $\uparrow$ & \textcolor{blue}{0.855} & \textcolor{blue}{0.689} & \textcolor{blue}{0.798} & \textcolor{red}{0.894} & \textcolor{blue}{0.873}\\
\hline
\hline

UHDLL~\cite{li2023uhdfour} & Real & LPIPS $\downarrow$ & \textcolor{red}{0.127} & 0.218 & 0.240 & 0.158 & 0.138\\
LSRW~\cite{Hai2023R2RNet} & Real & LPIPS $\downarrow$ & 0.154 & \textcolor{red}{0.190} & 0.235 & 0.165 & 0.158\\
SICE~\cite{Cai2018SICE} & Real & LPIPS $\downarrow$ & 0.202 & 0.224 & \textcolor{red}{0.177} & 0.143 & 0.156\\
MSEC~\cite{Afifi2021Learning} & Raw-Syn & LPIPS $\downarrow$ & 0.184 & 0.218 & 0.206 & \textcolor{red}{0.121} & 0.153\\
BAID~\cite{Lv2022BacklitNet} & Real & LPIPS $\downarrow$ & 0.229 & 0.267 & 0.247 & 0.187 & \textcolor{red}{0.121}\\
RCD~\cite{Cui2024UEC} & Raw-Syn & LPIPS $\downarrow$ & 0.185 & 0.229 & 0.228 & 0.145 & 0.145\\

Gamma~\cite{Lore2017LLNet} & RGB-Syn & LPIPS $\downarrow$ & 0.240 & 0.331 & 0.298 & 0.230 & 0.241\\
CSNorm-Fourier~\cite{Yao2023CSNorm} & RGB-Syn & LPIPS $\downarrow$ & 0.278 & 0.284 & 0.257 & 0.205 & 0.201\\
UEC-based~\cite{Cui2024UEC} & RGB-Syn & LPIPS $\downarrow$ & 0.198 & 0.250 & 0.211 & 0.178 & 0.165\\
MESS-generated & RGB-Syn & LPIPS $\downarrow$ & \textcolor{blue}{0.134} & \textcolor{blue}{0.197} & \textcolor{blue}{0.184} & \textcolor{blue}{0.125} & \textcolor{blue}{0.125}\\
\hline
\end{tabular}
\end{table*}

\begin{table*}[t]
\centering
\caption{No-reference results using ExSLOT~\cite{jung2025exposure_slot} backbone. The model trained on each data source is evaluated on all test sets without fine-tuning. Values in \textcolor{red}{red} and \textcolor{blue}{blue} indicate the best and second-best results, respectively.}
\label{tab:exslot_nriqa}
\small
\setlength{\tabcolsep}{4pt}
\begin{tabular}{l|c|c|ccccc}
\hline
Training Data & Type & Metric & UHDLL & LSRW & SICE & MSEC & BAID\\
\hline

UHDLL~\cite{li2023uhdfour} & Real & ARNIQA $\uparrow$ & \textcolor{blue}{0.661} & 0.650 & 0.693 & 0.692 & 0.680\\
LSRW~\cite{Hai2023R2RNet} & Real & ARNIQA $\uparrow$ & 0.640 & \textcolor{red}{0.673} & 0.690 & 0.690 & 0.675\\
SICE~\cite{Cai2018SICE} & Real & ARNIQA $\uparrow$ & 0.646 & 0.633 & \textcolor{red}{0.727} & 0.709 & 0.669\\
MSEC~\cite{Afifi2021Learning} & Raw-Syn & ARNIQA $\uparrow$ & 0.641 & 0.652 & 0.720 & \textcolor{red}{0.728} & 0.687\\
BAID~\cite{Lv2022BacklitNet} & Real & ARNIQA $\uparrow$ & 0.610 & 0.613 & 0.654 & 0.669 & \textcolor{blue}{0.705}\\
RCD~\cite{Cui2024UEC} & Raw-Syn & ARNIQA $\uparrow$ & 0.639 & 0.649 & 0.707 & 0.707 & 0.682\\

Gamma~\cite{Lore2017LLNet} & RGB-Syn & ARNIQA $\uparrow$ & 0.604 & 0.607 & 0.664 & 0.673 & 0.662\\
CSNorm-Fourier~\cite{Yao2023CSNorm} & RGB-Syn & ARNIQA $\uparrow$ & 0.610 & 0.615 & 0.678 & 0.682 & 0.675\\
UEC-based~\cite{Cui2024UEC} & RGB-Syn & ARNIQA $\uparrow$ & 0.633 & 0.640 & 0.694 & 0.700 & 0.695\\
MESS-generated & RGB-Syn & ARNIQA $\uparrow$ & \textcolor{red}{0.678} & \textcolor{blue}{0.663} & \textcolor{blue}{0.721} & \textcolor{blue}{0.725} & \textcolor{red}{0.716}\\
\hline
\hline

UHDLL~\cite{li2023uhdfour} & Real & MUSIQ $\uparrow$ & \textcolor{blue}{57.446} & 50.672 & 52.614 & 52.452 & 56.011\\
LSRW~\cite{Hai2023R2RNet} & Real & MUSIQ $\uparrow$ & 56.085 & \textcolor{red}{52.205} & 52.996 & 53.064 & 56.266\\
SICE~\cite{Cai2018SICE} & Real & MUSIQ $\uparrow$ & 55.039 & 50.460 & \textcolor{blue}{56.700} & 54.161 & 55.777\\
MSEC~\cite{Afifi2021Learning} & Raw-Syn & MUSIQ $\uparrow$ & 55.782 & 50.815 & 54.446 & \textcolor{blue}{55.648} & 55.843\\
BAID~\cite{Lv2022BacklitNet} & Real & MUSIQ $\uparrow$ & 52.403 & 49.264 & 52.691 & 51.901 & \textcolor{blue}{58.217}\\
RCD~\cite{Cui2024UEC} & Raw-Syn & MUSIQ $\uparrow$ & 55.431 & 50.475 & 53.725 & 53.998 & 56.359\\

Gamma~\cite{Lore2017LLNet} & RGB-Syn & MUSIQ $\uparrow$ & 52.746 & 48.979 & 53.211 & 52.339 & 55.288\\
CSNorm-Fourier~\cite{Yao2023CSNorm} & RGB-Syn & MUSIQ $\uparrow$ & 53.669 & 49.489 & 53.941 & 53.037 & 55.923\\
UEC-based~\cite{Cui2024UEC} & RGB-Syn & MUSIQ $\uparrow$ & 54.643 & 49.963 & 54.779 & 53.801 & 56.759\\
MESS-generated & RGB-Syn & MUSIQ $\uparrow$ & \textcolor{red}{57.678} & \textcolor{blue}{52.112} & \textcolor{red}{57.406} & \textcolor{red}{55.814} & \textcolor{red}{58.441}\\
\hline
\hline

UHDLL~\cite{li2023uhdfour} & Real & MANIQA $\uparrow$ & 0.444 & 0.387 & 0.423 & 0.455 & 0.410\\
LSRW~\cite{Hai2023R2RNet} & Real & MANIQA $\uparrow$ & 0.440 & \textcolor{blue}{0.397} & 0.430 & 0.446 & 0.401\\
SICE~\cite{Cai2018SICE} & Real & MANIQA $\uparrow$ & 0.428 & 0.378 & \textcolor{blue}{0.452} & 0.460 & 0.399\\
MSEC~\cite{Afifi2021Learning} & Raw-Syn & MANIQA $\uparrow$ & \textcolor{blue}{0.446} & 0.380 & 0.447 & \textcolor{blue}{0.486} & 0.408\\
BAID~\cite{Lv2022BacklitNet} & Real & MANIQA $\uparrow$ & 0.400 & 0.365 & 0.413 & 0.434 & \textcolor{blue}{0.426}\\
RCD~\cite{Cui2024UEC} & Raw-Syn & MANIQA $\uparrow$ & 0.433 & 0.373 & 0.431 & 0.468 & 0.415\\

Gamma~\cite{Lore2017LLNet} & RGB-Syn & MANIQA $\uparrow$ & 0.396 & 0.362 & 0.412 & 0.420 & 0.385\\
CSNorm-Fourier~\cite{Yao2023CSNorm} & RGB-Syn & MANIQA $\uparrow$ & 0.412 & 0.370 & 0.420 & 0.441 & 0.394\\
UEC-based~\cite{Cui2024UEC} & RGB-Syn & MANIQA $\uparrow$ & 0.427 & 0.388 & 0.434 & 0.450 & 0.402\\
MESS-generated & RGB-Syn & MANIQA $\uparrow$ & \textcolor{red}{0.456} & \textcolor{red}{0.407} & \textcolor{red}{0.453} & \textcolor{red}{0.493} & \textcolor{red}{0.434}\\
\hline
\end{tabular}
\end{table*}

\begin{table*}[t]
\centering
\caption{Full-reference results using CSEC~\cite{Li2024CSEC} backbone. The model trained on each data source is evaluated on all test sets without fine-tuning. Values in \textcolor{red}{red} and \textcolor{blue}{blue} indicate the best and second-best results, respectively.}
\label{tab:csec_friqa}
\small
\setlength{\tabcolsep}{4pt}
\begin{tabular}{l|c|c|ccccc}
\hline
Training Data & Type & Metric & UHDLL & LSRW & SICE & MSEC & BAID\\
\hline

UHDLL~\cite{li2023uhdfour} & Real & PSNR $\uparrow$ & \textcolor{red}{19.145} & 16.386 & 11.687 & 18.706 & 15.269\\
LSRW~\cite{Hai2023R2RNet} & Real & PSNR $\uparrow$ & 16.860 & \textcolor{red}{17.850} & 12.807 & 18.961 & 14.696\\
SICE~\cite{Cai2018SICE} & Real & PSNR $\uparrow$ & 14.421 & 15.039 & \textcolor{blue}{17.321} & 20.334 & 14.905\\
MSEC~\cite{Afifi2021Learning} & Raw-Syn & PSNR $\uparrow$ & 14.824 & 15.197 & 15.910 & \textcolor{blue}{22.728} & 15.169\\
BAID~\cite{Lv2022BacklitNet} & Real & PSNR $\uparrow$ & 12.278 & 13.928 & 13.841 & 18.745 & \textcolor{red}{16.763}\\
RCD~\cite{Cui2024UEC} & Raw-Syn & PSNR $\uparrow$ & 15.177 & 15.184 & 14.654 & 20.761 & 15.508\\

Gamma~\cite{Lore2017LLNet} & RGB-Syn & PSNR $\uparrow$ & 13.139 & 11.701 & 12.778 & 17.687 & 12.958\\
CSNorm-Fourier~\cite{Yao2023CSNorm} & RGB-Syn & PSNR $\uparrow$ & 13.526 & 13.433 & 14.397 & 19.204 & 14.217\\
UEC-based~\cite{Cui2024UEC} & RGB-Syn & PSNR $\uparrow$ & 14.564 & 14.845 & 15.787 & 20.580 & 15.306\\
MESS-generated & RGB-Syn & PSNR $\uparrow$ & \textcolor{blue}{18.867} & \textcolor{blue}{17.515} & \textcolor{red}{17.937} & \textcolor{red}{22.866} & \textcolor{blue}{16.502}\\
\hline
\hline

UHDLL~\cite{li2023uhdfour} & Real & SSIM $\uparrow$ & \textcolor{red}{0.873} & 0.672 & 0.701 & 0.819 & 0.857\\
LSRW~\cite{Hai2023R2RNet} & Real & SSIM $\uparrow$ & 0.852 & \textcolor{red}{0.702} & 0.725 & 0.814 & 0.846\\
SICE~\cite{Cai2018SICE} & Real & SSIM $\uparrow$ & 0.782 & 0.661 & \textcolor{red}{0.805} & 0.842 & 0.850\\
MSEC~\cite{Afifi2021Learning} & Raw-Syn & SSIM $\uparrow$ & 0.789 & 0.666 & 0.756 & \textcolor{red}{0.873} & 0.847\\
BAID~\cite{Lv2022BacklitNet} & Real & SSIM $\uparrow$ & 0.706 & 0.611 & 0.706 & 0.791 & \textcolor{red}{0.885}\\
RCD~\cite{Cui2024UEC} & Raw-Syn & SSIM $\uparrow$ & 0.784 & 0.659 & 0.736 & 0.859 & 0.854\\

Gamma~\cite{Lore2017LLNet} & RGB-Syn & SSIM $\uparrow$ & 0.740 & 0.539 & 0.653 & 0.727 & 0.807\\
CSNorm-Fourier~\cite{Yao2023CSNorm} & RGB-Syn & SSIM $\uparrow$ & 0.756 & 0.587 & 0.710 & 0.781 & 0.756\\
UEC-based~\cite{Cui2024UEC} & RGB-Syn & SSIM $\uparrow$ & 0.793 & 0.630 & 0.744 & 0.828 & 0.843\\
MESS-generated & RGB-Syn & SSIM $\uparrow$ & \textcolor{blue}{0.869} & \textcolor{blue}{0.689} & \textcolor{blue}{0.792} & \textcolor{blue}{0.868} & \textcolor{blue}{0.876}\\
\hline
\hline

UHDLL~\cite{li2023uhdfour} & Real & LPIPS $\downarrow$ & \textcolor{red}{0.103} & 0.218 & 0.235 & 0.158 & 0.127\\
LSRW~\cite{Hai2023R2RNet} & Real & LPIPS $\downarrow$ & 0.131 & \textcolor{red}{0.190} & 0.227 & 0.163 & 0.147\\
SICE~\cite{Cai2018SICE} & Real & LPIPS $\downarrow$ & 0.174 & 0.221 & \textcolor{red}{0.170} & 0.145 & 0.145\\
MSEC~\cite{Afifi2021Learning} & Raw-Syn & LPIPS $\downarrow$ & 0.162 & 0.220 & 0.196 & \textcolor{blue}{0.128} & 0.141\\
BAID~\cite{Lv2022BacklitNet} & Real & LPIPS $\downarrow$ & 0.202 & 0.267 & 0.240 & 0.189 & \textcolor{red}{0.110}\\
RCD~\cite{Cui2024UEC} & Raw-Syn & LPIPS $\downarrow$ & 0.154 & 0.226 & 0.218 & 0.148 & 0.134\\

Gamma~\cite{Lore2017LLNet} & RGB-Syn & LPIPS $\downarrow$ & 0.255 & 0.333 & 0.290 & 0.233 & 0.229\\
CSNorm-Fourier~\cite{Yao2023CSNorm} & RGB-Syn & LPIPS $\downarrow$ & 0.218 & 0.283 & 0.251 & 0.203 & 0.194\\
UEC-based~\cite{Cui2024UEC} & RGB-Syn & LPIPS $\downarrow$ & 0.174 & 0.250 & 0.205 & 0.180 & 0.154\\
MESS-generated & RGB-Syn & LPIPS $\downarrow$ & \textcolor{blue}{0.110} & \textcolor{blue}{0.197} & \textcolor{blue}{0.177} & \textcolor{red}{0.126} & \textcolor{blue}{0.114}\\
\hline
\end{tabular}
\end{table*}

\begin{table*}[t]
\centering
\caption{No-reference results using CSEC~\cite{Li2024CSEC} backbone. The model trained on each data source is evaluated on all test sets without fine-tuning. Values in \textcolor{red}{red} and \textcolor{blue}{blue} indicate the best and second-best results, respectively.}
\label{tab:csec_nriqa}
\small
\setlength{\tabcolsep}{4pt}
\begin{tabular}{l|c|c|ccccc}
\hline
Training Data & Type & Metric & UHDLL & LSRW & SICE & MSEC & BAID\\
\hline

UHDLL~\cite{li2023uhdfour} & Real & ARNIQA $\uparrow$ & \textcolor{blue}{0.682} & 0.659 & 0.708 & 0.692 & 0.700\\
LSRW~\cite{Hai2023R2RNet} & Real & ARNIQA $\uparrow$ & 0.661 & \textcolor{red}{0.682} & 0.701 & 0.692 & 0.695\\
SICE~\cite{Cai2018SICE} & Real & ARNIQA $\uparrow$ & 0.669 & 0.643 & \textcolor{blue}{0.734} & 0.709 & 0.690\\
MSEC~\cite{Afifi2021Learning} & Raw-Syn & ARNIQA $\uparrow$ & 0.664 & 0.657 & 0.725 & \textcolor{blue}{0.722} & 0.701\\
BAID~\cite{Lv2022BacklitNet} & Real & ARNIQA $\uparrow$ & 0.629 & 0.623 & 0.668 & 0.667 & \textcolor{red}{0.734}\\
RCD~\cite{Cui2024UEC} & Raw-Syn & ARNIQA $\uparrow$ & 0.664 & 0.655 & 0.719 & 0.706 & 0.705\\

Gamma~\cite{Lore2017LLNet} & RGB-Syn & ARNIQA $\uparrow$ & 0.627 & 0.614 & 0.674 & 0.674 & 0.682\\
CSNorm-Fourier~\cite{Yao2023CSNorm} & RGB-Syn & ARNIQA $\uparrow$ & 0.633 & 0.623 & 0.689 & 0.680 & 0.691\\
UEC-based~\cite{Cui2024UEC} & RGB-Syn & ARNIQA $\uparrow$ & 0.654 & 0.649 & 0.710 & 0.701 & 0.713\\
MESS-generated & RGB-Syn & ARNIQA $\uparrow$ & \textcolor{red}{0.699} & \textcolor{blue}{0.672} & \textcolor{red}{0.740} & \textcolor{red}{0.729} & \textcolor{blue}{0.723}\\
\hline
\hline

UHDLL~\cite{li2023uhdfour} & Real & MUSIQ $\uparrow$ & \textcolor{blue}{57.028} & 50.210 & 52.349 & 52.464 & 55.537\\
LSRW~\cite{Hai2023R2RNet} & Real & MUSIQ $\uparrow$ & 55.667 & \textcolor{red}{51.738} & 52.746 & 53.064 & 55.811\\
SICE~\cite{Cai2018SICE} & Real & MUSIQ $\uparrow$ & 54.582 & 49.989 & \textcolor{blue}{56.478} & 54.142 & 55.266\\
MSEC~\cite{Afifi2021Learning} & Raw-Syn & MUSIQ $\uparrow$ & 55.360 & 50.302 & 54.223 & \textcolor{blue}{55.400} & 55.376\\
BAID~\cite{Lv2022BacklitNet} & Real & MUSIQ $\uparrow$ & 51.984 & 48.770 & 52.446 & 51.863 & \textcolor{blue}{57.750}\\
RCD~\cite{Cui2024UEC} & Raw-Syn & MUSIQ $\uparrow$ & 55.017 & 49.972 & 53.494 & 53.997 & 55.884\\

Gamma~\cite{Lore2017LLNet} & RGB-Syn & MUSIQ $\uparrow$ & 52.359 & 48.506 & 53.002 & 52.338 & 54.791\\
CSNorm-Fourier~\cite{Yao2023CSNorm} & RGB-Syn & MUSIQ $\uparrow$ & 53.274 & 48.989 & 53.752 & 53.060 & 55.466\\
UEC-based~\cite{Cui2024UEC} & RGB-Syn & MUSIQ $\uparrow$ & 54.173 & 49.521 & 54.570 & 53.769 & 56.253\\
MESS-generated & RGB-Syn & MUSIQ $\uparrow$ & \textcolor{red}{57.260} & \textcolor{blue}{51.645} & \textcolor{red}{57.184} & \textcolor{red}{55.554} & \textcolor{red}{57.974}\\
\hline
\hline

UHDLL~\cite{li2023uhdfour} & Real & MANIQA $\uparrow$ & \textcolor{blue}{0.448} & 0.384 & 0.427 & 0.455 & 0.417\\
LSRW~\cite{Hai2023R2RNet} & Real & MANIQA $\uparrow$ & 0.447 & \textcolor{blue}{0.393} & 0.433 & 0.446 & 0.411\\
SICE~\cite{Cai2018SICE} & Real & MANIQA $\uparrow$ & 0.432 & 0.374 & \textcolor{blue}{0.455} & 0.460 & 0.408\\
MSEC~\cite{Afifi2021Learning} & Raw-Syn & MANIQA $\uparrow$ & 0.433 & 0.378 & 0.452 & \textcolor{blue}{0.479} & 0.419\\
BAID~\cite{Lv2022BacklitNet} & Real & MANIQA $\uparrow$ & 0.403 & 0.360 & 0.415 & 0.434 & \textcolor{blue}{0.435}\\
RCD~\cite{Cui2024UEC} & Raw-Syn & MANIQA $\uparrow$ & 0.437 & 0.373 & 0.443 & 0.461 & 0.421\\

Gamma~\cite{Lore2017LLNet} & RGB-Syn & MANIQA $\uparrow$ & 0.400 & 0.360 & 0.414 & 0.422 & 0.394\\
CSNorm-Fourier~\cite{Yao2023CSNorm} & RGB-Syn & MANIQA $\uparrow$ & 0.418 & 0.366 & 0.427 & 0.440 & 0.401\\
UEC-based~\cite{Cui2024UEC} & RGB-Syn & MANIQA $\uparrow$ & 0.431 & 0.388 & 0.434 & 0.450 & 0.414\\
MESS-generated & RGB-Syn & MANIQA $\uparrow$ & \textcolor{red}{0.460} & \textcolor{red}{0.403} & \textcolor{red}{0.461} & \textcolor{red}{0.486} & \textcolor{red}{0.443}\\
\hline
\end{tabular}
\end{table*}

\begin{table*}[t]
\centering
\caption{Full-reference results using RAVE~\cite{gaintseva2024RAVE} backbone. The model trained on each data source is evaluated on all test sets without fine-tuning. Values in \textcolor{red}{red} and \textcolor{blue}{blue} indicate the best and second-best results, respectively.}
\label{tab:rave_friqa}
\small
\setlength{\tabcolsep}{4pt}
\begin{tabular}{l|c|c|ccccc}
\hline
Training Data & Type & Metric & UHDLL & LSRW & SICE & MSEC & BAID\\
\hline

UHDLL~\cite{li2023uhdfour} & Real & PSNR $\uparrow$ & \textcolor{red}{19.276} & 16.488 & 12.454 & 18.718 & 16.573\\
LSRW~\cite{Hai2023R2RNet} & Real & PSNR $\uparrow$ & 16.983 & \textcolor{red}{18.003} & 13.557 & 18.981 & 15.976\\
SICE~\cite{Cai2018SICE} & Real & PSNR $\uparrow$ & 14.521 & 15.214 & \textcolor{red}{18.673} & 20.311 & 16.182\\
MSEC~\cite{Afifi2021Learning} & Raw-Syn & PSNR $\uparrow$ & 14.922 & 15.362 & 16.650 & \textcolor{red}{22.001} & 16.473\\
BAID~\cite{Lv2022BacklitNet} & Real & PSNR $\uparrow$ & 13.249 & 14.094 & 14.574 & 18.772 & \textcolor{red}{18.035}\\
RCD~\cite{Cui2024UEC} & Raw-Syn & PSNR $\uparrow$ & 15.459 & 15.332 & 15.398 & 20.761 & 16.803\\

Gamma~\cite{Lore2017LLNet} & RGB-Syn & PSNR $\uparrow$ & 12.441 & 13.639 & 13.513 & 17.681 & 14.222\\
CSNorm-Fourier~\cite{Yao2023CSNorm} & RGB-Syn & PSNR $\uparrow$ & 13.668 & 11.857 & 15.140 & 19.249 & 15.470\\
UEC-based~\cite{Cui2024UEC} & RGB-Syn & PSNR $\uparrow$ & 14.709 & 15.006 & 16.531 & 20.536 & 16.586\\
MESS-generated & RGB-Syn & PSNR $\uparrow$ & \textcolor{blue}{18.998} & \textcolor{blue}{17.668} & \textcolor{blue}{18.365} & \textcolor{blue}{21.869} & \textcolor{blue}{17.774}\\
\hline
\hline

UHDLL~\cite{li2023uhdfour} & Real & SSIM $\uparrow$ & \textcolor{red}{0.857} & 0.671 & 0.709 & 0.820 & 0.855\\
LSRW~\cite{Hai2023R2RNet} & Real & SSIM $\uparrow$ & 0.833 & \textcolor{red}{0.702} & 0.735 & 0.813 & 0.841\\
SICE~\cite{Cai2018SICE} & Real & SSIM $\uparrow$ & 0.769 & 0.661 & \textcolor{blue}{0.799} & 0.828 & 0.845\\
MSEC~\cite{Afifi2021Learning} & Raw-Syn & SSIM $\uparrow$ & 0.775 & 0.670 & 0.763 & \textcolor{blue}{0.870} & 0.845\\
BAID~\cite{Lv2022BacklitNet} & Real & SSIM $\uparrow$ & 0.724 & 0.614 & 0.712 & 0.791 & \textcolor{red}{0.881}\\
RCD~\cite{Cui2024UEC} & Raw-Syn & SSIM $\uparrow$ & 0.785 & 0.651 & 0.743 & 0.855 & 0.855\\

Gamma~\cite{Lore2017LLNet} & RGB-Syn & SSIM $\uparrow$ & 0.689 & 0.538 & 0.661 & 0.781 & 0.752\\
CSNorm-Fourier~\cite{Yao2023CSNorm} & RGB-Syn & SSIM $\uparrow$ & 0.743 & 0.584 & 0.716 & 0.729 & 0.804\\
UEC-based~\cite{Cui2024UEC} & RGB-Syn & SSIM $\uparrow$ & 0.775 & 0.630 & 0.754 & 0.829 & 0.840\\
MESS-generated & RGB-Syn & SSIM $\uparrow$ & \textcolor{blue}{0.853} & \textcolor{blue}{0.689} & \textcolor{red}{0.812} & \textcolor{red}{0.874} & \textcolor{blue}{0.872}\\
\hline
\hline

UHDLL~\cite{li2023uhdfour} & Real & LPIPS $\downarrow$ & \textcolor{red}{0.132} & 0.219 & 0.243 & 0.157 & 0.144\\
LSRW~\cite{Hai2023R2RNet} & Real & LPIPS $\downarrow$ & 0.161 & \textcolor{red}{0.189} & 0.235 & 0.164 & 0.160\\
SICE~\cite{Cai2018SICE} & Real & LPIPS $\downarrow$ & 0.206 & 0.222 & \textcolor{red}{0.178} & 0.143 & 0.161\\
MSEC~\cite{Afifi2021Learning} & Raw-Syn & LPIPS $\downarrow$ & 0.191 & 0.217 & 0.206 & \textcolor{blue}{0.141} & 0.152\\
BAID~\cite{Lv2022BacklitNet} & Real & LPIPS $\downarrow$ & 0.230 & 0.266 & 0.248 & 0.186 & \textcolor{red}{0.124}\\
RCD~\cite{Cui2024UEC} & Raw-Syn & LPIPS $\downarrow$ & 0.181 & 0.227 & 0.227 & 0.144 & 0.147\\

Gamma~\cite{Lore2017LLNet} & RGB-Syn & LPIPS $\downarrow$ & 0.283 & 0.334 & 0.299 & 0.232 & 0.243\\
CSNorm-Fourier~\cite{Yao2023CSNorm} & RGB-Syn & LPIPS $\downarrow$ & 0.247 & 0.284 & 0.258 & 0.205 & 0.205\\
UEC-based~\cite{Cui2024UEC} & RGB-Syn & LPIPS $\downarrow$ & 0.203 & 0.246 & 0.214 & 0.180 & 0.168\\
MESS-generated & RGB-Syn & LPIPS $\downarrow$ & \textcolor{blue}{0.139} & \textcolor{blue}{0.196} & \textcolor{blue}{0.185} & \textcolor{red}{0.138} & \textcolor{blue}{0.128}\\
\hline
\end{tabular}
\end{table*}

\begin{table*}[t]
\centering
\caption{No-reference results using RAVE~\cite{gaintseva2024RAVE} backbone. The model trained on each data source is evaluated on all test sets without fine-tuning. Values in \textcolor{red}{red} and \textcolor{blue}{blue} indicate the best and second-best results, respectively.}
\label{tab:rave_nriqa}
\small
\setlength{\tabcolsep}{4pt}
\begin{tabular}{l|c|c|ccccc}
\hline
Training Data & Type & Metric & UHDLL & LSRW & SICE & MSEC & BAID\\
\hline

UHDLL~\cite{li2023uhdfour} & Real & ARNIQA $\uparrow$ & \textcolor{blue}{0.657} & 0.650 & 0.692 & 0.693 & 0.679\\
LSRW~\cite{Hai2023R2RNet} & Real & ARNIQA $\uparrow$ & 0.637 & \textcolor{red}{0.672} & 0.687 & 0.692 & 0.672\\
SICE~\cite{Cai2018SICE} & Real & ARNIQA $\uparrow$ & 0.642 & 0.633 & \textcolor{blue}{0.718} & 0.699 & 0.667\\
MSEC~\cite{Afifi2021Learning} & Raw-Syn & ARNIQA $\uparrow$ & 0.638 & 0.645 & 0.708 & 0.697 & 0.680\\
BAID~\cite{Lv2022BacklitNet} & Real & ARNIQA $\uparrow$ & 0.603 & 0.613 & 0.649 & 0.668 & \textcolor{blue}{0.701}\\
RCD~\cite{Cui2024UEC} & Raw-Syn & ARNIQA $\uparrow$ & 0.632 & 0.643 & 0.691 & \textcolor{blue}{0.702} & 0.689\\

Gamma~\cite{Lore2017LLNet} & RGB-Syn & ARNIQA $\uparrow$ & 0.601 & 0.604 & 0.660 & 0.671 & 0.658\\
CSNorm-Fourier~\cite{Yao2023CSNorm} & RGB-Syn & ARNIQA $\uparrow$ & 0.608 & 0.614 & 0.675 & 0.680 & 0.667\\
UEC-based~\cite{Cui2024UEC} & RGB-Syn & ARNIQA $\uparrow$ & 0.629 & 0.641 & 0.691 & 0.700 & 0.691\\
MESS-generated & RGB-Syn & ARNIQA $\uparrow$ & \textcolor{red}{0.674} & \textcolor{blue}{0.662} & \textcolor{red}{0.724} & \textcolor{red}{0.704} & \textcolor{red}{0.712}\\
\hline
\hline

UHDLL~\cite{li2023uhdfour} & Real & MUSIQ $\uparrow$ & \textcolor{blue}{57.529} & 50.730 & 52.641 & 52.449 & 56.127\\
LSRW~\cite{Hai2023R2RNet} & Real & MUSIQ $\uparrow$ & 56.168 & \textcolor{red}{52.299} & 53.067 & 53.058 & 56.357\\
SICE~\cite{Cai2018SICE} & Real & MUSIQ $\uparrow$ & 55.096 & 50.518 & \textcolor{blue}{56.744} & 54.119 & 55.867\\
MSEC~\cite{Afifi2021Learning} & Raw-Syn & MUSIQ $\uparrow$ & 55.830 & 50.856 & 54.476 & \textcolor{blue}{54.413} & 55.936\\
BAID~\cite{Lv2022BacklitNet} & Real & MUSIQ $\uparrow$ & 52.473 & 49.331 & 52.705 & 51.879 & \textcolor{blue}{58.310}\\
RCD~\cite{Cui2024UEC} & Raw-Syn & MUSIQ $\uparrow$ & 55.492 & 50.526 & 53.759 & 53.994 & 56.447\\

Gamma~\cite{Lore2017LLNet} & RGB-Syn & MUSIQ $\uparrow$ & 52.846 & 49.091 & 53.247 & 52.308 & 55.360\\
CSNorm-Fourier~\cite{Yao2023CSNorm} & RGB-Syn & MUSIQ $\uparrow$ & 53.735 & 49.582 & 54.003 & 53.045 & 55.989\\
UEC-based~\cite{Cui2024UEC} & RGB-Syn & MUSIQ $\uparrow$ & 54.678 & 50.033 & 54.834 & 53.760 & 56.816\\
MESS-generated & RGB-Syn & MUSIQ $\uparrow$ & \textcolor{red}{57.761} & \textcolor{blue}{52.206} & \textcolor{red}{57.450} & \textcolor{red}{54.573} & \textcolor{red}{58.534}\\
\hline
\hline

UHDLL~\cite{li2023uhdfour} & Real & MANIQA $\uparrow$ & \textcolor{blue}{0.443} & 0.388 & 0.422 & 0.452 & 0.406\\
LSRW~\cite{Hai2023R2RNet} & Real & MANIQA $\uparrow$ & 0.438 & \textcolor{blue}{0.398} & 0.430 & 0.446 & 0.400\\
SICE~\cite{Cai2018SICE} & Real & MANIQA $\uparrow$ & 0.430 & 0.378 & \textcolor{blue}{0.451} & \textcolor{blue}{0.460} & 0.399\\
MSEC~\cite{Afifi2021Learning} & Raw-Syn & MANIQA $\uparrow$ & 0.426 & 0.381 & 0.448 & 0.449 & 0.408\\
BAID~\cite{Lv2022BacklitNet} & Real & MANIQA $\uparrow$ & 0.395 & 0.365 & 0.412 & 0.433 & \textcolor{blue}{0.424}\\
RCD~\cite{Cui2024UEC} & Raw-Syn & MANIQA $\uparrow$ & 0.423 & 0.379 & 0.436 & \textcolor{red}{0.469} & 0.419\\

Gamma~\cite{Lore2017LLNet} & RGB-Syn & MANIQA $\uparrow$ & 0.397 & 0.363 & 0.409 & 0.422 & 0.380\\
CSNorm-Fourier~\cite{Yao2023CSNorm} & RGB-Syn & MANIQA $\uparrow$ & 0.414 & 0.370 & 0.420 & 0.440 & 0.389\\
UEC-based~\cite{Cui2024UEC} & RGB-Syn & MANIQA $\uparrow$ & 0.425 & 0.392 & 0.433 & 0.452 & 0.402\\
MESS-generated & RGB-Syn & MANIQA $\uparrow$ & \textcolor{red}{0.455} & \textcolor{red}{0.408} & \textcolor{red}{0.457} & 0.445 & \textcolor{red}{0.432}\\
\hline
\end{tabular}
\end{table*}

\begin{table}[t]
\centering
\caption{
Ablation study of the proposed MESS framework.
All variants are used to generate training data for the same FourLLIE backbone, and the trained models are evaluated on five test sets without fine-tuning.
The reported results are averaged over UHDLL, LSRW, SICE, MSEC, and BAID.
}
\label{tab:ablation_main}
\resizebox{\linewidth}{!}{
\begin{tabular}{lcccccc}
\toprule
Variant & PSNR $\uparrow$ & SSIM $\uparrow$ & LPIPS $\downarrow$ & ARNIQA $\uparrow$ & MUSIQ $\uparrow$ & MANIQA $\uparrow$ \\
\midrule
w/o representation consistency loss      & 21.354 & 0.835 & 0.132 & 0.698 & 54.821 & 0.456 \\
w/o frequency-based enhancement           & 20.936 & 0.828 & 0.140 & 0.692 & 54.307 & 0.449 \\
w/o diffusion-based detail enhancement    & 21.112 & 0.832 & 0.136 & 0.695 & 54.563 & 0.453 \\
w/o exposure-conditioned rendering        & 20.781 & 0.821 & 0.145 & 0.688 & 54.018 & 0.445 \\
FFT instead of DWT                        & 21.447 & 0.837 & 0.130 & 0.700 & 54.963 & 0.458 \\
\textbf{Full MESS}                        & \textbf{21.923} & \textbf{0.846} & \textbf{0.121} & \textbf{0.709} & \textbf{55.510} & \textbf{0.467} \\
\bottomrule
\end{tabular}
}
\end{table}

\begin{table}[t]
\centering
\caption{
Ablation study on the design of the exposure-conditioned rendering module.
All variants use the same frequency-based enhancement module and diffusion-based detail enhancement module.
Only the way of injecting the target exposure condition is changed.
}
\label{tab:ablation_renderer}
\resizebox{\linewidth}{!}{
\begin{tabular}{lcccccc}
\toprule
Renderer Design & PSNR $\uparrow$ & SSIM $\uparrow$ & LPIPS $\downarrow$ & ARNIQA $\uparrow$ & MUSIQ $\uparrow$ & MANIQA $\uparrow$ \\
\midrule
No exposure condition                  & 20.781 & 0.821 & 0.145 & 0.688 & 54.018 & 0.445 \\
Concatenate EV with RGB features        & 21.082 & 0.831 & 0.137 & 0.695 & 54.576 & 0.452 \\
Global FiLM on RGB features             & 21.324 & 0.836 & 0.132 & 0.701 & 54.905 & 0.459 \\
DWT-FiLM w/o content descriptor         & 21.541 & 0.840 & 0.128 & 0.704 & 55.102 & 0.462 \\
\textbf{DWT-FiLM with content descriptor} & \textbf{21.923} & \textbf{0.846} & \textbf{0.121} & \textbf{0.709} & \textbf{55.510} & \textbf{0.467} \\
\bottomrule
\end{tabular}
}
\end{table}

\clearpage
\end{document}